\documentclass[sigconf]{acmart}

\usepackage{multirow}
\usepackage{algorithm, algpseudocode}
\usepackage{subcaption}
\usepackage{array} 
\usepackage{xcolor}

\AtBeginDocument{%
  }

\setcopyright{acmlicensed}
\copyrightyear{2018}
\acmYear{2018}
\acmDOI{XXXXXXX.XXXXXXX}
\acmConference[Conference acronym 'XX]{Make sure to enter the correct
  conference title from your rights confirmation email}{June 03--05,
  2018}{Woodstock, NY}
\acmISBN{978-1-4503-XXXX-X/2018/06}

\makeatletter
\renewcommand{\@copyrightpermission}{}
\renewcommand{\@copyrightowner}{}
\let\@printcopyright\relax
\let\@printpermission\relax

\def\acmConference#1#2#3#4{}
\def\acmDOI#1{}
\def\acmISBN#1{}
\def\copyrightyear#1{}
\def\setcopyright#1{}

\renewcommand{\footnotetextcopyrightpermission}[1]{}

\AtBeginDocument{\let\@mkbibcitation\@empty}

\fancypagestyle{firstpagestyle}{}
\makeatother

\begin{document}

\title{TDDM-Melatt: A Decoupled Memory and Diffusion Framework for Generalizable Encrypted Traffic Classification}


\author{Ze Chen}
\affiliation{
	\institution{National University of \\ Defense Technology}
	\city{Hefei}
	\state{Anhui}
	\country{China}
}
\email{chenze@nudt.edu.cn}

\author{Qiming Yu}
\affiliation{
	\institution{National University of \\ Defense Technology}
	\city{Hefei}
	\state{Anhui}
	\country{China}
}
\email{yuqiming24@nudt.edu.cn}

\author{Zijia Song}
\affiliation{
	\institution{National University of \\ Defense Technology}
	\city{Hefei}
	\state{Anhui}
	\country{China}
}
\email{songzijia@nudt.edu.cn}

\author{Guozheng Yang}
\authornote{Corresponding author.}
\affiliation{
	\institution{National University of \\ Defense Technology}
	\city{Hefei}
	\state{Anhui}
	\country{China}
}
\email{yangguozheng17@nudt.edu.cn}

\author{Wei Yan}
\affiliation{
	\institution{National University of \\ Defense Technology}
	\city{Hefei}
	\state{Anhui}
	\country{China}
}
\email{yan.wei2023@nudt.edu.cn}


\begin{abstract}
The widespread adoption of encrypted traffic poses severe challenges to current security situational awareness systems based on network traffic monitoring. In existing dataset-driven training and testing studies, limitations such as shortcut learning induced by spurious feature correlations and sample imbalance caused by the long-tail distribution of real-world traffic result in weak generalization of traffic identification performance to real-world network traffic. To address these limitations, we propose TDDM-Melatt, a disentangled memory-based traffic classification framework with diffusion-based data augmentation. First, we design Melatt, a memory-decoupled traffic representation model, which employs Competitive Gating Long Short-Term Memory (CG-LSTM) to construct the encoder and decoder. We design a spurious-correlation-free pre-training and inference paradigm, employing strict topology anonymization and a frozen pre-trained encoder strategy to cut off the model's learning pathways for spurious features. During pre-training, computing the reconstruction loss between different traffic classes and each memory prototype forces the model to learn prototype-aligned representations. During inference, classification is performed efficiently by a downstream classifier on the frozen representations. Second, we propose a Traffic Denoising Diffusion Model (TDDM) tailored to the characteristics of traffic data. Addressing the structured nature of high-dimensional sparsity and strong feature correlations in traffic data, we design a lightweight residual noise prediction network and a compact noise sampling mechanism, solving the problem of having only class labels as conditional guidance for traffic diffusion models. Extensive experiments are conducted on 4 representative public benchmark datasets. Under strict flow-level splitting and anonymization, TDDM-Melatt outperforms 6 basic classification models and 6 SOTA representation learning models. The proposed method provides a new and effective technical pathway for encrypted traffic classification in real-world network environments.
\end{abstract}




\maketitle

\section{Introduction}

Object recognition in the human brain relies on prototype memories stored in the hippocampus, forming a closed-loop cognitive process of ``memory storage--feature matching--reconstructive recognition.'' Visual information captured by the eyes is encoded into abstract features by the visual cortex and subsequently undergoes precise matching and reconstruction with memory prototypes via the hippocampus. High matching confidence leads to successful recognition, whereas low confidence results in classification as unknown or anomalous. This cognitive mechanism offers profound inspiration for network traffic inspection. If a ``memory module'' storing behavioral prototypes of traffic can be constructed, the reconstruction error naturally serves as a training objective when traffic of a given class cannot be accurately reconstructed by its own class-specific prototypes. Motivated by this insight, this paper focuses on encrypted traffic classification---a challenge in network security---and proposes a novel solution.

In recent years, with the increasing sophistication of cybersecurity threats and the strengthening of privacy protection regulations, the adoption of encryption protocols in Internet communications has exceeded 80\% \cite{TLS>80}. While encryption effectively safeguards user privacy, it also introduces substantial technical obstacles to network traffic monitoring and security auditing. Traditional Deep Packet Inspection (DPI) \cite{DPI,DPI-all} techniques are rendered ineffective due to the inability to decrypt encrypted payloads, compelling security researchers to shift toward deep learning approaches.

Explicit identifiers commonly retained in public datasets \cite{IDS,VPN,USTC,ET-BERT-TLS120}, such as IP, Port, and flow IDs, exhibit strong yet non-causal statistical correlations with traffic labels. Models tend to prioritize learning such ``shortcut features''. This reliance on spurious correlations severely degrades generalization capability, manifesting as inflated accuracy on public test sets followed by precipitous performance drops upon real-world deployment \cite{Pcap-Encoder,SoK}. Furthermore, most studies \cite{ET-BERT-TLS120,TrafficFormer} adopt packet-level sample partitioning, whereby packets from the same flow are distributed across both training and test sets. This practice causes models to learn flow-specific idiosyncrasies instead of class-generalizable patterns, further amplifying the detrimental effects of spurious correlations. In the context of fine-tuning traffic classification models, high accuracy on encrypted datasets has led to the controversial claim that pre-training enables models to extract patterns from encrypted payloads.

In real-world network environments, traffic exhibits highly imbalanced class distributions. Extreme data skewness causes models to be dominated by majority classes during training. Generative Adversarial Networks (GANs) \cite{GAN,WGAN-GP,ACGAN} have been explored to mitigate data imbalance. However, GANs are prone to training instability and mode collapse. Diffusion models, as an emerging class of generative models, have achieved remarkable success in image synthesis \cite{DDPM,Diffusion-img,CRoSS}, yet their application to network traffic remains in its infancy and struggles to accommodate the structured nature of traffic data.

To address the aforementioned limitations, we propose TDDM-Melatt, a decoupled memory and diffusion framework for generalizable encrypted traffic classification. The main contributions are summarized as follows:

\begin{itemize}
\item We propose Melatt, a modular traffic memory model. The architecture comprises a CG-LSTM-based encoder and a memory-augmented multi-head cross-attention decoder, enhancing the capacity to extract and reconstruct dynamic traffic information. 

\item We establish a pre-training and inference paradigm free of spurious correlations. During data preprocessing, explicit flow identifiers are rigorously removed, and timestamp information is generalized. Freezing the parameters of the pre-trained encoder during downstream classification tasks ensures the cross-scenario generalization capability in real network environments.

\item We design TDDM, a traffic denoising diffusion model tailored to the structural characteristics of traffic data. In light of the structured characteristics of traffic data, we construct a lightweight diffusion generation architecture that achieves a reduction of more than 30\% in the dimensionality of the features. By adopting a class-wise training paradigm, we propose a condition-guided compact noise sampling mechanism to improve class convergence during generation.

\item We conduct comprehensive experimental evaluations on 4 public datasets. Under strict flow-level sample partitioning and topology anonymization, TDDM-Melatt outperforms 6 baseline classifiers and 6 SOTA models.
\end{itemize}

\section{Problem Statement}

Encrypted traffic classification constitutes a fundamental capability within cyberspace security defense frameworks, and its technological evolution is deeply intertwined with the upgrade of encryption protocols, the mutation of network attack patterns, and the shifting landscape of regulatory requirements. With the widespread deployment of next-generation encryption protocols such as TLS 1.3 and QUIC, traditional payload-based DPI \cite{DPI,DPI-all} techniques have become increasingly ineffective, prompting a pivot toward representation learning-based classification methods as Table~\ref{tab:full_comparison}. ET-BERT \cite{ET-BERT-TLS120}, YaTC \cite{YaTC}, NetMamba \cite{NetMamba}, and TrafficFormer \cite{TrafficFormer} have achieved improved classification performance. Nevertheless, when migrating from laboratory settings to real-world environments, existing approaches frequently encounter precipitous performance degradation.

\begin{table}[tbp]
    \centering
    \small
    \caption{Comparison of the Traffic Analysis Methods.}
    \label{tab:full_comparison}
    \begin{tabular}{@{}lcccc@{}}
        \toprule
        \multirow{2}{*}{Methods} & \multirow{2}{*}{ML Models} & \multirow{2}{*}{Modality} & \multicolumn{2}{c}{Detection Abilities} \\
        \cmidrule(lr){4-5}
        & & & Encrypted & Non IP/Port \\
        \midrule
        netFound      & BERT & Flow           & \textcolor{blue}{\checkmark} & \textcolor{blue}{\checkmark} \\
        YaTC          & ViT   & Flow           & \textcolor{blue}{\checkmark} & \textcolor{blue}{\checkmark} \\
        NetMamba      & Mamba & Flow           & \textcolor{blue}{\checkmark} & $\textcolor{red}{\times}$ \\
        ET-BERT       & BERT  & Packet & \textcolor{blue}{\checkmark} & \textcolor{blue}{\checkmark} \\
        TrafficFormer & BERT  & Packet & \textcolor{blue}{\checkmark} & $\textcolor{red}{\times}$ \\
        Pcap-Encoder  & T5    & Flow         & \textcolor{blue}{\checkmark} & $\textcolor{red}{\times}$ \\
        \bottomrule
    \end{tabular}
\end{table}

The prevailing research paradigm in encrypted traffic classification relies on supervised learning over labeled datasets, implicitly assuming the existence of stable causal relationships between features and labels. This assumption, however, engenders a severe ``shortcut learning'' phenomenon, which has become a critical bottleneck restricting model generalization. Public datasets routinely retain explicit identifiers such as IP, Port, and flow IDs. These features exhibit strong yet non-causal correlations with traffic labels \cite{data-3,data-collect}. For instance, specific applications often utilize well-known ports, and particular attack traffic frequently originates from specific IP prefixes. During training, supervised models preferentially learn these ``low-cost, high-discriminability'' shortcut features. Such reliance on spurious correlations yields outstanding performance on test sets drawn from the same distribution as the training data \cite{Pcap-Encoder,SoK}, yet once deployed in real networks where IP and Port diverge, model performance collapses dramatically. The packet-level sample partitioning commonly adopted in existing studies \cite{ET-BERT-TLS120,TrafficFormer} further amplifies the detrimental effects of spurious correlations. Several works split distinct packets belonging to the same flow across training and test sets, thereby enabling models to learn flow-specific idiosyncrasies. This partitioning scheme, which inherently involves data leakage, leads to severe overestimation of actual model performance in experimental evaluations.

More critically, full fine-tuning (unfrozen encoder) of existing representation learning models severely erodes the generic features acquired during pre-training \cite{fine-tuning-1}. When deployment environments shift, the model forfeits both the generalization benefits of pre-trained knowledge and becomes excessively reliant on training-set-specific shortcut features, ultimately resulting in sharp performance attenuation in real-world network contexts. This phenomenon underscores a fundamental tension between theoretical innovation and practical deployment in encrypted traffic classification.

Another major challenge is the long-tail distribution of traffic in real network environments \cite{long-tail,long-tail-2}. In operational networks, benign traffic and flows from popular applications overwhelmingly dominate the total volume. In contrast, malicious traffic and niche applications flows are not only scarce in aggregate but also exhibit sample counts that may differ by several orders of magnitude across categories. Such extreme imbalance causes models to be dominated by majority class samples during training, leading to decision boundaries severely skewed against minority classes \cite{data-inba-2}. In pursuit of high overall accuracy, models tend to classify samples as majority classes, resulting in unacceptably low recognition rates for minority classes. In cybersecurity defense scenarios, however, it is precisely these exceedingly rare minority samples that carry the highest security value \cite{Net-ID}.

To mitigate the scarcity of minority class samples \cite{small-data}, data augmentation techniques have been extensively explored in encrypted traffic classification. In recent years, GANs \cite{GAN,WGAN-GP,ACGAN} have been applied to address data imbalance. However, GANs are plagued by training instability and mode collapse. Diffusion models, as an emerging class of generative models, have achieved breakthrough successes in image synthesis and natural language processing \cite{DDPM,Diffusion-img,CRoSS}, offering a promising new avenue for traffic data augmentation. Nonetheless, the application of diffusion models to network traffic remains in its infancy, and existing efforts have merely performed a straightforward model transfer by converting traffic data into images. Moreover, existing traffic classification models \cite{ET-BERT-TLS120,YaTC,NetMamba,TrafficFormer,Lens,Pat,SHAPE,MIETT,SNAKE} are typically designed with a deep coupling among core components such as the encoder, the pre-trained backbone, and the classification head. When transitioning across tasks, extensive architectural redesign and model retraining are often required, incurring prohibitively high development and tuning costs \cite{GGFAST}. This ``one-task-one-model'' development paradigm impedes the iterative evolution of traffic classification systems and hinders their capacity to adapt to rapidly evolving cybersecurity threats \cite{MEMENTO}.

This section systematically delineates major challenges confronting contemporary traffic classification research: (1) the trap of shortcut learning and spurious correlations under the supervised learning paradigm; (2) the class imbalance of real-world network traffic and the limitations of existing data augmentation methods; (3) the scenario-binding nature of model architectures and deficiencies in cross-task adaptability.

In light of the aforementioned challenges, the core research question of our work is formulated as follows: {\itshape How can we construct an encrypted traffic classification framework that is capable of escaping spurious correlations, accommodating extreme data distributions, supporting rapid cross-scenario adaptation, and generating high-fidelity traffic samples?} To address this question, we draw inspiration from the cognitive mechanisms of the human brain and propose a decoupled memory and diffusion framework for generalizable encrypted traffic classification.

\section{Methodology Overview}

\begin{figure*}[htbp]
  \centering
  \includegraphics[width=\textwidth]{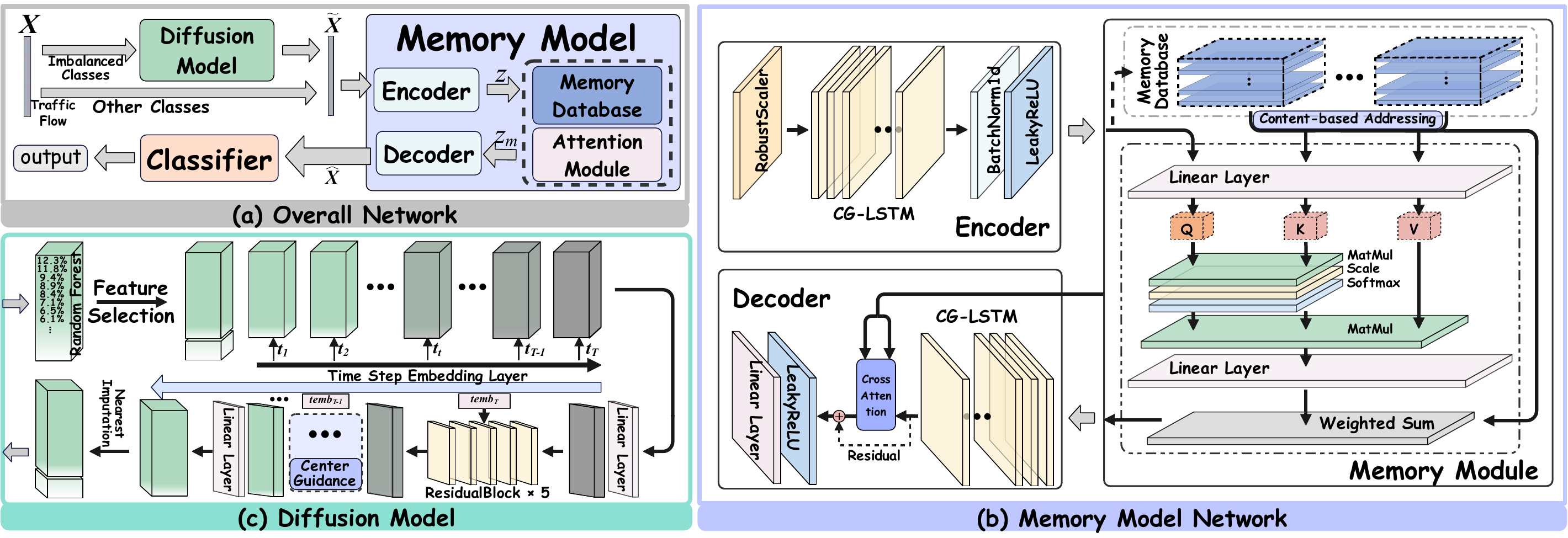}
  {
  \caption{Framework Overview.}
  \Description{(a) The proposed TDDM-Melatt framework consists of three core components: the memory representation model Melatt, the denoising diffusion-based traffic data augmentation module TDDM, and a downstream task classifier. (b) Architecture of the memory-augmented encoder-decoder. (c) The diffusion Model.}
  \label{fig:framework}
  }
\end{figure*}

The overall framework proposed in our work is illustrated in Figure~\ref{fig:framework}(a). It comprises three core components: a memory representation model termed Melatt, a traffic data augmentation module based on denoising diffusion probabilistic models termed TDDM, and a downstream task classifier. The traffic memory representation model, which serves as a pre-training architecture, consists of an encoder, a memory module, and a decoder. This model performs a sequence of operations on input traffic features, including encoding, memory bank addressing, and reconstruction decoding. The conditional diffusion augmentation module functions as an independent generative component capable of synthesizing high-fidelity minority class samples conditioned on specified class labels. The downstream task classifier is designed as a flexible and extensible classification engine. Embracing a modular design philosophy, it supports seamless transitions from lightweight traditional models to complex deep architectures, thereby accommodating diverse deployment scenarios (e.g., resource-constrained edge devices or high-performance servers). To highlight the decoupled nature of the framework, and unless otherwise specified, the classifier is implemented using a straightforward XGBoost \cite{XGBoost} model.

We construct an independent external memory module, the Melatt architecture illustrated in Figure~\ref{fig:framework}(b). Emulating the cognitive process of ``perception--memory--decision'' in the human brain, this module realizes the decoupling of feature extraction and knowledge storage. The module consists of multiple learnable memory matrices and a content-based addressing mechanism \cite{MemAE}. Each memory slot corresponds to the prototype features of a specific traffic class, supporting independent pre-training and flexible insertion or deletion. To overcome the prevalent ``fuzzy memory'' problem in neural networks, we introduce a negative entropy-based regularization constraint that enforces sharpening of the attention distribution. To address the dual nature of network traffic characterized by both periodicity and burstiness, we design a Competitive Gating Long Short-Term Memory (CG-LSTM) in both the encoder and the decoder. By placing its 3 gates within a shared softmax competitive framework, CG-LSTM forces the model to allocate a finite total “attentional budget” at each time step, thereby enhancing its discriminative capacity for dynamic traffic patterns. Furthermore, we integrate a multi-head cross-attention mechanism \cite{Mul-Att-Tra-1,MIETT,Mul-Att-Tra-3} within the decoder, which enables dynamic querying of multiple prototype patterns stored in the memory module at each decoding step, ultimately generating more accurate reconstructed sequences.

Traffic data is characterized by high dimensionality, sparsity, and strong feature interdependencies \cite{BARS}, while generic generative models often suffer from distribution drift and feature distortion when applied to this domain. To address these issues, we propose TDDM, a denoising diffusion model specifically tailored for structured traffic data. First, we design a feature selection method based on Random Forest Gini importance. This method prunes redundant features that exhibit high zero-value ratios or low classification contributions, achieving a dimensionality reduction of over 30\% while preserving 95\% of the critical information. Second, we eschew the complex convolutional and attention-based architectures commonly employed in image diffusion models. Instead, we adopt a lightweight MLP-based residual network as the noise predictor and adopt a class-wise training paradigm to avoid unnecessary cross-modal computational overhead. To further enhance the distributional consistency of generated samples, we propose a compact noise sampling mechanism. Finally, a nearest-neighbor feature padding strategy is employed to restore the complete feature dimensionality by leveraging the manifold consistency in the low-dimensional latent space.

\section{Methodological Details}

\subsection{Encoder and Decoder}

\textbf{Encoder Based on Competitive Gating LSTM.}

\begin{figure}[htbp]
  \centering
  \includegraphics[width=\linewidth]{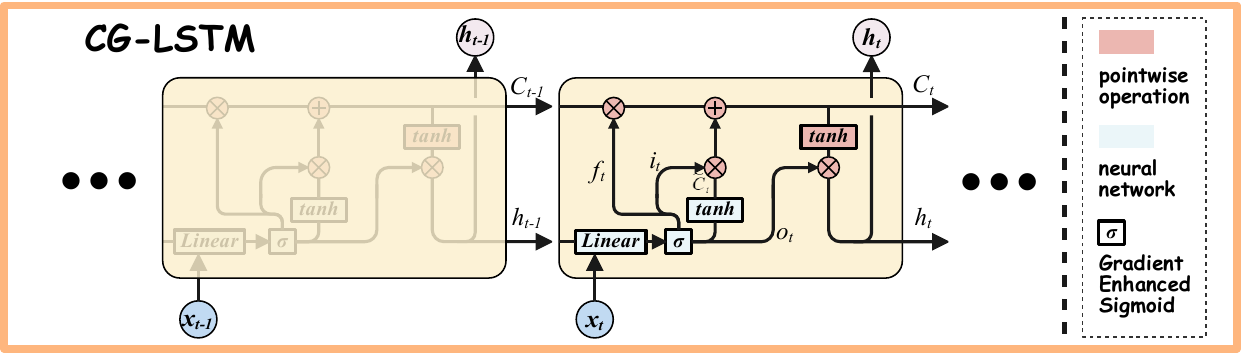}
  {
  \caption{Structure of CG-LSTM.}
  \Description{}
  \label{fig:CG-LSTM}
  }
\end{figure}

Traditional LSTM networks \cite{LSTM-2000,LSTM-2023} employ 3 independent sigmoidal gating mechanisms, allowing the forget gate, input gate, and output gate to simultaneously approach unity. This results in a lack of selectivity in the information flow through the memory cell, making it difficult to decisively switch between bursty traffic and quiescent background states. To address these challenges, we design a CG-LSTM, the structure of which is depicted in Figure~\ref{fig:CG-LSTM}.

In conventional LSTM networks \cite{LSTM-1997,LSTM-2000,LSTM-2023}, the forget gate \( f_t \), input gate \( i_t \), and output gate \( o_t \) are each computed via independent sigmoidal activation functions. Consequently, all 3 gating values may simultaneously approach unity, leading to scenarios where the model erases substantial portions of historical information while concurrently writing new content, thereby inducing memory confusion. The CG-LSTM replaces the independent sigmoidal activations with a joint softmax operation, thereby constraining 3 gates to form a probability distribution at each time step:
\begin{equation}
[f_t, i_t, o_t] = \text{softmax}\left(\boldsymbol{W}_g[\boldsymbol{h}_{t-1}; \boldsymbol{x}_t] + \boldsymbol{b}_g\right),
\label{eq:gates}
\end{equation}
where \( \boldsymbol{W}_g \in \mathbb{R}^{3 \times (d_h + d_x)} \), hidden state of previous step \( \boldsymbol{h}_{t-1} \), traffic vector \( \boldsymbol{x}_t \), and \( \boldsymbol{b}_g \in \mathbb{R}^3 \). This formulation enforces the following constraints:
\begin{equation}
f_t + i_t + o_t = 1, \quad f_t, i_t, o_t \geq 0.
\label{eq:constraint}
\end{equation}
Under this design, the model allocates a finite ``gating resource budget'' across the forget, input, and output operations at each time step. When encountering a traffic burst, the model may allocate the majority of its budget to the input gate \( i_t \) to rapidly incorporate novel features. During quiescent periods, the budget is preferentially assigned to the forget gate \( f_t \) to maintain the baseline representation of benign traffic. When the model needs to propagate long-range periodic patterns, the budget is concentrated on the output gate \( o_t \), enabling high-fidelity propagation of the memory state. Define the element-wise multiplication as $\odot$. The candidate memory cell state $\tilde{\boldsymbol{C}}_t = \tanh\left(\boldsymbol{W}_c[\boldsymbol{h}_{t-1}; \boldsymbol{x}_t] + \boldsymbol{b}_c\right)$. The update equations for the CG-LSTM cell state \( \boldsymbol{C}_t \) and hidden state \( \boldsymbol{h}_t \) are given by:
\begin{equation}
\boldsymbol{C}_t = f_t \odot \boldsymbol{C}_{t-1} + i_t \odot \tilde{\boldsymbol{C}}_t, \quad \boldsymbol{h}_t = o_t \odot \tanh(\boldsymbol{C}_t).
\label{eq:update}
\end{equation}

Compared with standard LSTM, CG-LSTM shares a single weight matrix across 3 gates, reducing the parameter count by approximately one-third. Furthermore, the extreme values (0 or 1) produced by the softmax output are more conducive to gradient propagation than the intermediate values typically output by sigmoidal functions, thereby alleviating the vanishing gradient problem in long sequences. When minority class traffic appears, the model is compelled to shift its gating budget from the forget gate toward the input gate in order to record novel features. This mechanism inherently amplifies the learning weight assigned to minority class samples, thereby enhancing robustness in class-imbalanced scenarios.

The encoder is constructed by stacking multiple CG-LSTM layers. For an input traffic feature sequence $\tilde{\boldsymbol{X}} = \{\boldsymbol{x}_1, \boldsymbol{x}_2, \dots, \boldsymbol{x}_T\}$ (\( T \) \text{ denotes the sequence length}), the encoder iteratively updates the cell state and hidden state at each time step according to Equation~\eqref{eq:update}. We adopt the hidden state \( \boldsymbol{h}_T \) from the final time step as the global temporal fingerprint of the traffic sequence. This fingerprint is subsequently transformed via a linear projection to yield a latent vector of dimension \( L \). A batch normalization layer and a LeakyReLU activation function are appended to the encoder output. Ultimately, the encoder produces a latent vector \( \boldsymbol{z}_t \in \mathbb{R}^L \), which serves dually as the initial state for the decoder and as the query vector for the subsequent memory module.

\textbf{Decoder Based on Cross-Attention.}

The decoder takes the memory vector \( \boldsymbol{z}_m \) output by the memory module as its initial state. It employs a multi-layer stack of CG-LSTM units as its backbone and incorporates a multi-head cross-attention mechanism \cite{Multi-Att-2017} to facilitate the fusion of global memory information with locally generated features. The overall architecture of the decoder comprises a CG-LSTM module, a multi-head cross-attention module, a residual connection, and layer normalization module and a fully connected reconstruction layer.

The CG-LSTM structure within the decoder is identical to that of the encoder. The multi-head cross-attention module serves as the core enhancement component of the decoder. By dynamically retrieving canonical traffic patterns from the global memory matrix, it mitigates the information closure problem. In this module, the intermediate feature sequence output by the CG-LSTM serves as the Query, representing the local temporal characteristics at the current generation step. The external memory matrix \( \boldsymbol{M} \in \mathbb{R}^{N \times L} \) serves simultaneously as both the Key and the Value, where \( N \) denotes the number of memory slots and \( L \) denotes the dimension of each memory slot. A multi-head attention mechanism \cite{Mul-Att-Tra-1,MIETT,Mul-Att-Tra-3} is employed to concurrently retrieve the most pertinent global prototype patterns from multiple distinct feature subspaces conditioned on the current local features:
\begin{equation}
\text{MultiHead}(\boldsymbol{Q}, \boldsymbol{M}, \boldsymbol{M}) = \text{Concat}(\text{head}_1, \dots, \text{head}_H)\boldsymbol{W}^O,
\label{eq:multihead}
\end{equation}
\begin{equation}
\text{head}_i = \text{Attention}(\boldsymbol{Q}\boldsymbol{W}_i^Q, \boldsymbol{M}\boldsymbol{W}_i^K, \boldsymbol{M}\boldsymbol{W}_i^V).
\label{eq:head}
\end{equation}
where \( \boldsymbol{W}^O \) is the learnable projection matrix for the output.
\begin{equation}
\text{Attention}(\boldsymbol{Q}, \boldsymbol{K}, \boldsymbol{V}) = \text{softmax}\left(\frac{\boldsymbol{Q}\boldsymbol{K}^T}{\sqrt{d_k}}\right)\boldsymbol{V}.
\label{eq:attention}
\end{equation}
The outputs of multiple attention heads are concatenated and subsequently transformed via a linear projection to yield the attention-augmented feature sequence. To alleviate the vanishing gradient problem during the training of deep neural networks, residual connections and layer normalization are introduced after the multi-head cross-attention module. These mechanisms fuse the original CG-LSTM output with the attention-enhanced features. Finally, a fully connected layer maps the fused feature sequence back to the original \( D \)-dimensional feature space of the input, producing the final reconstructed traffic sequence.

\subsection{Memory Model and Pre-training Constraints}

\textbf{Memory Module.}

The Melatt emulates the cognitive process of ``perception--memory--decision'' in the human brain by introducing an independent external memory unit, thereby realizing the decoupling of feature extraction and knowledge storage. This liberates the encoder from the burdensome task of parameter storage, allowing it to concentrate exclusively on extracting temporal features, while long-term knowledge retention is delegated to the external memory module.

The memory module constitutes the core component of Melatt. It consists of multiple learnable memory matrices \( \boldsymbol{M} \in \mathbb{R}^{N \times L} \) and a content-based addressing mechanism. Each \( \boldsymbol{M} \)  comprises \( N \) memory slots, each corresponding to a prototype feature of a target traffic class. These memory slots are iteratively optimized and updated during the pre-training process. Notably, each memory matrix can be pre-trained independently, and different pre-trained memory matrices can be flexibly inserted into or removed from the memory bank, provided that their dimensionality remains consistent.

Given the latent vector \( \boldsymbol{z}_i \) output by the encoder, the memory module computes its similarity with all memory slots via a content-based addressing mechanism. The relevance score \( \boldsymbol{e} \) for the memory slots is expressed as \( \boldsymbol{e} = {\boldsymbol{z}_i \cdot \boldsymbol{M}^T} / {\sqrt{L}} \).
The attention weight \( \omega = \text{Softmax}(\boldsymbol{e}) \) is computed via the softmax function, where the weight represents the ``memory strength'' of the input traffic with respect to each memory prototype. To address the prevalent ``fuzzy memory'' problem in neural networks (requiring that a given traffic flow correspond unambiguously to a single distinct memory prototype), we introduce a regularization constraint based on negative entropy. During training, the entropy of the attention distribution is minimized:
\begin{equation}
\mathcal{L}_{\text{ent}} = -\sum \omega_i \log \omega_i.
\label{eq:entropy_loss}
\end{equation}
This constraint enforces the attention weight distribution to become sharper and sparser, compelling the model to match a specific memory prototype with high precision during reconstruction. Ultimately, the model reconstructs the sequence using the attention-weighted memory content \( \boldsymbol{z}_m = \omega \boldsymbol{M} \), yielding the encoder output.

\textbf{Loss Function.}

The efficacy of Melatt is highly contingent upon the constraints imposed during the pre-training phase. A conventional, singular reconstruction error criterion is insufficient to guarantee that the latent space acquires structured semantic information. To this end, we devise 4 pre-training constraints.

\textbf{Reconstruction Loss.} We employ the Smooth L1 Loss as the reconstruction loss to mitigate the influence of outliers on gradient updates. Compared with the traditional MSE loss, Smooth L1 exhibits L1 robustness in large error regions, making it less susceptible to domination by outliers, and smoothly transitions to L2 loss in small error regions, thereby ensuring convergence stability.
\begin{equation}
\mathcal{L}_{\text{rec}} = \frac{1}{N} \sum_{i=1}^{N} \text{SmoothL1}(\tilde{\boldsymbol{X}}_i, \tilde{\boldsymbol{X}}_{mi}),
\label{eq:reconstruction_loss}
\end{equation}
where \( \tilde{\boldsymbol{X}}_m \) denotes the traffic reconstructed by the decoder.

\textbf{Compactness Constraint.} To prevent the latent vector \( \boldsymbol{z}_i \) output by the encoder from deviating substantially from the memory readout \( \boldsymbol{z}_m \), we introduce a compactness constraint. This constraint enforces the feature vectors in the latent space to closely adhere to the memory prototypes, thereby enhancing the continuity of the feature manifold:
\begin{equation}
\mathcal{L}_{\text{comp}} = \mathbb{E} \|\boldsymbol{z}_i - \boldsymbol{z}_m\|_2^2.
\label{eq:compactness_loss}
\end{equation}
By minimizing the distance between the latent vector and the memory readout vector, the model is encouraged to encode inputs into a subspace consistent with the memory prototypes, thus preventing excessive dispersion of latent features.

\textbf{Entropy Regularization and Diversity Constraints.} Equation~\eqref{eq:entropy_loss} prescribes a negative entropy-based regularization constraint, \( \mathcal{L}_{\text{ent}} \), which compels the model to match a specific memory prototype with high precision during traffic reconstruction.

Concurrently, redundant prototype features may appear in the memory bank, where multiple memory slots learn similar or identical feature patterns, causing ``mode collapse''. To prevent this, we maximize the distance between any pair of distinct memory slots \( \boldsymbol{m}_i \) and \( \boldsymbol{m}_j \). This enforces each memory slot to learn a unique and distinguishable traffic pattern, thereby augmenting the capacity and richness of feature representations:
\begin{equation}
\mathcal{L}_{\text{div}} = -\frac{1}{N(N-1)} \sum_{i \neq j} \|\boldsymbol{m}_i - \boldsymbol{m}_j\|_2^2.
\label{eq:diversity_loss}
\end{equation}
The negative sign ensures that minimizing \( \mathcal{L}_{\text{div}} \) is equivalent to maximizing the pairwise distances among memory slots.

\textbf{Total Loss.} The aggregate loss function during the pre-training phase is formulated as a weighted sum of the losses defined in Equations~\eqref{eq:entropy_loss}, ~\eqref{eq:reconstruction_loss}, \eqref{eq:compactness_loss}, and \eqref{eq:diversity_loss}. In practice, while the reconstruction loss retains its original scale (weight of 1), we set the weighting coefficients as \( \alpha_{\text{comp}} = 0.05 \), \( \alpha_{\text{ent}} = 0.05 \), and \( \alpha_{\text{div}} = 0.0005 \). This configuration yielded favorable outcomes across all our experiments.

It is important to emphasize that the nearest-prototype matching paradigm, driven by \(\mathcal{L}_{\mathrm{rec}}\), is the pre-training objective. During inference, the decoder outputs the reconstructed feature vector \(\tilde{\mathbf{X}}_{mi} \in \mathbb{R}^{D}\), which is concatenated with the encoder's latent vector \(\mathbf{z}_i \in \mathbb{R}^{L}\) to form a compact representation \([\tilde{\mathbf{X}}_{mi}; \mathbf{z}_i]\). This representation is provided to downstream classifier for final classification. The pre-training process ensures that \(\tilde{\mathbf{X}}_{mi}\) inherently encodes the quality of prototype matching, making the inference pipeline efficient and decoupled from the memory module's per-prototype computational overhead.

\subsection{Feature Selection and Padding}
\label{4.3}

\textbf{Preprocessed Flow-Level Statistical Features.}
We extract a total of 76 flow-level statistical features from each bidirectional flow, all derived exclusively from packet headers (IP and TCP/UDP layers) without accessing any encrypted payload. These features span five functional categories: length/size statistics, temporal and IAT features, packet/byte count features, rate features, and flag/initialization features. The complete list of all 76 features, organized by category, is provided in Appendix Table~\ref{tab:full-feature-list}.

\textbf{Feature Selection Based on Gini Importance.}

Traffic data differs fundamentally from image data \cite{BARS}, the domain in which diffusion models have been most extensively applied. A substantial proportion of traffic features assume zero values, as summarized in Table~\ref{tab:zero_ratio}. The table presents, for each of the four datasets, the fraction of features whose zero proportion exceeds the threshold \( \gamma \). In each dataset, over 10\% of the features are identically zero across all samples. A higher zero-value ratio for a given feature indicates a greater prevalence of zero entries across all classes within the dataset. This phenomenon may correspond to one of two scenarios: (1) the feature constitutes a distinguishing characteristic of a specific traffic class relative to others; (2) the feature exhibits consistent behavior across different classes, manifesting as a uniformly high zero-value ratio. In the latter case, the feature conveys negligible information, contributes virtually nothing to downstream tasks, and simultaneously degrades the training efficiency of the diffusion model while increasing the risk of overfitting.

\begin{table}[htbp]
\centering
\small
{
\renewcommand{\arraystretch}{0.85}
\caption{Statistics of Feature Zero-Value Ratios}
\label{tab:zero_ratio}
\begin{tabular*}{\columnwidth}{@{\extracolsep{\fill}} lcccc}
\toprule
Dataset & \( \gamma = 100\% \) & \( \gamma > 95\% \) & \( \gamma > 90\% \) & \( \gamma > 80\% \) \\
\midrule
CIC-IDS-2017   & 10.39\% & 19.48\% & 23.38\% & 27.27\% \\
ISCX-VPN-2016  & 11.84\% & 27.63\% & 46.05\% & 64.47\% \\
USTC-TFC2016   & 11.84\% & 18.42\% & 21.05\% & 38.16\% \\
CSTNET-TLS1.3  & 10.53\% & 15.79\% & 19.74\% & 21.05\% \\
\bottomrule
\end{tabular*}
}
\end{table}

To address this issue, we introduce a feature selection method based on Random Forest importance scores. Specifically, a Random Forest classifier is first trained on the training partition designated for the diffusion model. The Gini Importance \cite{Gini} of each feature is subsequently computed, which quantifies the contribution of a feature to classification performance via the mean reduction in node impurity. The Gini impurity at a given node is defined as: $\text{Gini}(i) = 1 - \sum_{k} p_k^2,$ where \( p_k \) denotes the proportion of samples belonging to the \( k \)-th class at forest node \( i \). For a node with \( n_i \) total samples, the reduction in Gini impurity resulting from a split is given by:
\begin{equation}
\Delta \text{Gini} = \text{Gini}(i) - \frac{n_{\text{left}}}{n_i} \text{Gini}(i_{\text{left}}) - \frac{n_{\text{right}}}{n_i} \text{Gini}(i_{\text{right}}).
\label{eq:gini_reduction}
\end{equation}
After normalization, the Gini Importance for each feature is obtained. Features are then ranked in descending order of their importance. We retain the minimal subset of features whose cumulative importance reaches a predefined threshold, while the remaining features are deemed low-contribution and are consequently pruned.

\begin{figure}[htbp]
\centering
{
\includegraphics[width=0.8\columnwidth]{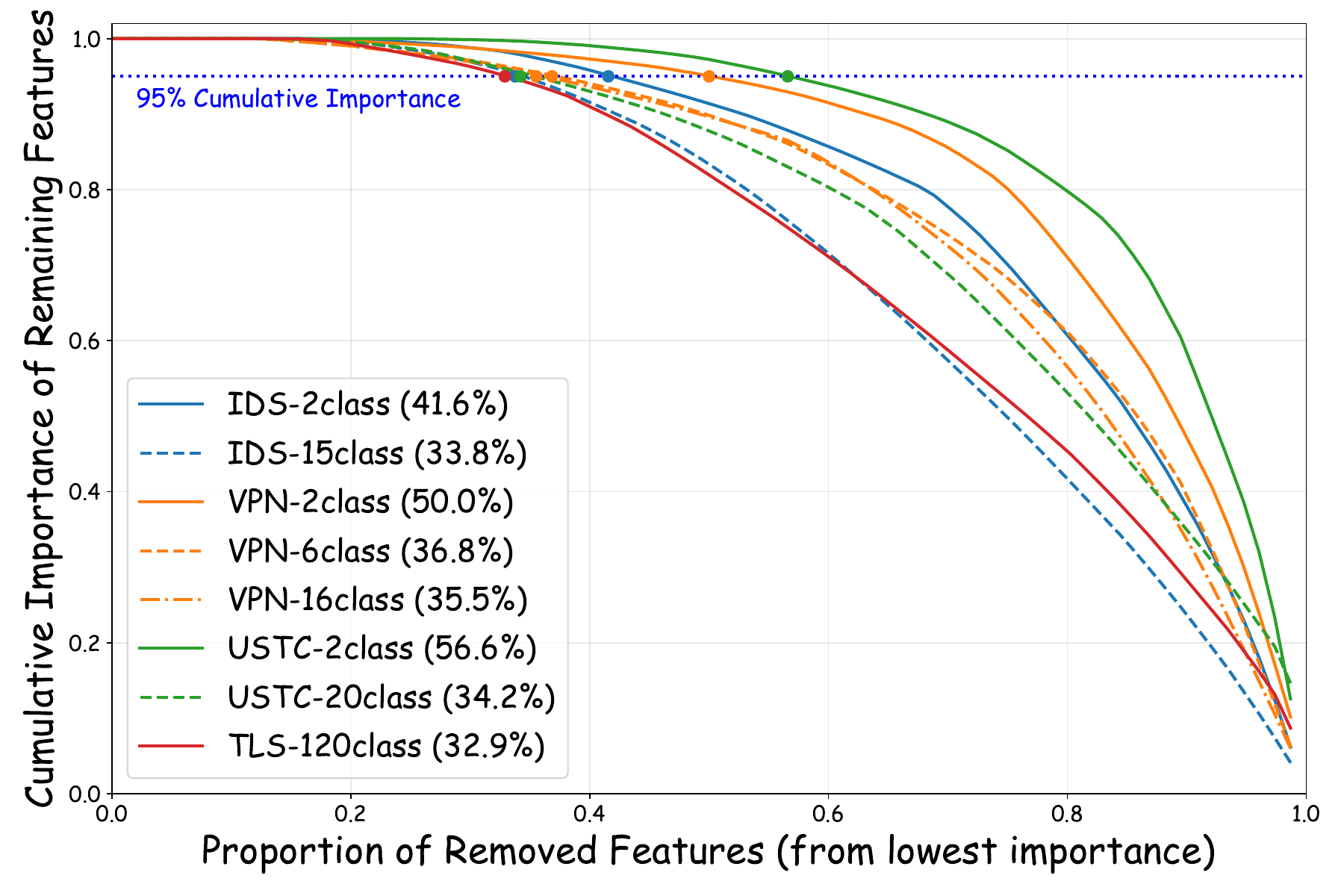}
\caption{Feature Importance Cumulative Curves.}
\Description{Feature Importance Cumulative Curves for Different Datasets and Tasks. The legend entries correspond to: IDS-2class (41.6\%), IDS-15class (33.8\%), VPN-2class (50.0\%), VPN-6class (36.8\%), VPN-16class (35.5\%), USTC-2class (56.6\%), USTC-20class (34.2\%), and TLS-120class (32.9\%).}
\label{fig:feature_importance}
}
\end{figure}

As illustrated in Figure~\ref{fig:feature_importance}, experimental results across 8 tasks spanning 4 datasets demonstrate that setting the cumulative importance threshold to 95\% yields a reduction in feature dimension exceeding 30\%. We present the feature importance bar charts for 4 tasks in Appendix Figure~\ref{fig:bar_all_remaining}.

Features that are pruned are not simply discarded. Instead, after the diffusion model synthesizes samples in the reduced feature space, we restore the full feature dimension using a nearest-neighbor padding strategy. 

\textbf{Feature Padding Based on Nearest-Neighbor.}

After feature selection, the diffusion model is trained and generates samples exclusively within the important feature subspace \( \mathbb{R}^d \). However, downstream classification tasks require the complete \( D \)-dimensional feature vectors. Directly discarding the pruned features leads to inevitable information loss, whereas naive random padding disrupts the intrinsic correlations among features, thereby inducing uncontrolled deviation between the distribution of generated samples and that of real data. To address this issue, we propose a nearest-neighbor-based padding strategy. The key insight is that the manifold structure of feature vectors in the original high-dimensional space is approximately preserved under the low-dimensional projection. Consequently, if two samples are proximate in the low-dimensional space, they should also exhibit similar values for the pruned features in the full-dimensional space.

Let \( \boldsymbol{X}_c^{\text{full}} \in \mathbb{R}^{n_c \times D} \) denote the matrix of authentic data belonging to class \( c \). Its low-dimensional projection is given by: $\boldsymbol{X}_c^{\text{low}} = \boldsymbol{X}_c^{\text{full}}[:, \mathcal{K}] \in \mathbb{R}^{n_c \times d},$ where \( \mathcal{K} \) denotes the index set of retained features. For a generated low-dimensional sample \( \tilde{\boldsymbol{y}} \in \mathbb{R}^d \), we define the index of its nearest neighbor as:
\begin{equation}
i^* = \arg \min_{i \in \{1, \dots, n_c\}} \left\| \tilde{\boldsymbol{y}} - \boldsymbol{X}_c^{\text{low}}[i, :] \right\|_2.
\label{eq:nn_index}
\end{equation}
The complete generated sample \( \tilde{\boldsymbol{\boldsymbol{x}}} \in \mathbb{R}^D \) is then constructed as: $\tilde{\boldsymbol{x}}[\mathcal{K}] = \tilde{\boldsymbol{y}}, \quad \tilde{\boldsymbol{x}}[\mathcal{D}] = \boldsymbol{X}_c^{\text{full}}[i^*, \mathcal{D}],$ where \( \mathcal{D} \) denotes the index set of pruned features.

\subsection{Traffic Feature Generation via Diffusion}

\begin{figure}[htbp]
\centering
{
\includegraphics[width=\columnwidth]{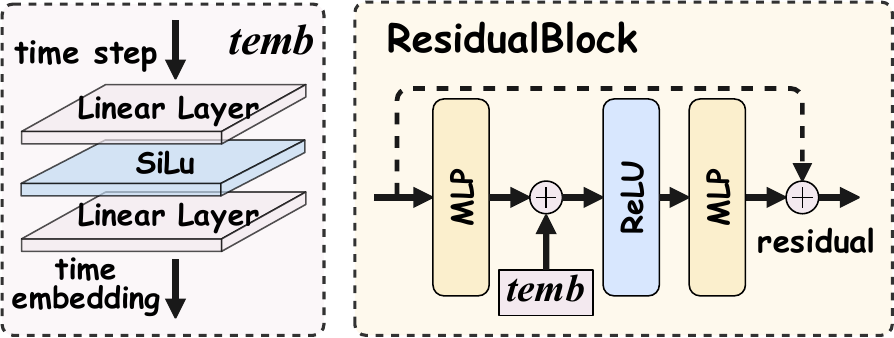}
\caption{Details of TDDM Architecture.}
\Description{The diagram depicts the feature selection stage, the time step embedding layer, and the stacked residual blocks, each comprising an MLP, ReLU activation, and a second MLP.}
\label{fig:TDDM}
}
\end{figure}

The overall architecture of our proposed TDDM, is built upon the denoising diffusion probabilistic framework. It comprises two core modules: the forward noise-adding diffusion process and the reverse denoising network \cite{DDPM,DDIM}. TDDM is specifically tailored to accommodate the characteristics of traffic feature data, thereby addressing the generative requirements of non-image, structured traffic data. The detailed structure is illustrated in Figure~\ref{fig:framework}(c) and Figure~\ref{fig:TDDM}.

Traffic feature data differs intrinsically from image data. In the image domain, an encoder is typically employed to map pixel-space representations into a lower-dimensional latent space to reduce the computational burden of diffusion models. In contrast, traffic features are inherently structured and comparatively low in volume; thus, a similar dimensionality reduction mapping is unnecessary, and diffusion modeling can be performed directly in the original feature space \cite{traffic-img}. Given a feature sample \( \boldsymbol{x}_0 \sim q(\boldsymbol{x}_0) \) belonging to class \( c \), the forward diffusion process progressively injects Gaussian noise over \( T \) steps, yielding a sequence of increasingly noisy latent variables. The transition probability at each step is defined as:
\begin{equation}
q(\boldsymbol{x}_t \mid \boldsymbol{x}_{t-1}) = \mathcal{N}(\boldsymbol{x}_t; \sqrt{1 - \beta_t}\, \boldsymbol{x}_{t-1}, \beta_t \mathbf{I}),
\label{eq:forward_transition}
\end{equation}
where \( \beta_t \in (0,1) \) represents a predefined noise schedule. We adopt a linear schedule defined by $\beta_t = \beta_{\text{start}} + \frac{t-1}{T-1} (\beta_{\text{end}} - \beta_{\text{start}})$. In our experiments, we set \( \beta_{\text{start}} = 10^{-4} \) and \( \beta_{\text{end}} = 0.02 \). Leveraging the reparameterization trick, the noisy sample at an arbitrary timestep \( t \) can be directly expressed as: $\boldsymbol{x}_t = \sqrt{\bar{\alpha}_t}\, \boldsymbol{x}_0 + \sqrt{1 - \bar{\alpha}_t}\, \boldsymbol{\varepsilon}$, where \( \boldsymbol{\varepsilon} \sim \mathcal{N}(0, \mathbf{I}) \), \( \alpha_t = 1 - \beta_t \), and \( \bar{\alpha}_t = \prod_{i=1}^t \alpha_i \).

We eschew complex convolutional and attention-based architectures for the noise prediction network and instead adopt a lightweight structure for noise estimation and denoising. The network takes as input the current noisy sample \( \boldsymbol{x}_t \in \mathbb{R}^d \) and a time-step embedding \( temb \in \mathbb{R}^{128} \) produced by a two-layer linear projection. The main network body consists of an input MLP layer (\( d \to 512 \)), five cascaded residual blocks, and an output MLP layer (\( 512 \to d \)). Each residual block contains two fully connected layers and integrates the time-step embedding and the residual path via a linear layer. The diffusion model is trained according to Algorithm~\ref{alg:diffusion_training}, minimizing the following simplified loss function:
\begin{equation}
\mathcal{L}_{\text{simple}} = \mathbb{E}_{t, \boldsymbol{x}_0, \boldsymbol{\varepsilon}} \left[ \left\| \boldsymbol{\varepsilon} - \boldsymbol{\varepsilon}_\theta(\boldsymbol{x}_t, t) \right\|^2 \right].
\label{eq:diffusion_loss}
\end{equation}

\newcommand{\algorithmicinput}{\textbf{Input:}}
\newcommand{\algorithmicoutput}{\textbf{Output:}}
\newcommand{\Input}{\item[\algorithmicinput]}
\newcommand{\Output}{\item[\algorithmicoutput]}
\begin{algorithm}[htbp]
\caption{Class-Wise Diffusion Model Training}
\label{alg:diffusion_training}
\begin{algorithmic}[1]
\Input Low-dimensional features \( \boldsymbol{X}_c \) of class \( c \), number of diffusion steps \( T \), learning rate \( \eta \), batch size \( B \), patience \( p \)
\Output Trained noise prediction network \( \boldsymbol{\varepsilon}_\theta \)
\State Initialize network parameters \( \theta \)
\Repeat
    \State Sample a mini-batch \( \{\boldsymbol{x}_0^{(i)}\}_{i=1}^B \) randomly from \( \boldsymbol{X}_c \)
    \For{each sample \( i \) in the batch}
        \State Sample timestep \( t \sim \text{Uniform}(\{1, \dots, T\}) \)
        \State Sample noise \( \boldsymbol{\varepsilon}^{(i)} \sim \mathcal{N}(0, \mathbf{I}) \)
        \State Compute \( \boldsymbol{x}_t^{(i)} = \sqrt{\bar{\alpha}_t}\, \boldsymbol{x}_0^{(i)} + \sqrt{1 - \bar{\alpha}_t}\, \boldsymbol{\varepsilon}^{(i)} \)
    \EndFor
    \State Compute loss \( \mathcal{L}_{\text{simple}} = \frac{1}{B} \sum_{i=1}^B \left\| \boldsymbol{\varepsilon}^{(i)} - \boldsymbol{\varepsilon}_\theta(\boldsymbol{x}_t^{(i)}, t) \right\|^2 \)
    \State Update \( \theta \) using gradient descent
\Until{validation loss does not decrease for \( p \) consecutive epochs}
\State \Return \( \boldsymbol{\varepsilon}_\theta \)
\end{algorithmic}
\end{algorithm}

\subsection{Condition-Guided Compact Noise Sampling}

In Algorithm~\ref{alg:diffusion_training}, we adopt a class-wise training paradigm for the noise prediction network rather than a joint training scheme. Classical text-guided diffusion models require the introduction of cross-modal cross-attention mechanisms and rely on pre-trained text encoders to establish alignment between textual descriptions and generated content. In contrast, the diffusion model oriented toward traffic generation only utilizes class labels as conditional guidance. The limited semantic information inherent in class labels is insufficient to support effective training of a text encoder, while simultaneously leading to inefficient utilization of computational resources.

Nevertheless, a diffusion generative model that lacks conditional guidance, even when trained independently per class, still cannot avoid the drift of synthetic data away from the real distribution. To address this issue, we design a centroid-guided compact sampling mechanism during the denoising phase to enhance the distributional consistency of the generated samples.

The core idea of the compact sampling mechanism is to guide the samples toward the authentic data distribution of their corresponding class at each reverse denoising step. Since feature selection has already been performed during the forward diffusion process, and the compact sampling directly operates on the predicted noise, this method exerts only a minor influence on the diversity of the generated data. Let the feature centroid of class \( c \) be defined as: $\boldsymbol{\mu}_c = \frac{1}{n_c} \sum_{i=1}^{n_c} \boldsymbol{x}_i^{(c)}$. During sampling, starting from \( \boldsymbol{x}_T \sim \mathcal{N}(0, \mathbf{I}) \) and proceeding in reverse temporal order, the noise prediction network first estimates the current noise \( \hat{\boldsymbol{\varepsilon}} = \boldsymbol{\varepsilon}_\theta(\boldsymbol{x}_t, t) \). Subsequently, a guidance correction term is introduced:
\begin{equation}
\hat{\boldsymbol{\varepsilon}}' = \hat{\boldsymbol{\varepsilon}} - \gamma \cdot (1 - \bar{\alpha}_t)^{\frac{1}{2}} (\boldsymbol{x}_t - \boldsymbol{\mu}_c),
\label{eq:noise_correction}
\end{equation}
where \( \gamma \geq 0 \) is a hyperparameter that controls the guidance strength (we set \( \gamma = 0.3 \) in our experiments). The corrected noise prediction steers the update direction toward the real sample distribution, thereby yielding more compactly distributed samples. The sample update formula is given by:
\begin{equation}
\boldsymbol{x}_{t-1} = \frac{1}{\sqrt{\alpha_t}} \left( \boldsymbol{x}_t - \frac{\beta_t}{\sqrt{1-\bar{\alpha}_t}} \hat{\boldsymbol{\varepsilon}}' \right) + \sqrt{\beta_t}\, \boldsymbol{u},
\label{eq:sampling_update}
\end{equation}
where \( \boldsymbol{u} \sim \mathcal{N}(0, \mathbf{I}) \) is standard Gaussian noise. When \( \boldsymbol{x}_t \) deviates from the centroid \( \boldsymbol{\mu}_c \), the term \( (1-\bar{\alpha}_t)^{1/2} \) acts as a noise-level-dependent weight: during the early stages of denoising (large \( t \), strong noise), the guidance is relatively weak to avoid disrupting the global structure; during the later stages (small \( t \), weak noise), the guidance intensifies and forces the sample to converge toward the authentic class distribution.

\section{Evaluation}

\subsection{Experimental Setup}
\label{5.1}

\textbf{Experimental Environment.} All experiments were conducted on a server equipped with an AMD EPYC 7343 CPU @ 3.2\,GHz, 256\,GB of RAM, and an NVIDIA GeForce RTX 4090 GPU with 24\,GB of memory. The software stack comprised Python~3.9 and the deep learning framework PyTorch~2.8.0.

\textbf{Datasets.} Our evaluation employs four representative public benchmark datasets. Each dataset is partitioned into training, validation, and test sets following a 7:1:2 ratio.
\begin{itemize}
\item \textbf{CIC-IDS-2017} (denoted IDS) \cite{IDS}. Collected under emulated real-world network conditions, this dataset contains benign traffic alongside 14 categories of attack traffic. It supports binary classification (2-class) and 15-class classification tasks.
\item \textbf{ISCX-VPN-2016} (denoted VPN) \cite{VPN}. This dataset focuses on encrypted traffic classification under both VPN and non-VPN environments. It supports binary classification (VPN vs.\ non-VPN), 6-class classification (service types), and 16-class classification (specific application types).
\item \textbf{USTC-TFC2016} (denoted USTC) \cite{USTC}. This dataset is dedicated to Tor anonymous network traffic classification. It supports binary classification (Tor vs.\ non-Tor) and 20-class classification.
\item \textbf{CSTNET-TLS1.3} (denoted TLS) \cite{ET-BERT-TLS120}. This dataset is dedicated to the analysis of TLS~1.3 encrypted traffic. It encompasses traffic from 120 mainstream websites and supports a 120-class classification task.
\end{itemize}

\textbf{Data Preprocessing.} Our experiments adopt strict flow-level partitioning, using flow-level statistical features as the fundamental unit of analysis. The specific preprocessing pipeline is as follows:
\begin{itemize}
\item Parsing of raw traffic files is performed such that each flow sample retains only statistical features derived from transport-layer interactions.
\item Rigorous topological information anonymization is enforced. Identifiers that directly locate network entities---including flow IDs, IP, and Port---are removed.
\item Timestamp generalization is applied. 
\item Missing values arising from statistical features are uniformly filled with zero to maintain tensor integrity.
\end{itemize}

{\itshape It is worth emphasizing that the constraints described above apply exclusively to our own model. Unless otherwise specified, the baseline models described subsequently adopt the data preparation procedures recommended in their original publications.}

\textbf{Baseline Models.} To comprehensively validate the superiority of the proposed TDDM-Melatt framework in encrypted traffic classification scenarios, we select SOTA methods as comparative baselines.
\begin{itemize}
\item \textbf{Classification Baselines.} We compare our model against SOTA traffic classification models, including ET-BERT \cite{ET-BERT-TLS120}, YaTC \cite{YaTC}, NetMamba \cite{NetMamba}, TrafficFormer \cite{TrafficFormer}, netFound \cite{netFound}, and Pcap-Encoder \cite{Pcap-Encoder}. {\itshape All baseline models adhere as closely as possible to the data preparation procedures and hyperparameter settings recommended in their original publications. Our requirement is strict flow-level partitioning. For pre-trained models, we freeze the backbone parameters and fine-tune only the downstream classification head. Furthermore, we require that all models support recognition of classes containing only a small number of samples; discarding any class due to insufficient sample size is forbidden, as artificially modifying the original data distribution introduces bias that may adversely affect real-world deployment performance.}
\item \textbf{Data Generation Baselines.} We compare the proposed TDDM against current SOTA traffic generation models, including the diffusion-based models NetDiffus \cite{Netdiffus} and LRT-DDPM \cite{LRT-DDPM}, as well as GANs \cite{GAN} and their variants ACGAN \cite{ACGAN} and WGAN-GP \cite{WGAN-GP}. {\itshape All data generation procedures are applied exclusively to the training set, while the test set remains unaltered in its original state.}
\end{itemize}

\textbf{Evaluation Metrics.} We adopt Accuracy (AC), Precision (PR), and F1-Score (F1) as the primary evaluation metrics. In addition to classification metrics, we adopt Maximum Mean Discrepancy (MMD) and Average Pairwise Distance (APD) as complementary metrics to directly evaluate the quality of generated samples.

\subsection{Analysis of Classification Performance}
\label{5.2}

To evaluate the generalization capability and feature representation quality of the TDDM-Melatt across diverse network traffic classification scenarios, we construct 6 classification tasks spanning three datasets: ISCX-VPN-2016, USTC-TFC2016, and CSTNET-TLS1.3. These tasks cover varying levels of difficulty, including binary classification, multi-class classification, and fine-grained classification. 6 baseline classifiers and 6 SOTA representation models are selected for comparative validation. Given the sufficient sample sizes in binary classification tasks, TDDM is not applied to those tasks. For the remaining classification tasks, a class is defined as a "long-tail" (minority) class if its sample size is less than 15\% of that of the largest class. For such classes, TDDM is employed to augment their training set by a factor of 1.5$\times$. The full set of experimental results is summarized in Table~\ref{tab:classification_comparison}.

\begin{table*}[htbp]
\centering
\small
{
\caption{Classification Performance Comparison of TDDM-Melatt against Baseline and SOTA Models across 6 Tasks.}
\label{tab:classification_comparison}
\begin{tabular*}{\textwidth}{@{\extracolsep{\fill}} l|cc|cc|cc|cc|cc|cc@{}}
\toprule
\multirow{2}{*}{Model} & \multicolumn{2}{c|}{VPN-2class} & \multicolumn{2}{c|}{VPN-6class} & \multicolumn{2}{c|}{VPN-16class} & \multicolumn{2}{c|}{USTC-2class} & \multicolumn{2}{c|}{USTC-20class} & \multicolumn{2}{c}{TLS-120class} \\
\cmidrule(lr){2-3} \cmidrule(lr){4-5} \cmidrule(lr){6-7} \cmidrule(lr){8-9} \cmidrule(lr){10-11} \cmidrule(lr){12-13}
& AC & F1 & AC & F1 & AC & F1 & AC & F1 & AC & F1 & AC & F1 \\
\midrule
MLP \cite{MLP}       & 0.925 & 0.798 & 0.657 & 0.140 & 0.309 & 0.032 & 0.966 & 0.966 & 0.154 & 0.030 & 0.008 & 0.000 \\
1D-CNN \cite{1D-CNN}    & 0.876 & 0.558 & 0.683 & 0.136 & 0.311 & 0.030 & 0.915 & 0.915 & 0.136 & 0.021 & 0.005 & 0.000 \\
2D-CNN  \cite{2D-CNN}   & 0.918 & 0.804 & 0.681 & 0.140 & 0.308 & 0.035 & 0.986 & 0.986 & 0.062 & 0.033 & 0.014 & 0.000 \\
LSTM \cite{LSTM}      & 0.955 & 0.895 & 0.540 & 0.364 & 0.337 & 0.277 & \textbf{1.000} & \textbf{1.000} & 0.854 & 0.844 & 0.574 & 0.541 \\
GRU  \cite{GRU}      & 0.959 & 0.902 & 0.643 & 0.371 & 0.354 & 0.296 & \textbf{1.000} & \textbf{1.000} & 0.852 & 0.843 & 0.579 & 0.545 \\
Transformer \cite{Transformer} & 0.972 & 0.924 & 0.579 & 0.226 & 0.355 & 0.089 & 0.992 & 0.992 & 0.395 & 0.336 & 0.258 & 0.210 \\
netFound \cite{netFound}  & 0.760 & 0.619 & 0.473 & 0.365 & 0.329 & 0.153 & 0.994 & 0.994 & 0.580 & 0.307 & 0.019 & 0.005 \\
YaTC  \cite{YaTC}     & 0.839 & 0.839 & 0.692 & 0.601 & 0.609 & 0.443 & 0.995 & 0.995 & 0.852 & 0.780 & 0.155 & 0.096 \\
NetMamba  \cite{NetMamba} & 0.750 & 0.745 & 0.569 & 0.490 & 0.396 & 0.284 & 0.976 & 0.975 & 0.725 & 0.577 & 0.088 & 0.045 \\
ET-BERT  \cite{ET-BERT-TLS120}  & 0.847 & 0.846 & 0.717 & 0.642 & 0.592 & 0.437 & \textbf{1.000} & \textbf{1.000} & 0.849 & 0.796 & 0.109 & 0.067 \\
TrafficFormer \cite{TrafficFormer} & 0.909 & 0.909 & 0.765 & 0.694 & 0.677 & 0.544 & \textbf{1.000} & \textbf{1.000} & 0.720 & 0.650 & 0.297 & 0.240 \\
Pcap-Encoder \cite{Pcap-Encoder} & \textbf{0.999} & \textbf{0.999} & \textbf{0.921} & \textbf{0.898} & \textbf{0.835} & 0.710 & \textbf{1.000} & \textbf{1.000} & 0.910 & 0.871 & \textbf{0.710} & \underline{0.637} \\
\midrule
Melatt      & \textbf{0.999} & \textbf{0.999} & 0.838 & 0.833 & \underline{0.826} & \textbf{0.811} & \textbf{1.000} & \textbf{1.000} & \underline{0.966} & \textbf{0.966} & 0.586 & 0.581 \\
TDDM-Melatt & - & - & \underline{0.889} & \underline{0.889} & 0.824 & \underline{0.810} & - & - & \textbf{0.967} & \textbf{0.966} & \underline{0.684} & \textbf{0.681} \\
\bottomrule
\end{tabular*}
}
\par\smallskip\footnotesize\raggedright
Note: The best results are highlighted in \textbf{bold}, and the second-best results are indicated with \underline{underlining}. For the binary classification tasks (VPN-2class, USTC-2class), TDDM-Melatt does not employ data augmentation and its performance is identical to that of Melatt.
\end{table*}

The experimental results demonstrate that TDDM-Melatt achieves consistently superior performance across multi-scenario traffic classification tasks, exhibiting a significant performance advantage and strong generalization capability. The baseline classifiers, when unconstrained, perform reasonably on binary classification tasks but suffer precipitous performance drops on multi-class tasks of higher difficulty, revealing severely insufficient generalization. Although the representation learning models generally exhibit better generalization than the baseline classifiers, their performance still degrades substantially compared to the results originally reported in their respective publications. Notably, this performance attenuation stems from merely two constraints: (1) requiring that packets from the same flow not appear simultaneously in both the training and test sets; (2) freezing the encoder parameters of pre-trained models. In contrast, the memory-decoupled design of TDDM-Melatt enables the model to focus on learning the intrinsic behavioral characteristics of traffic rather than relying on spurious topological identifiers. As a result, it maintains stable, high performance under stringent experimental conditions---including topology anonymization and timestamp generalization.

\begin{table}[tbp]
\centering
\small
{
\caption{Performance under Data Generalization Scenarios.}
\label{tab:pcap_vs_melatt}
\begin{tabular}{l|l|c|c}
\toprule
Model & Processing & VPN-16class & TLS-120class \\
\midrule
\multirow{3}{*}{Pcap-Encoder} 
  & Retain (Original) & 0.710 & \underline{0.637} \\
  & Remove Payload    & 0.667 & 0.630 \\
  & Remove IPs/Ports  & 0.525 & 0.130 \\
\midrule
Melatt       &    -   & \textbf{0.811} & 0.581 \\
TDDM-Melatt  &    -   & \underline{0.810} & \textbf{0.681} \\
\bottomrule
\end{tabular}
}
\end{table}

As shown in Table~\ref{tab:classification_comparison}, Pcap-Encoder and TDDM-Melatt both achieve outstanding performance and appear comparable. To further simulate common real-world network scenarios involving traffic encryption and anonymization, we conduct a generalization robustness study in the context of multi-class classification tasks. We progressively apply the data preprocessing constraints adopted by TDDM-Melatt to the input of Pcap-Encoder. We first remove payload data to simulate scenarios where payload encryption renders payload inspection infeasible. Subsequently, we remove explicit identifiers such as IP and Port to emulate traffic classification under an anonymized network environment. Using the F1-score, which better reflects true performance under class imbalance, as the evaluation metric, the results are reported in Table~\ref{tab:pcap_vs_melatt}. As the level of data generalization increases, the detection performance of Pcap-Encoder exhibits a pronounced downward trend. In particular, after the removal of explicit identifiers, its F1-score on the TLS-120class task plummets from 0.637 to 0.130, ultimately degrading to a level comparable to the remaining baseline models. Under identical generalization scenarios, TDDM-Melatt consistently achieves the best performance, validating the superiority of its memory decoupling and diffusion-based data augmentation mechanisms in coping with the uncertainties intrinsic to real-world network environments.

\subsection{Generation Performance Analysis}
\label{5.3}

Examining the experimental results in Table~\ref{tab:classification_comparison}, it can be observed that, with the exception of the VPN-16class task, TDDM consistently improves the classification performance of Melatt across the remaining five tasks. To further validate the generation quality of TDDM, we conduct comparative experiments against NetDiffus, LRT-DDPM, and a suite of GAN-based models (GAN, WGAN-GP, ACGAN). A unified experimental setup is adopted for all comparisons: Melatt serves as the underlying pre-trained classification backbone, and the original training set is augmented exclusively with synthetically generated data. The experimental results are presented in Table~\ref{tab:tddm_vs_others}.

\begin{table*}[htbp]
\centering
{
\caption{Performance Comparison of Data Augmentation Methods across 4 Tasks.}
\label{tab:tddm_vs_others}
\small
\begin{tabular*}{\textwidth}{@{\extracolsep{\fill}} l|ccc|ccc|ccc|ccc@{}}
\toprule
\multirow{2}{*}{Model} & \multicolumn{3}{c|}{VPN-6class} & \multicolumn{3}{c|}{VPN-16class} & \multicolumn{3}{c|}{USTC-20class} & \multicolumn{3}{c}{TLS-120class} \\
\cmidrule(lr){2-4} \cmidrule(lr){5-7} \cmidrule(lr){8-10} \cmidrule(lr){11-13}
& PR & AC & F1 & PR & AC & F1 & PR & AC & F1 & PR & AC & F1 \\
\midrule
Melatt         & \underline{0.8502} & 0.8381 & 0.8329 & \underline{0.8282} & \textbf{0.8258} & \textbf{0.8113} & 0.9677 & 0.9664 & 0.9656 & \underline{0.5855} & \underline{0.5863} & \underline{0.5810} \\
\midrule
Melatt + GAN  \cite{GAN}            & 0.8490 & \underline{0.8393} & 0.8331 & 0.8281 & \underline{0.8256} & 0.8107 & 0.9687 & 0.9670 & 0.9661 & 0.5753 & 0.5742 & 0.5698 \\
Melatt + WGAN-GP \cite{WGAN-GP}         & 0.8499 & 0.8385 & \underline{0.8333} & \textbf{0.8288} & 0.8246 & 0.8105 & \textbf{0.9694} & \underline{0.9682} & \underline{0.9673} & 0.5790 & 0.5819 & 0.5752 \\
Melatt + ACGAN  \cite{ACGAN}          & 0.8494 & 0.8377 & 0.8325 & 0.8273 & 0.8245 & 0.8095 & 0.9690 & \textbf{0.9688} & \textbf{0.9687} & 0.5827 & 0.5821 & 0.5776 \\
Melatt + LRT-DDPM  \cite{LRT-DDPM}       & 0.8494 & 0.8387 & 0.8332 & 0.8274 & 0.8252 & \underline{0.8109} & 0.9689 & 0.9672 & 0.9664 & 0.5679 & 0.5672 & 0.5626 \\
Melatt + NetDiffus  \cite{Netdiffus}      & 0.8500 & 0.8376 & 0.8320 & 0.8276 & 0.8243 & 0.8102 & \textbf{0.9694} & 0.9681 & 0.9672 & 0.5618 & 0.5646 & 0.5587 \\
\midrule
TDDM-Melatt               & \textbf{0.9022} & \textbf{0.8889} & \textbf{0.8885} & 0.8272 & 0.8242 & 0.8102 & 0.9682 & 0.9667 & 0.9659 & \textbf{0.6834} & \textbf{0.6844} & \textbf{0.6808} \\
\bottomrule
\end{tabular*}
}
\end{table*}

\begin{figure}[tbp]
	\centering
	{
	}
	\begin{subfigure}[b]{0.49\columnwidth}
		\centering
		\includegraphics[width=\linewidth]{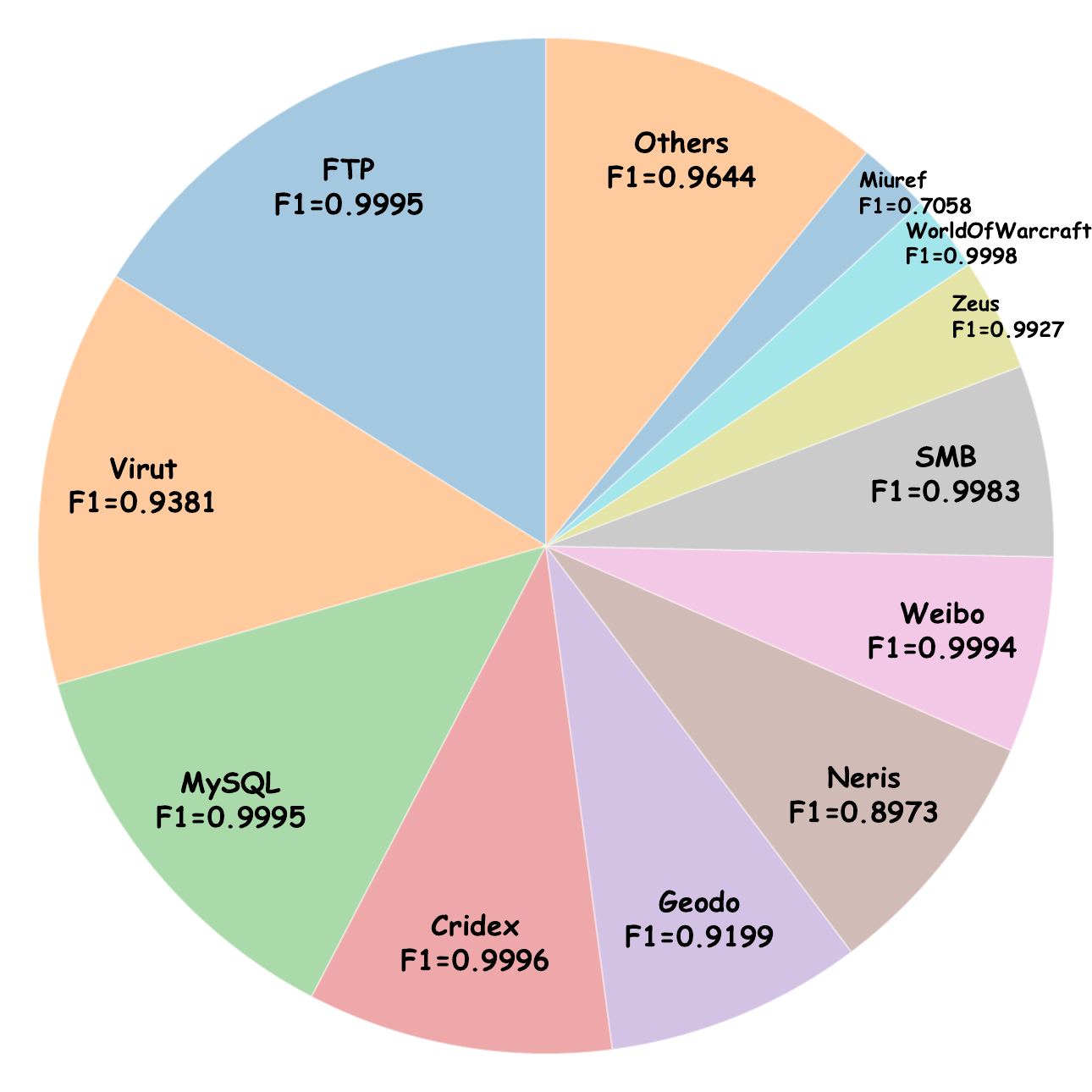}
		{
			\setlength{\abovecaptionskip}{-10pt}
			\setlength{\belowcaptionskip}{-5pt}
			\caption{Original F1 Score}
			\Description{The original per-class F1-score of Melatt on USTC-20class without data augmentation.}
		}
	\end{subfigure}
	\hfill
	\begin{subfigure}[b]{0.49\columnwidth}
		\centering
		\includegraphics[width=\linewidth]{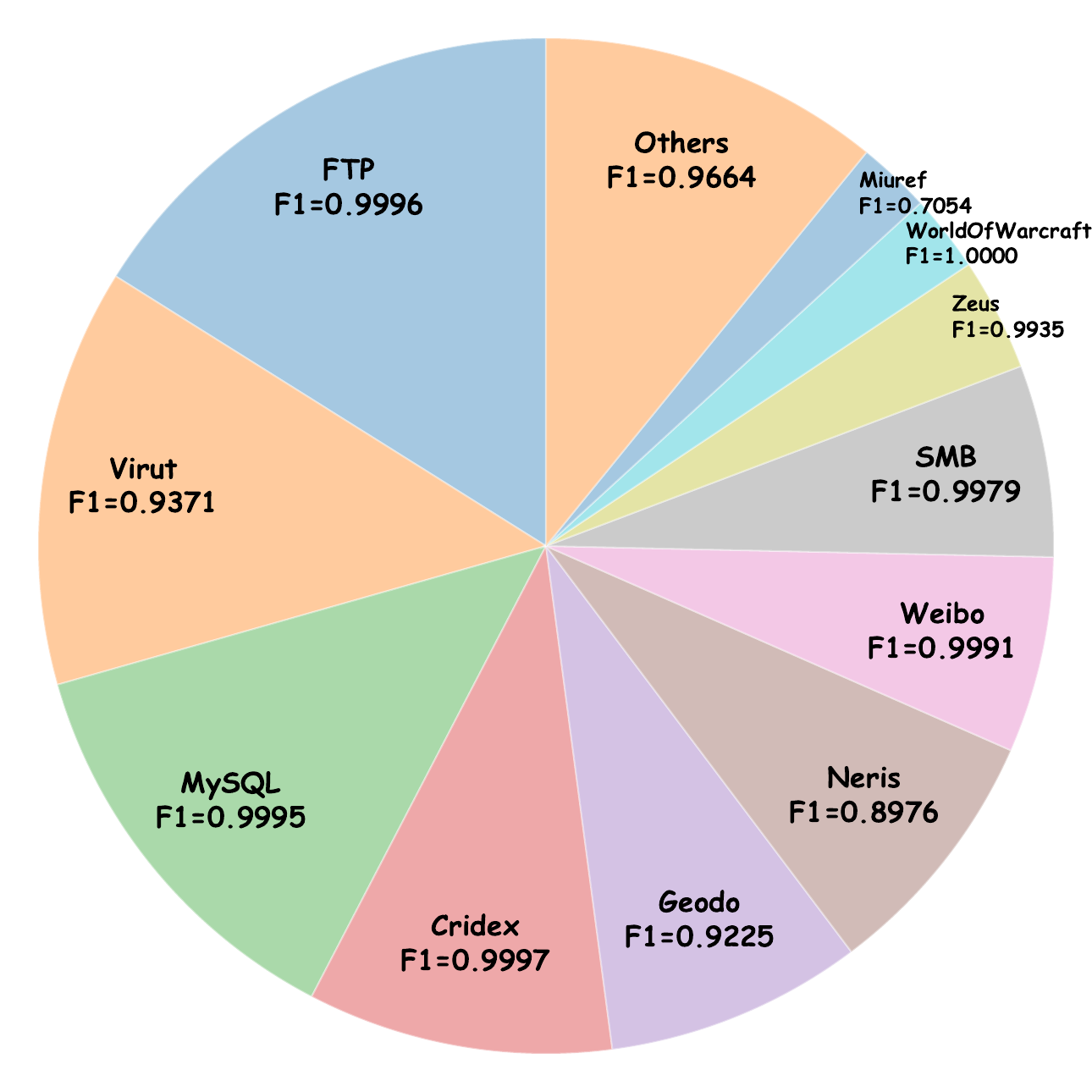}
		{
			\setlength{\abovecaptionskip}{-10pt}
			\setlength{\belowcaptionskip}{-5pt}
			\caption{Augmented F1 Score}
			\Description{The per-class F1-score of TDDM-Melatt on USTC-20class after TDDM data augmentation.}
		}
	\end{subfigure}
	{
		\caption{Per-class Performance on USTC-20class.}
		\Description{A set of four pie charts showing per-class classification results for USTC-20class. The top-left chart displays the original accuracy per class using Melatt without augmentation. The top-right chart displays the original F1-score per class using Melatt. The bottom-left chart displays the per-class accuracy after augmenting training data with TDDM. The bottom-right chart displays the per-class F1-score after TDDM augmentation. All charts indicate performance metrics across the 20 classes.}
		\label{fig:pie_chart}
	}
\end{figure}

Analysis of the experimental results reveals that, in VPN-16class and USTC-20class tasks, the performance gains contributed by all generative models are negligible, and several models even induce performance degradation. To further investigate this phenomenon, we conduct additional verification experiments, as illustrated in Figure~\ref{fig:pie_chart} (and in Appendix Figure~\ref{fig:0}). Taking USTC-20class as an example, certain classes with abundant samples still suffer from relatively low classification F1 (e.g., Neris 0.8973), whereas the improvement for classes with scarce samples is limited (e.g., Others from 0.9644 to 0.9664). Consequently, data augmentation targeted at minority classes struggles to yield a marked improvement in overall classification performance. Instead, the synthetic data may introduce noise that adversely affects the training dynamics of the majority classes.

In contrast, on the VPN-6class and TLS-120class tasks, TDDM is the sole method that achieves statistically significant performance improvements. The F1-score gains reach 9.98\% on TLS-120class and 5.56\% on VPN-6class, respectively. Conversely, all other baseline generative models fail to deliver any notable performance enhancement on these two tasks. These results indicate that the generalization capacity of existing data generation baselines is extremely limited. Most of these baselines do not adequately account for the high-dimensional sparsity and temporal dependencies inherent in network traffic data, and they may introduce substantial noise that disrupts the classifier's decision boundaries.

\begin{table}[tbp]
	\centering
	{
		\caption{Quality Metrics of Imbalance Handling Methods.}
		\label{tab:generation_quality}
		\begin{tabular}{p{0.25\linewidth}|>{\centering\arraybackslash}m{0.3\linewidth}|>{\centering\arraybackslash}m{0.3\linewidth}}
			\toprule
			Method & MMD $\downarrow$ & APD $\uparrow$ \\
			\midrule
			GAN & $0.3930$ & $2.4315$ \\
			WGAN-GP & $0.3310$ & $2.6534$ \\
			ACGAN & $0.3962$ & $2.4721$ \\
			\midrule
			Oversampling & $0.0591$ & $3.0292$ \\
			SMOTE & $0.0518$ & $3.0356$ \\
			ADASYN & $0.0506$ & $2.9781$ \\
			SimpleCopy & $0.0545$ & $3.0558$ \\
			\midrule
			TDDM & $0.2667$ & $2.9633$ \\
			\bottomrule
		\end{tabular}
	}
\end{table}

To further verify the necessity and effectiveness of TDDM, we conduct a systematic comparative experiment by incorporating five classical imbalance handling methods based on existing GAN-variant-based generative schemes, including Oversampling, Simple Copy, SMOTE, and ADASYN. We adopt MMD and APD as direct generation quality evaluation metrics, and carry out all experiments on the IDS-15class task. The experimental results are presented in Table \ref{tab:generation_quality}. It is worth noting that although the classical imbalance handling methods achieve favorable MMD and APD scores, they contribute little to downstream classification performance, or even produce negative effects. Compared with vanilla GAN, TDDM achieves competitive MMD and APD performance, while delivering the strongest classification gain. This demonstrates that the diffusion model achieves a better balance between sample fidelity and task-relevant signal preservation.

\begin{figure}[tbp]
	\centering
	\begin{minipage}{0.48\linewidth}
		\centering
		\includegraphics[width=\linewidth]{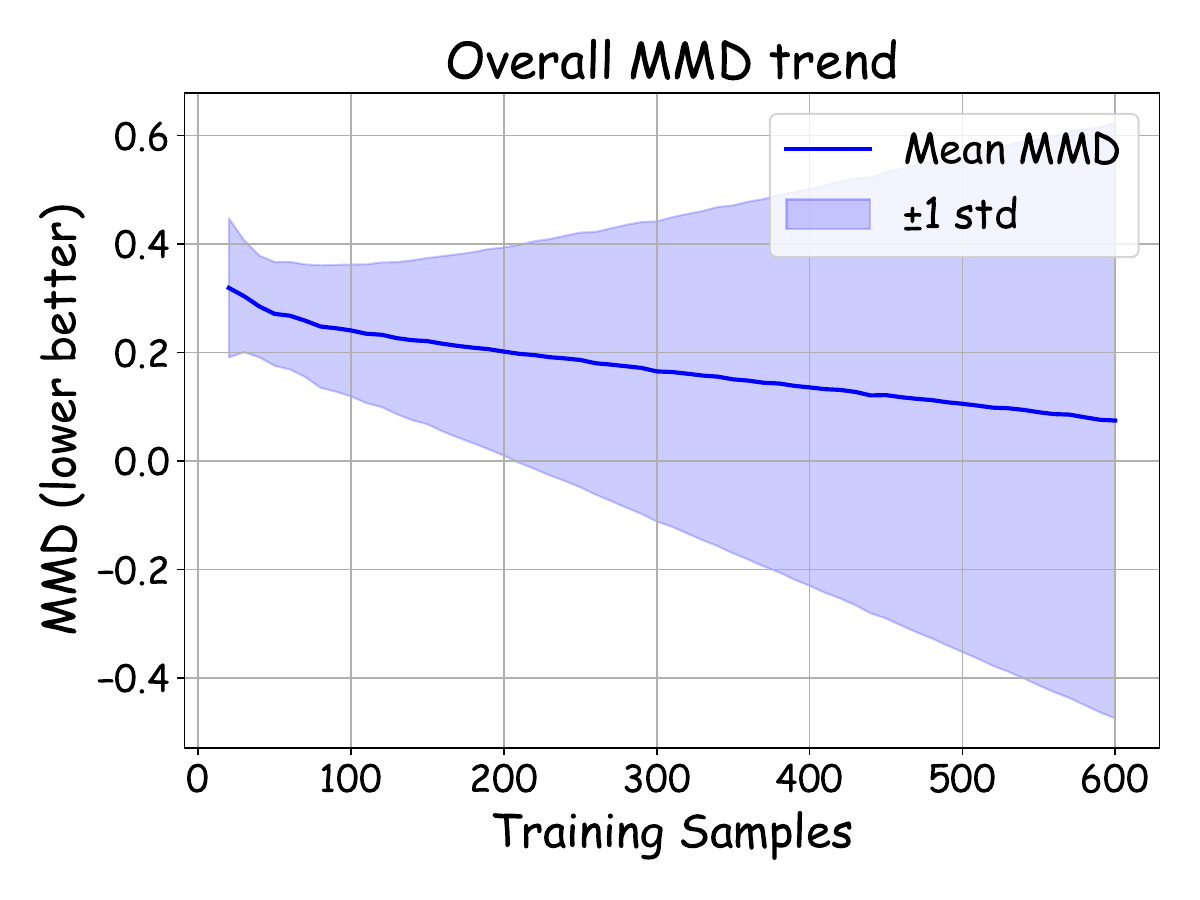}
	\end{minipage}
	\hfill
	\begin{minipage}{0.48\linewidth}
		\centering
		\includegraphics[width=\linewidth]{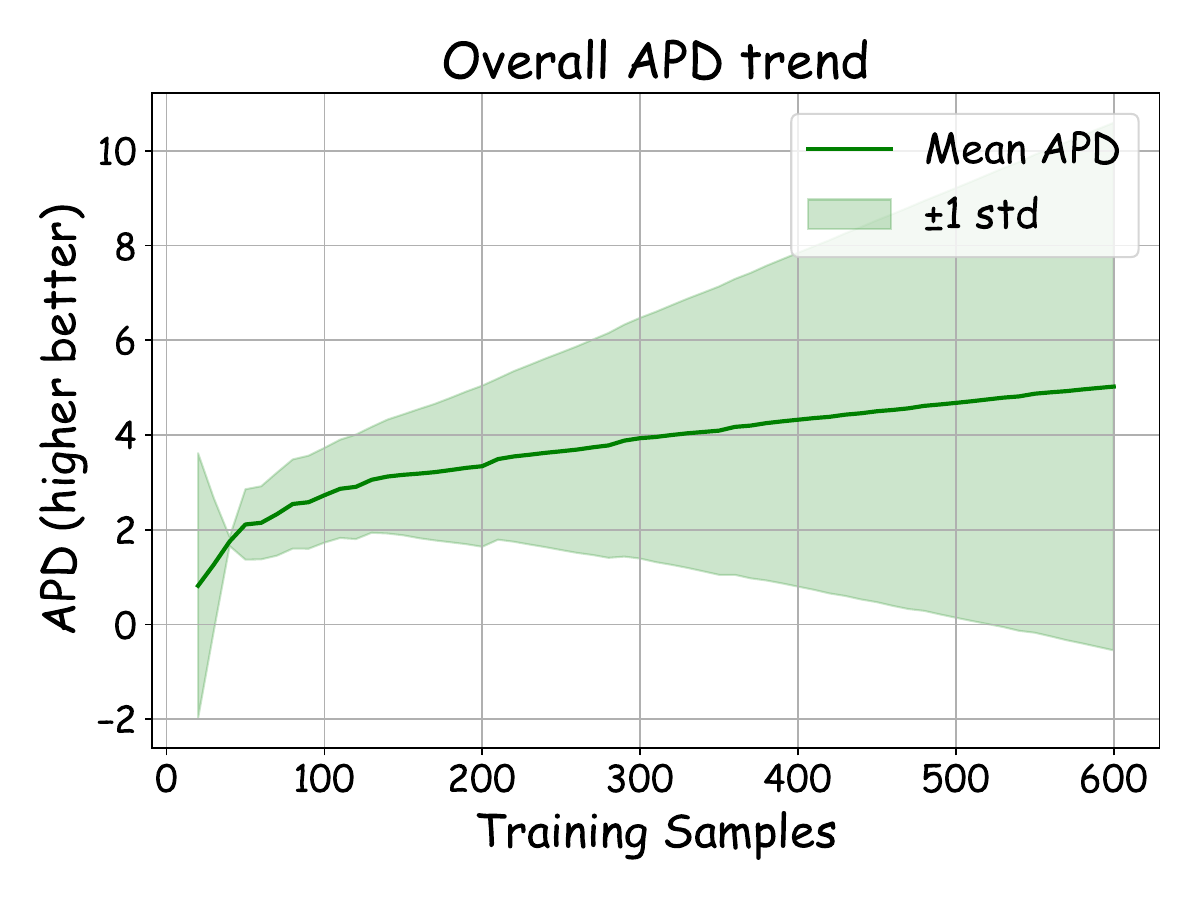}
	\end{minipage}
	\\
	\begin{minipage}{0.95\linewidth}
		\centering
		\includegraphics[width=\linewidth]{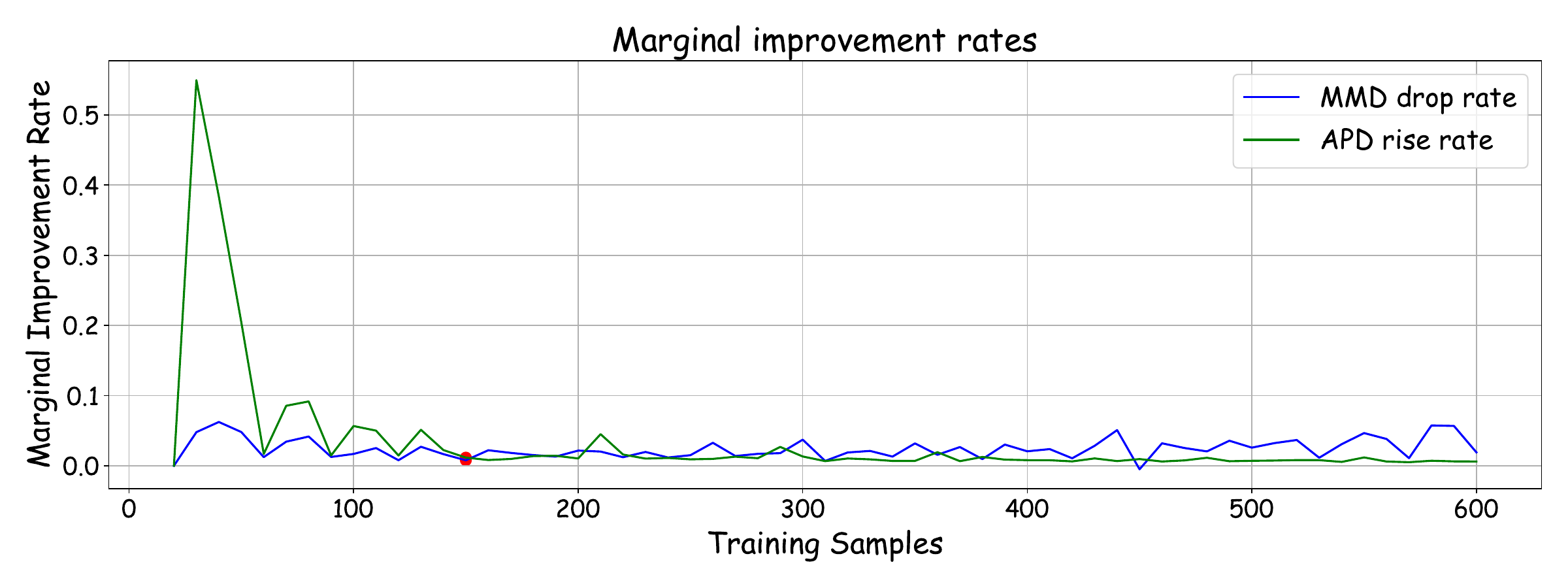}
	\end{minipage}
	\caption{Convergence Analysis in Training Sample Size.}
	\label{fig:tddm_convergence}
\end{figure}

To determine the minimum number of samples required to train a reliable class-wise diffusion model, we conduct an experiment where we progressively increase the training sample size from 20 to 600 (in steps of 5) for a single minority class and observe the MMD and APD curves, as shown in Figure~\ref{fig:tddm_convergence}. To identify the optimal minimum sample size, we adopt a unified marginal improvement criterion combining the 50\% convergence rule and standard deviation constraints. As shown in the marginal improvement rate curves in Figure~\ref{fig:tddm_convergence}, the inflection point is identified at 150 training samples, where the MMD improvement reaches 40.1\% of its total reduction, and the APD improvement reaches 55.7\% of its total increase. Beyond this point, the marginal gains diminish significantly, and the cross-class standard deviations of MMD and APD converge to stable levels (0.1533 and 1.2766, respectively). Based on these results, we recommend a minimum of 150 samples per class for training a reliable single-class diffusion model. This threshold ensures that the model captures sufficient intra-class diversity while avoiding overfitting to limited data. When the training sample size reaches 600, the MMD and APD scores of TDDM reach levels comparable to those of classical oversampling methods.

Through targeted enhancements to the diffusion generative process---including feature selection, class-wise training, centroid-guided compact sampling, and nearest-neighbor padding---TDDM provides an effective technical pathway for network traffic classification in scenarios characterized by limited samples and class imbalance.

\subsection{Condition-Guided Noise Sampling Analysis}
\label{5.4}

\begin{figure*}[tbp]
	\centering
	\setlength{\tabcolsep}{0pt} 
	\begin{tabular}{@{} 
			>{\centering\arraybackslash}m{0.12\textwidth} 
			*{5}{>{\centering\arraybackslash}m{0.176\textwidth}} 
			@{}}
		\toprule
		\textbf{Timestep} & 900 & 600 & 400 & 200 & 0 \\
		\midrule
		\textbf{IDS (UG)} &
		\includegraphics[width=0.176\textwidth]{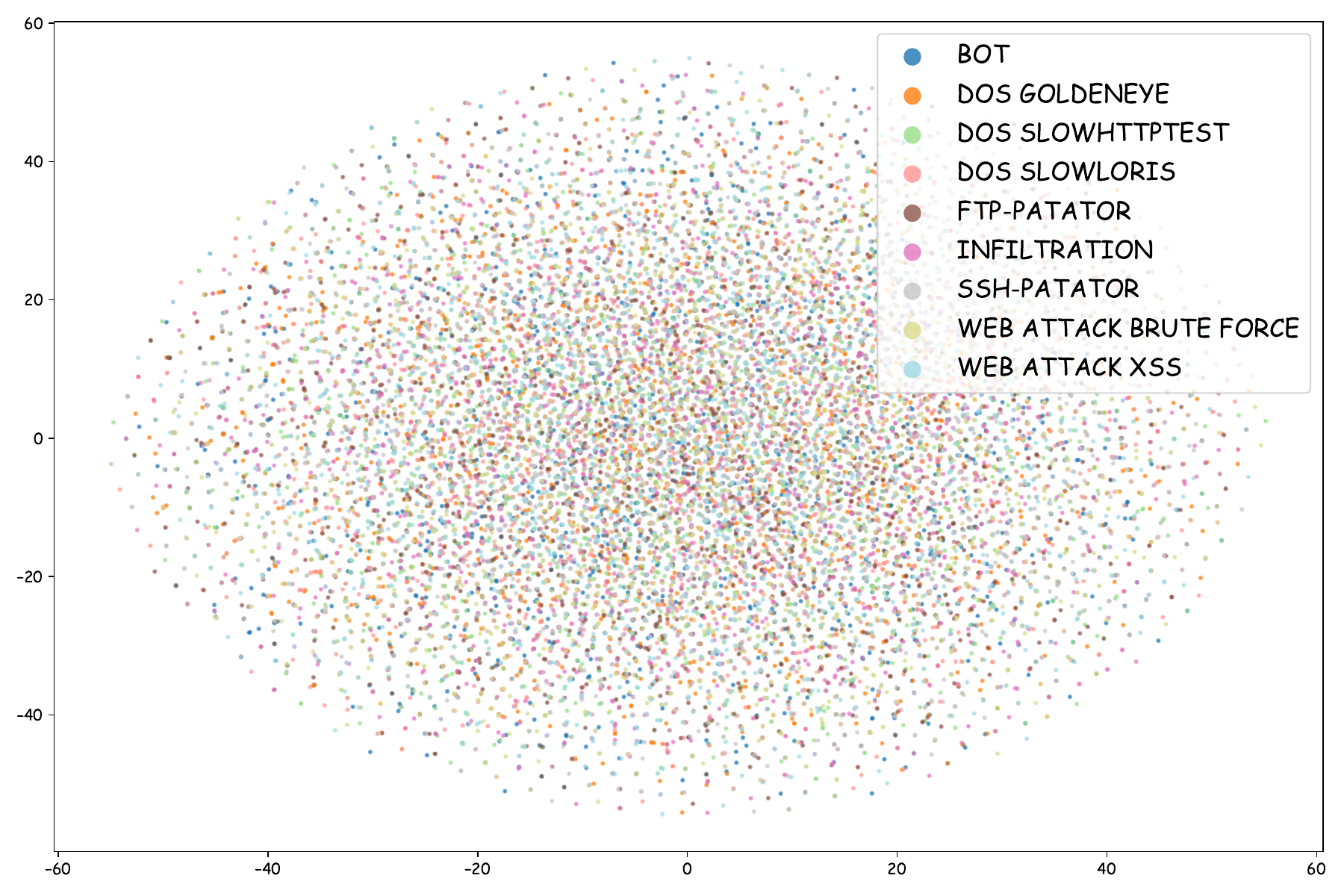} &
		\includegraphics[width=0.176\textwidth]{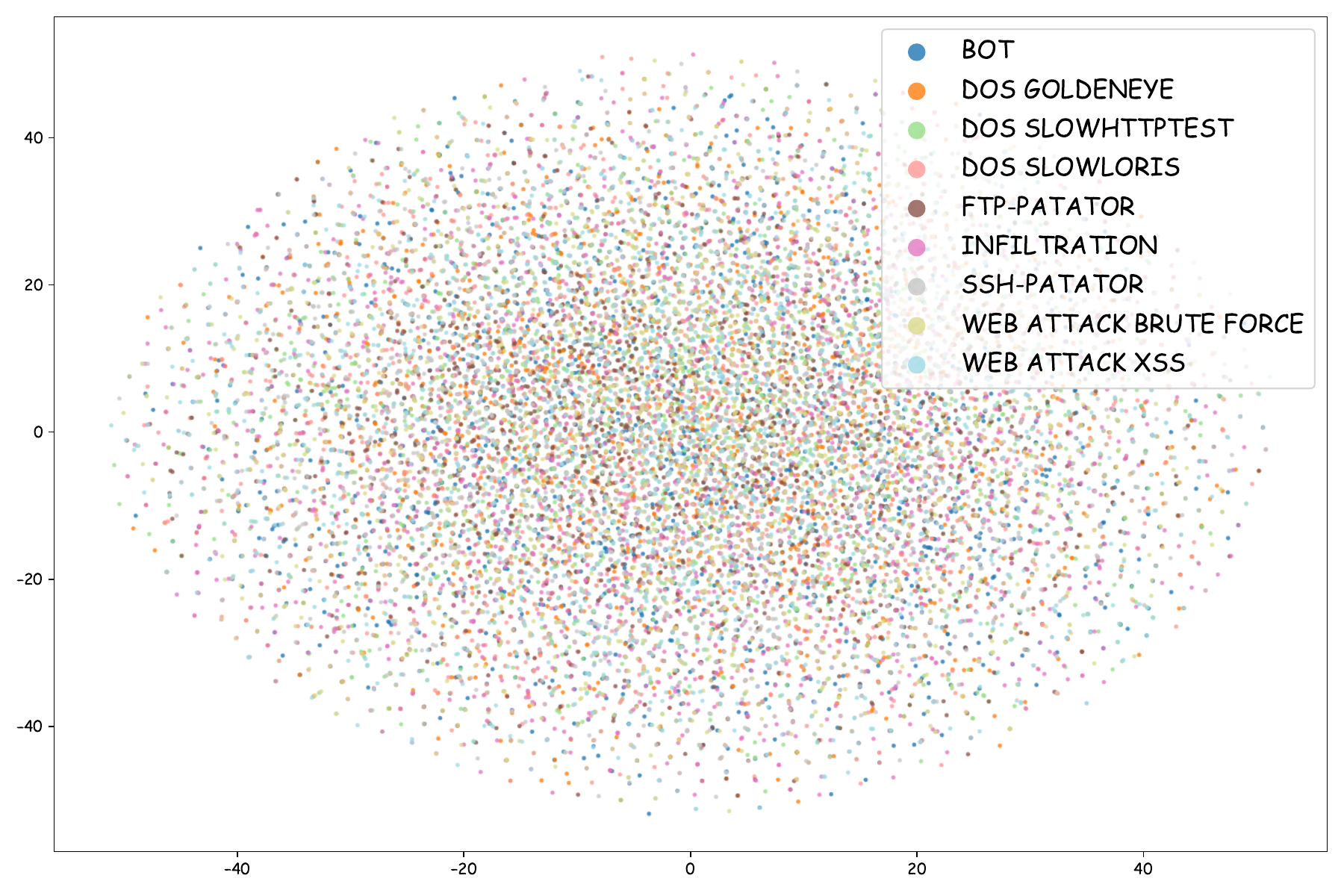} &
		\includegraphics[width=0.176\textwidth]{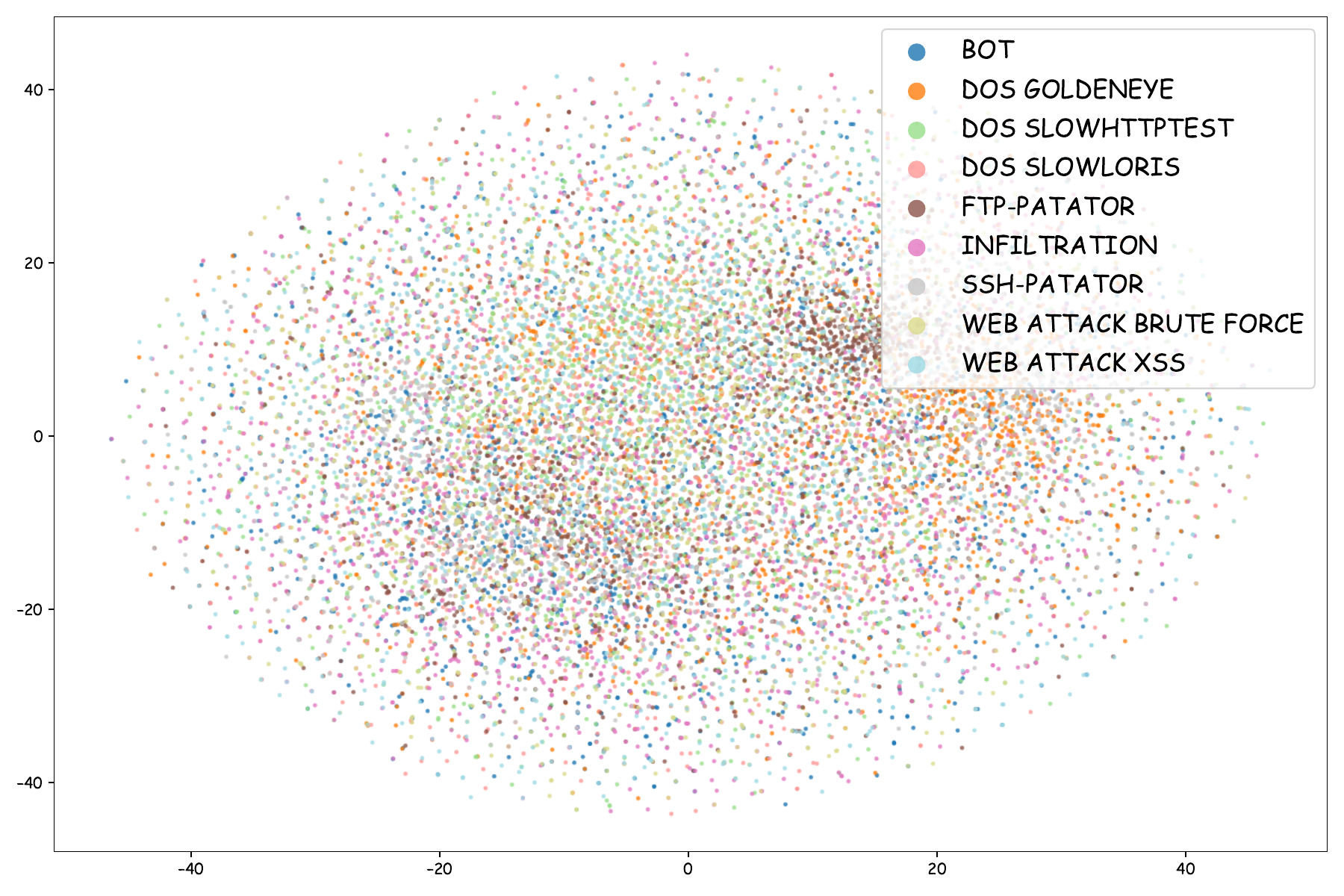} &
		\includegraphics[width=0.176\textwidth]{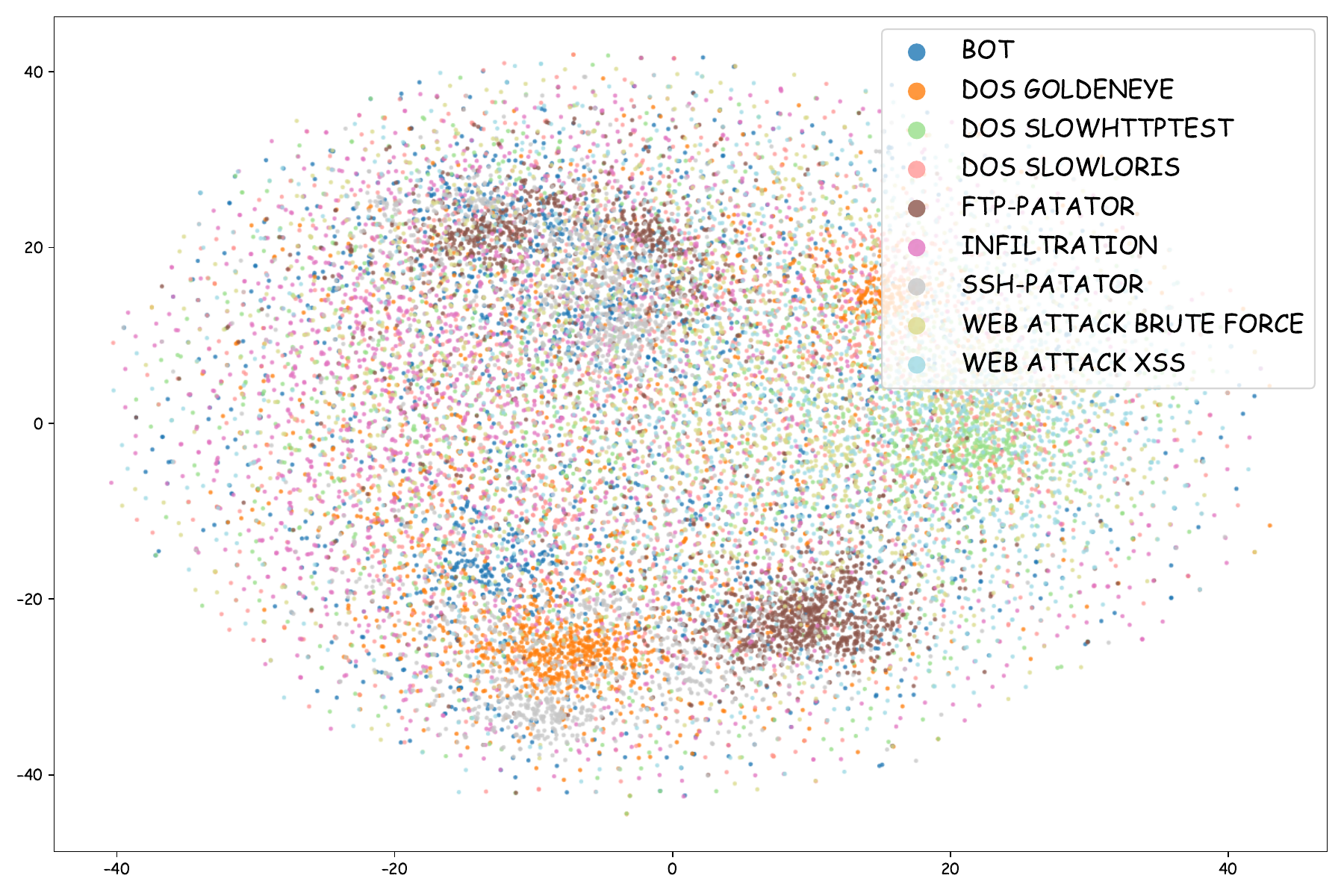} &
		\includegraphics[width=0.176\textwidth]{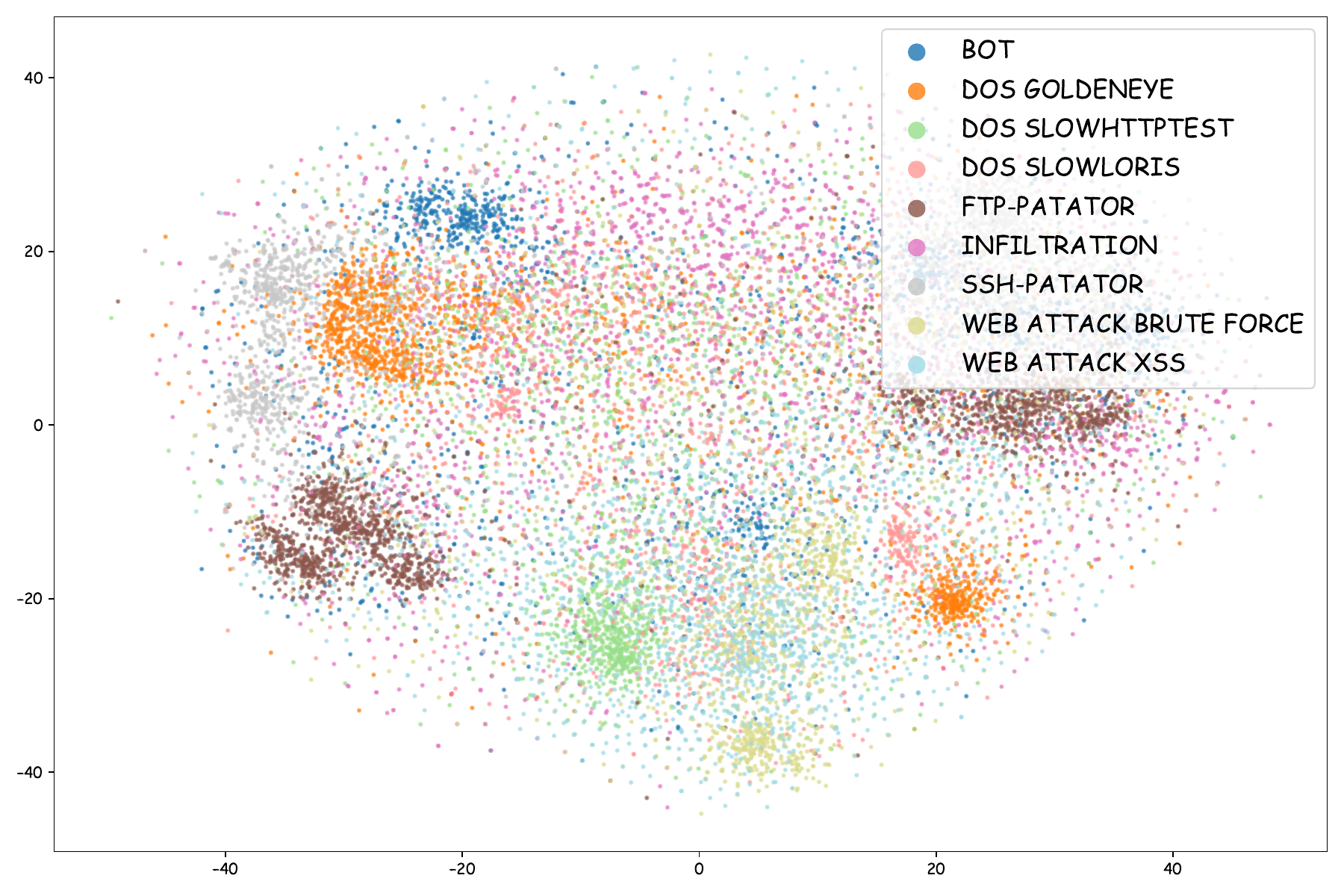} \\
		\midrule
		\textbf{IDS (G)} &
		\includegraphics[width=0.176\textwidth]{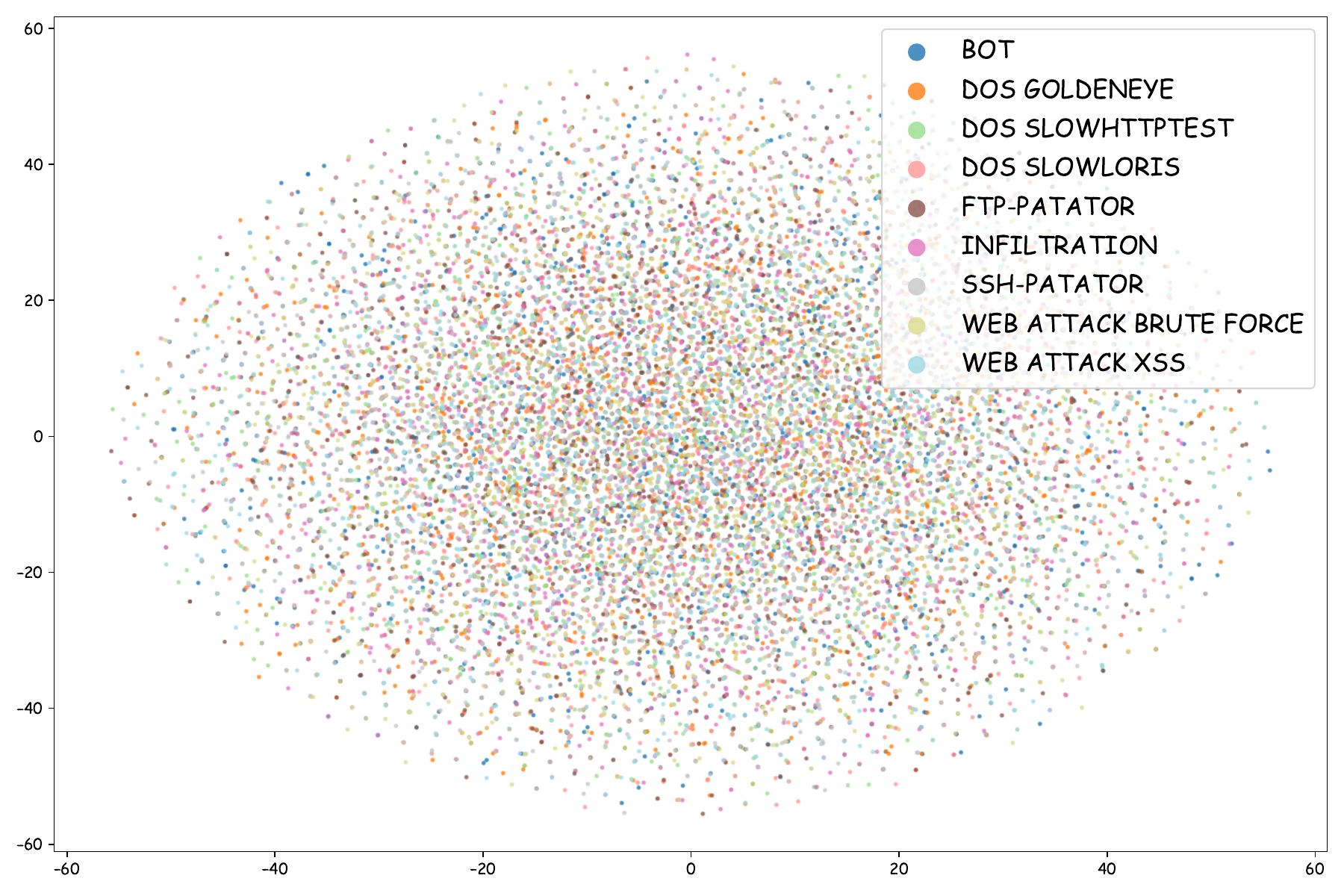} &
		\includegraphics[width=0.176\textwidth]{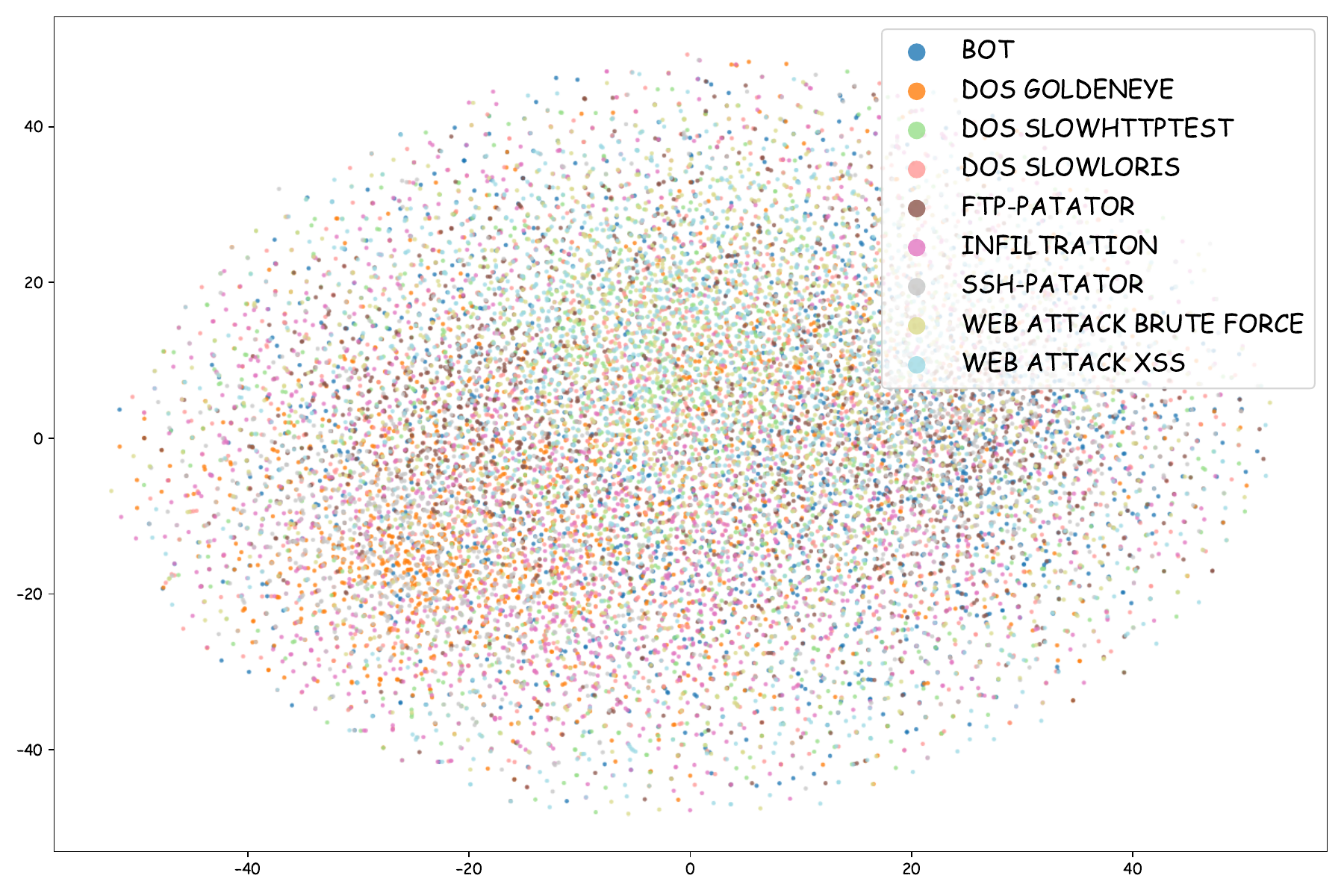} &
		\includegraphics[width=0.176\textwidth]{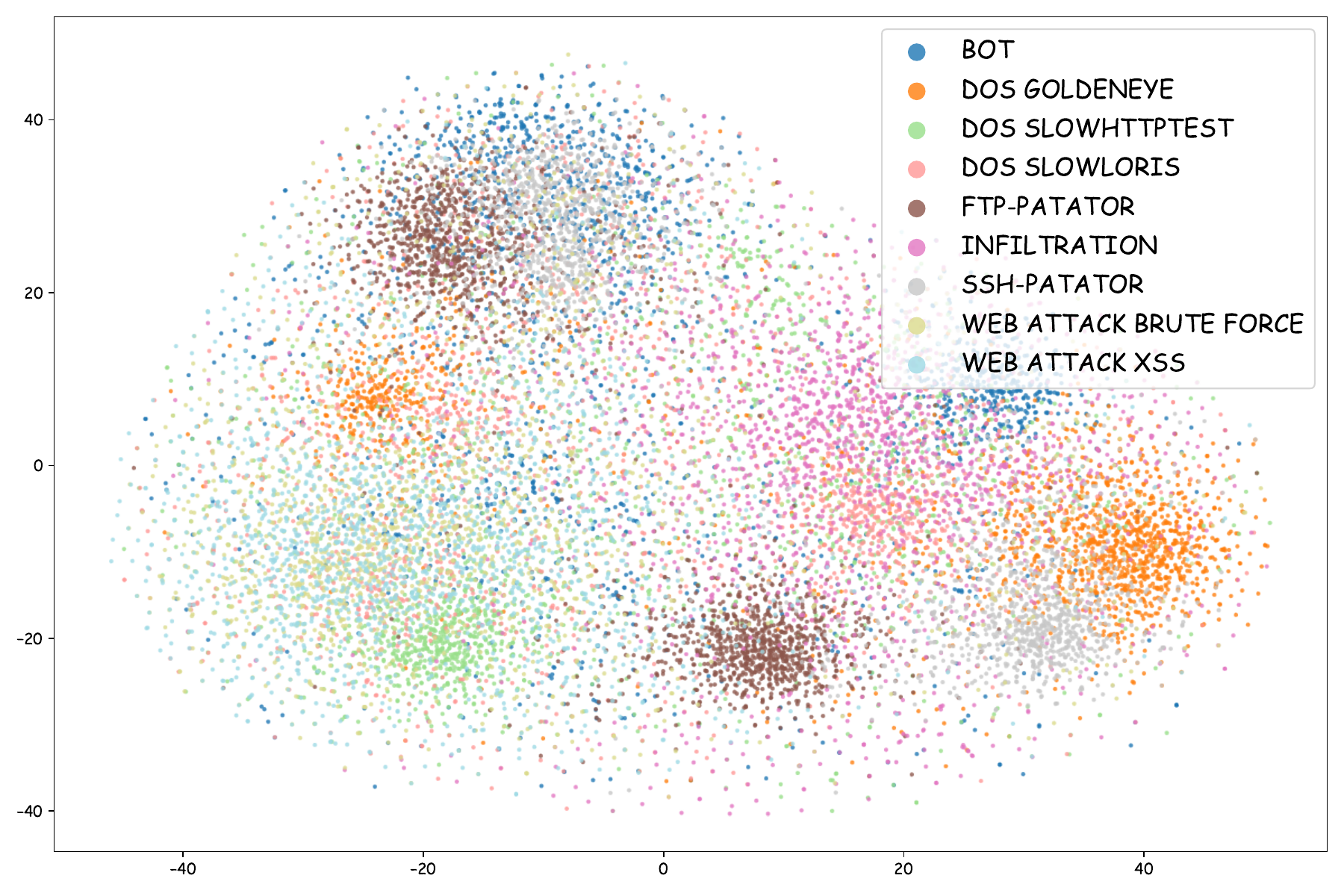} &
		\includegraphics[width=0.176\textwidth]{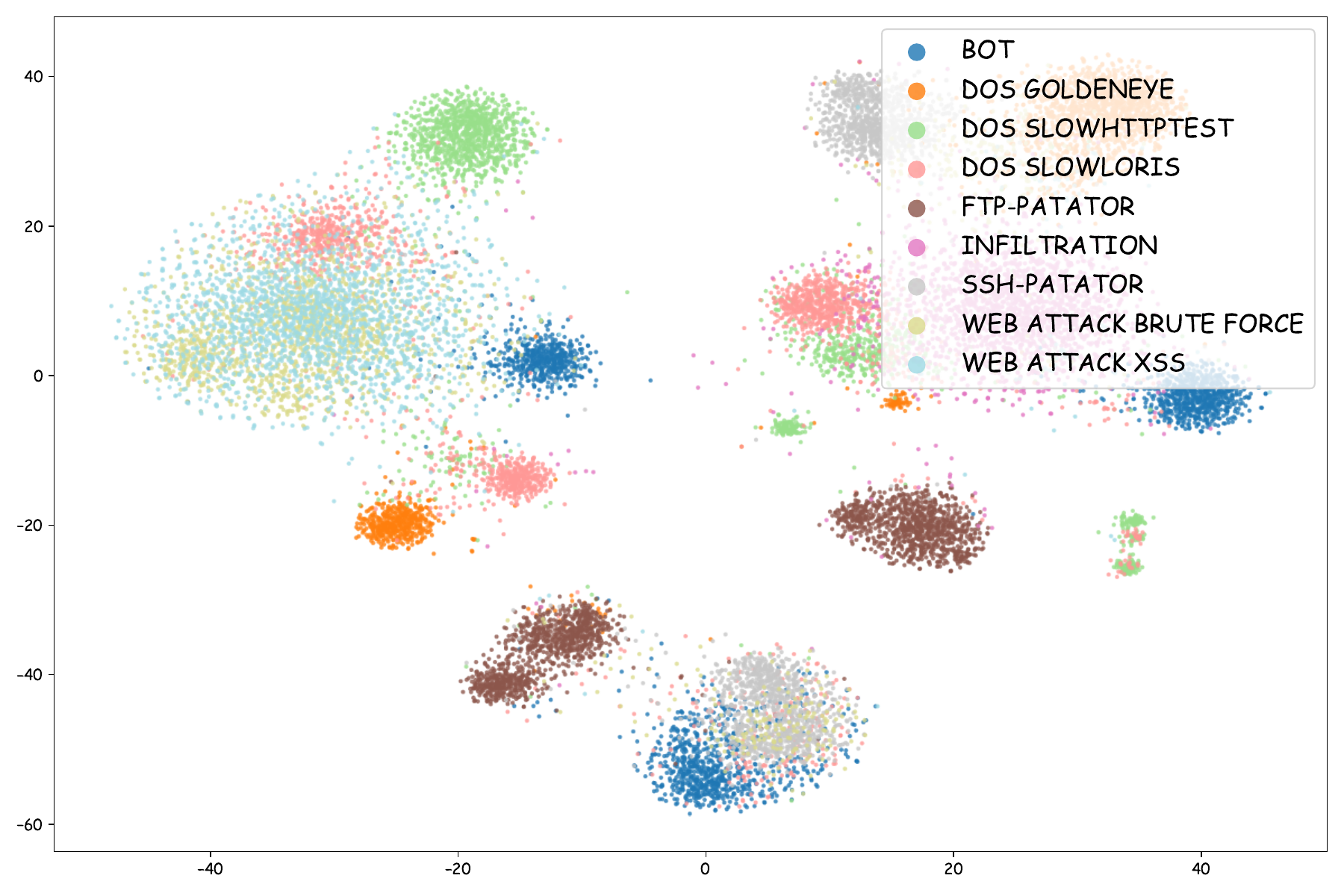} &
		\includegraphics[width=0.176\textwidth]{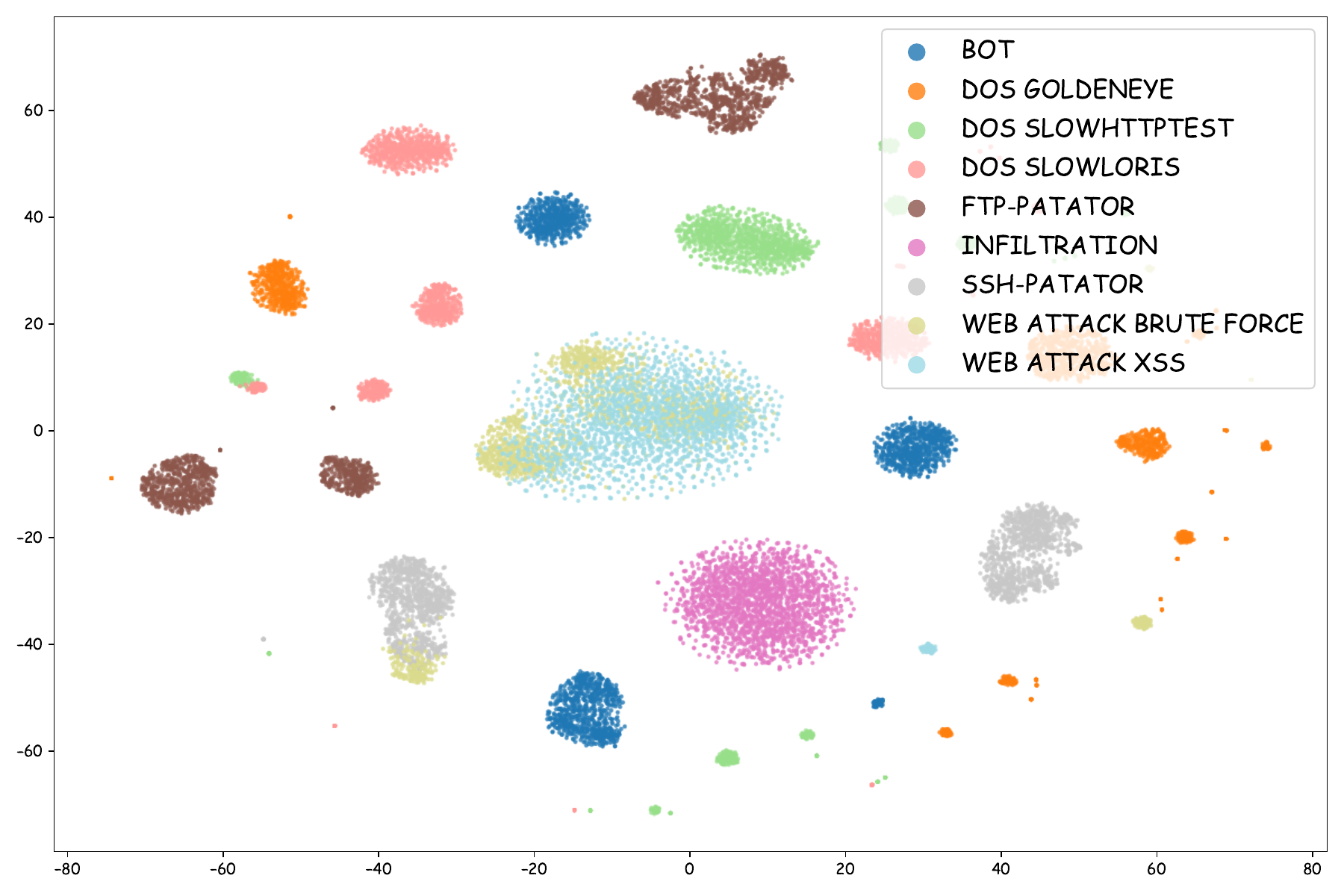} \\
		\midrule
		\textbf{TLS (UG)} &
		\includegraphics[width=0.176\textwidth]{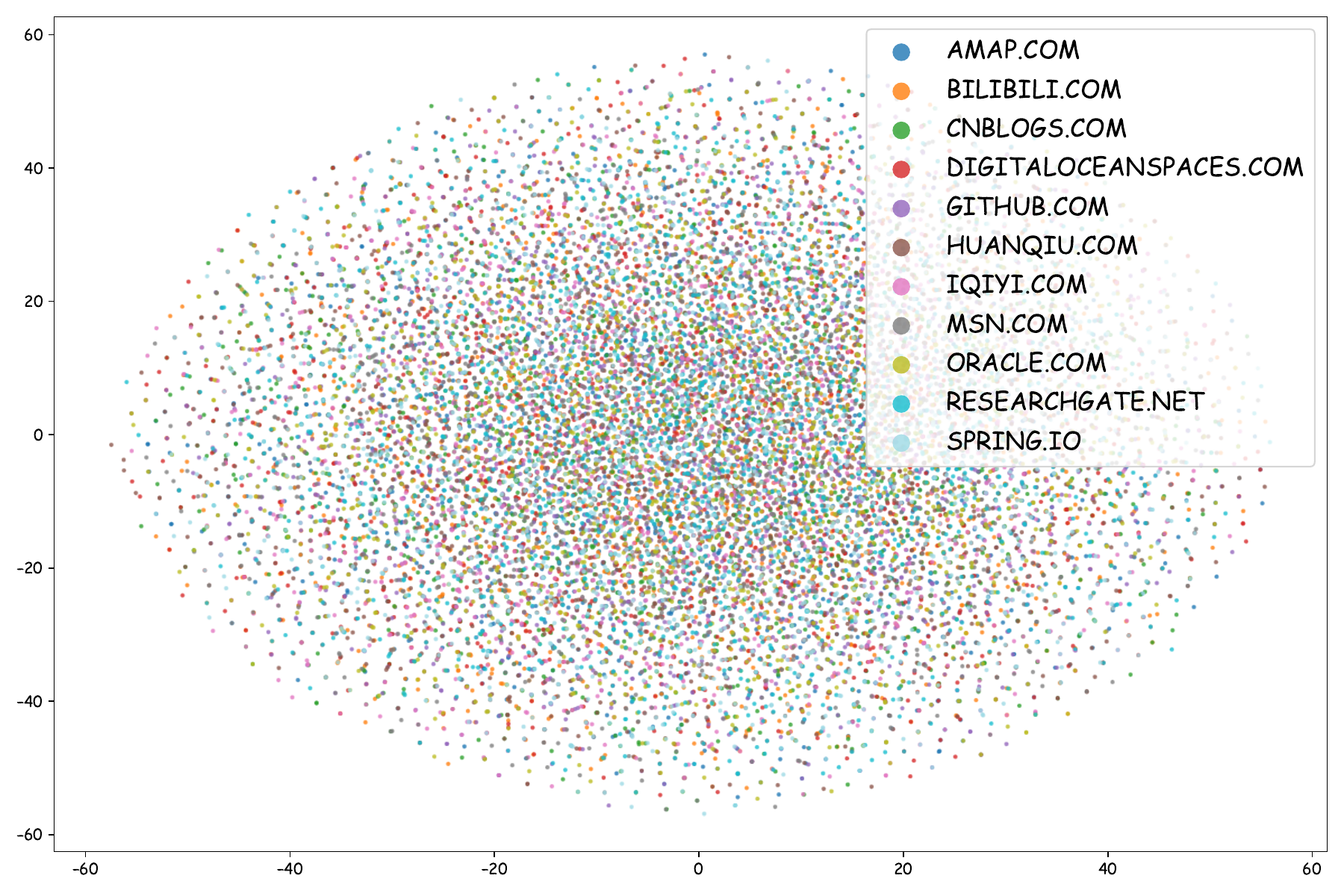} &
		\includegraphics[width=0.176\textwidth]{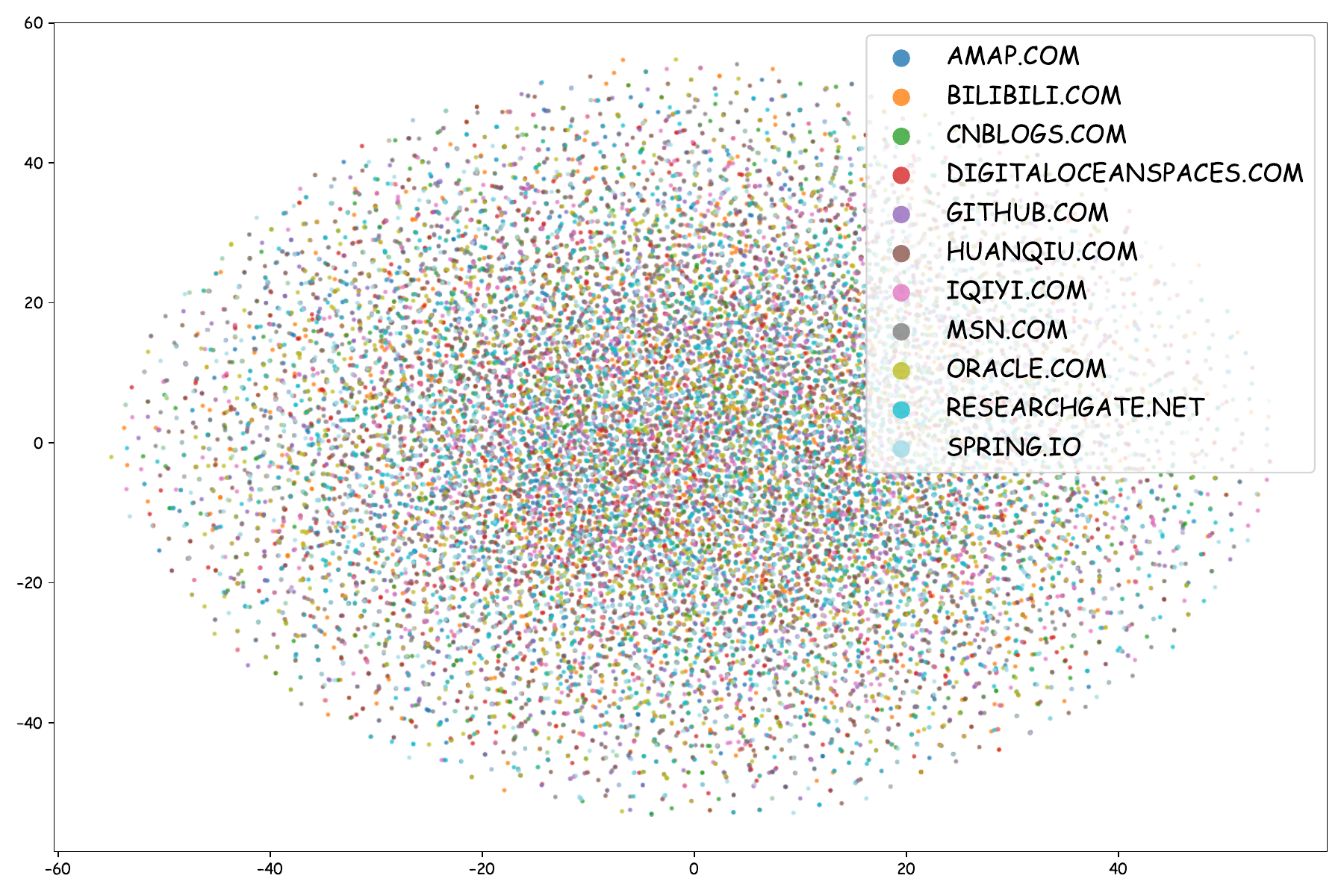} &
		\includegraphics[width=0.176\textwidth]{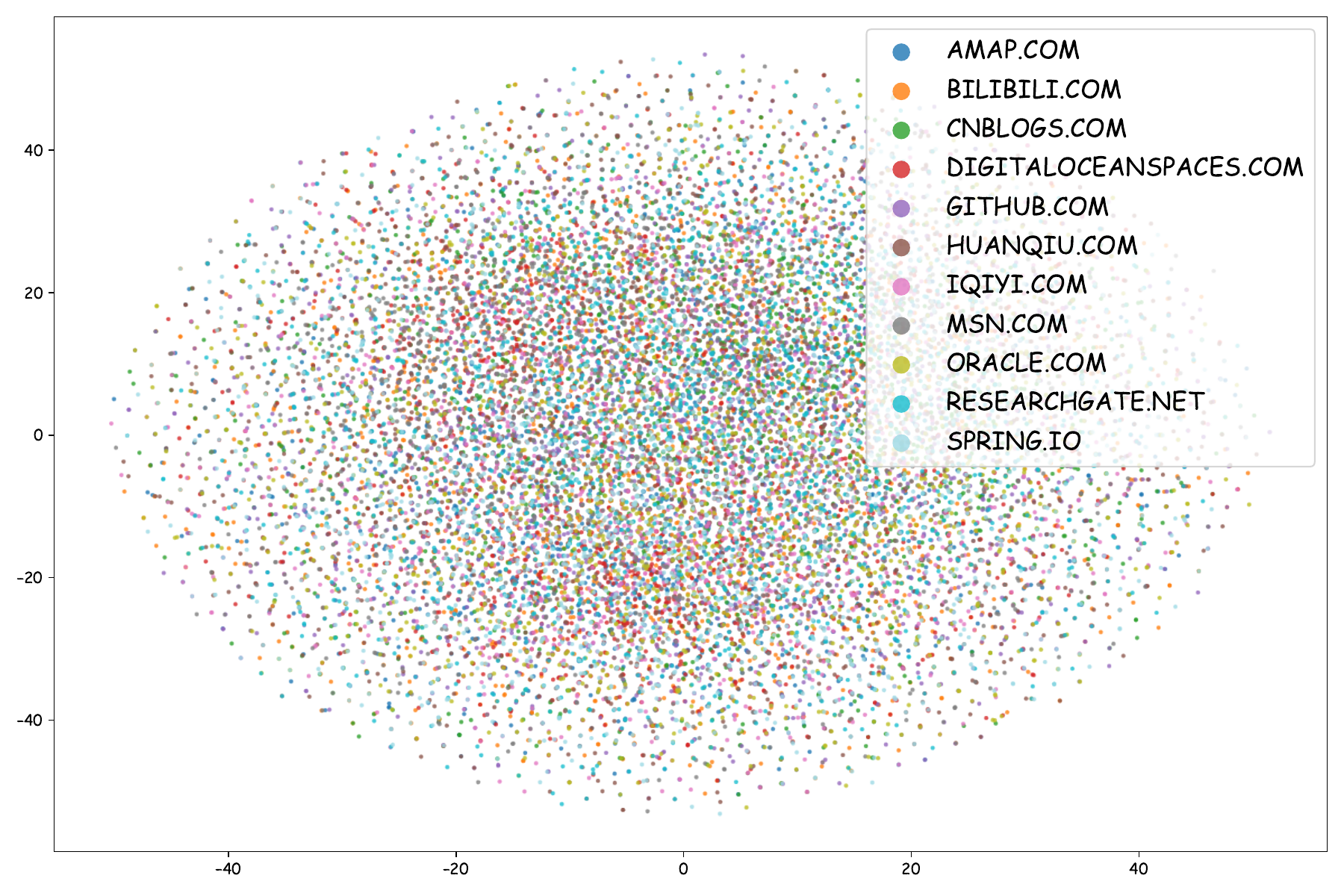} &
		\includegraphics[width=0.176\textwidth]{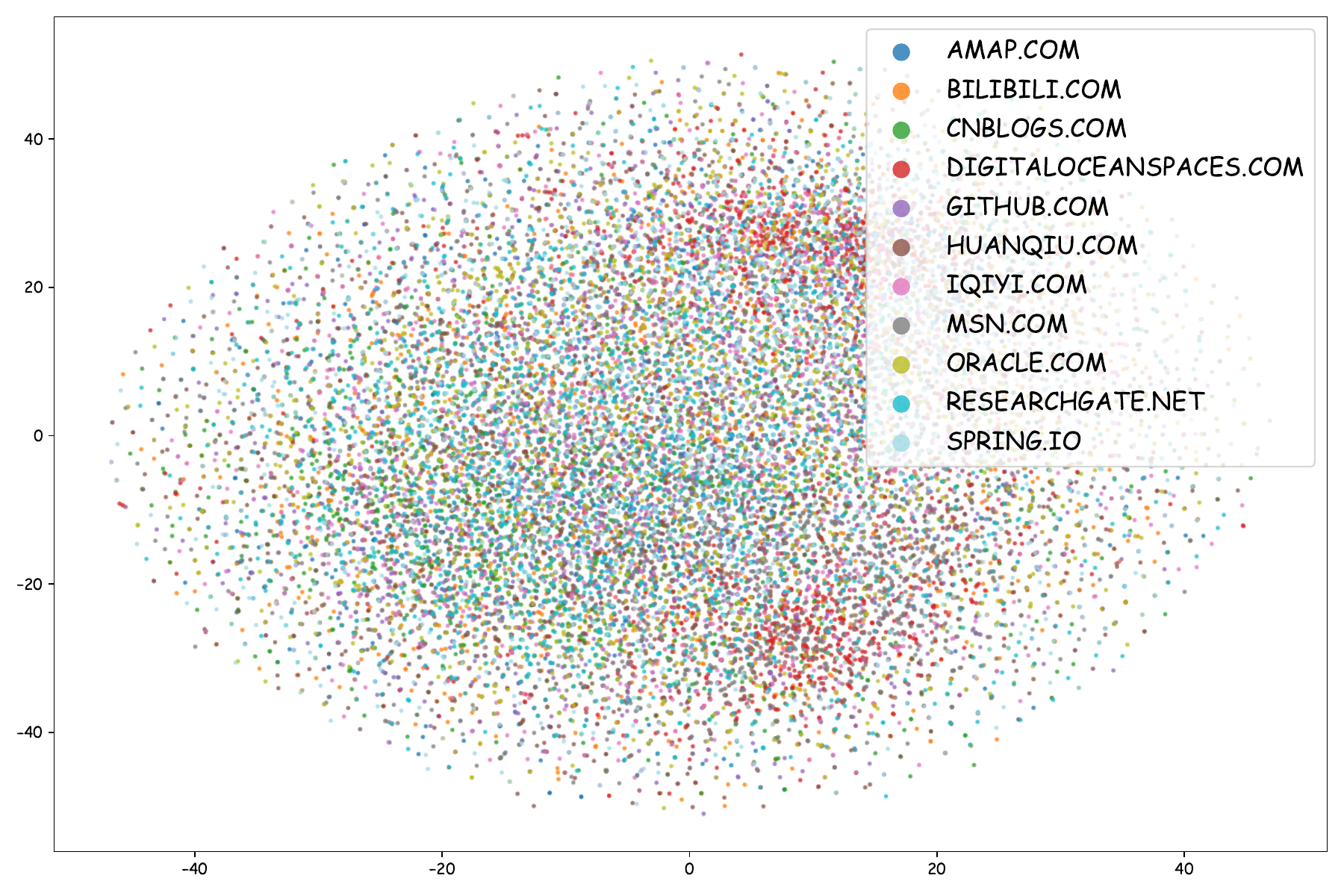} &
		\includegraphics[width=0.176\textwidth]{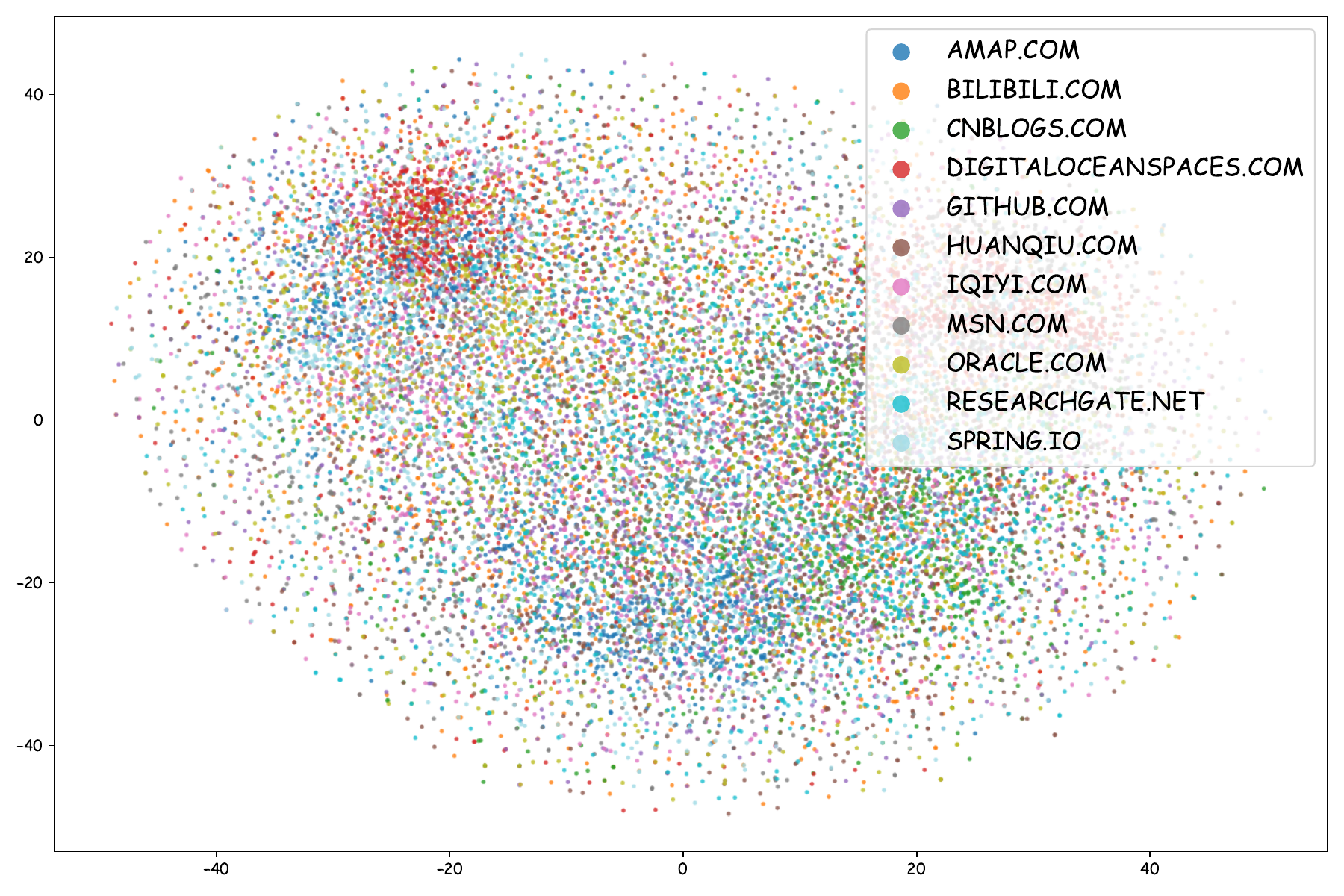} \\
		\midrule
		\textbf{TLS (G)} &
		\includegraphics[width=0.176\textwidth]{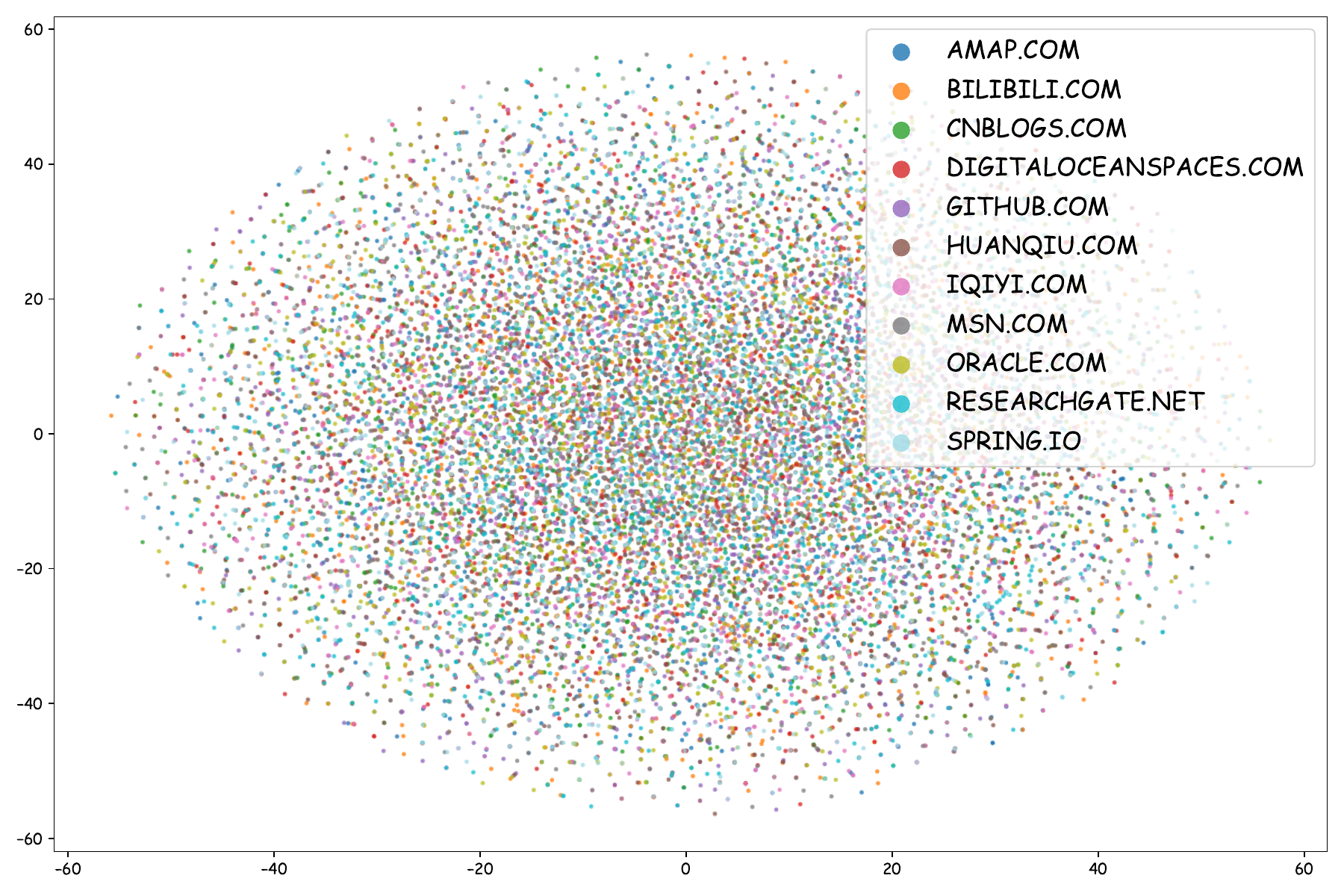} &
		\includegraphics[width=0.176\textwidth]{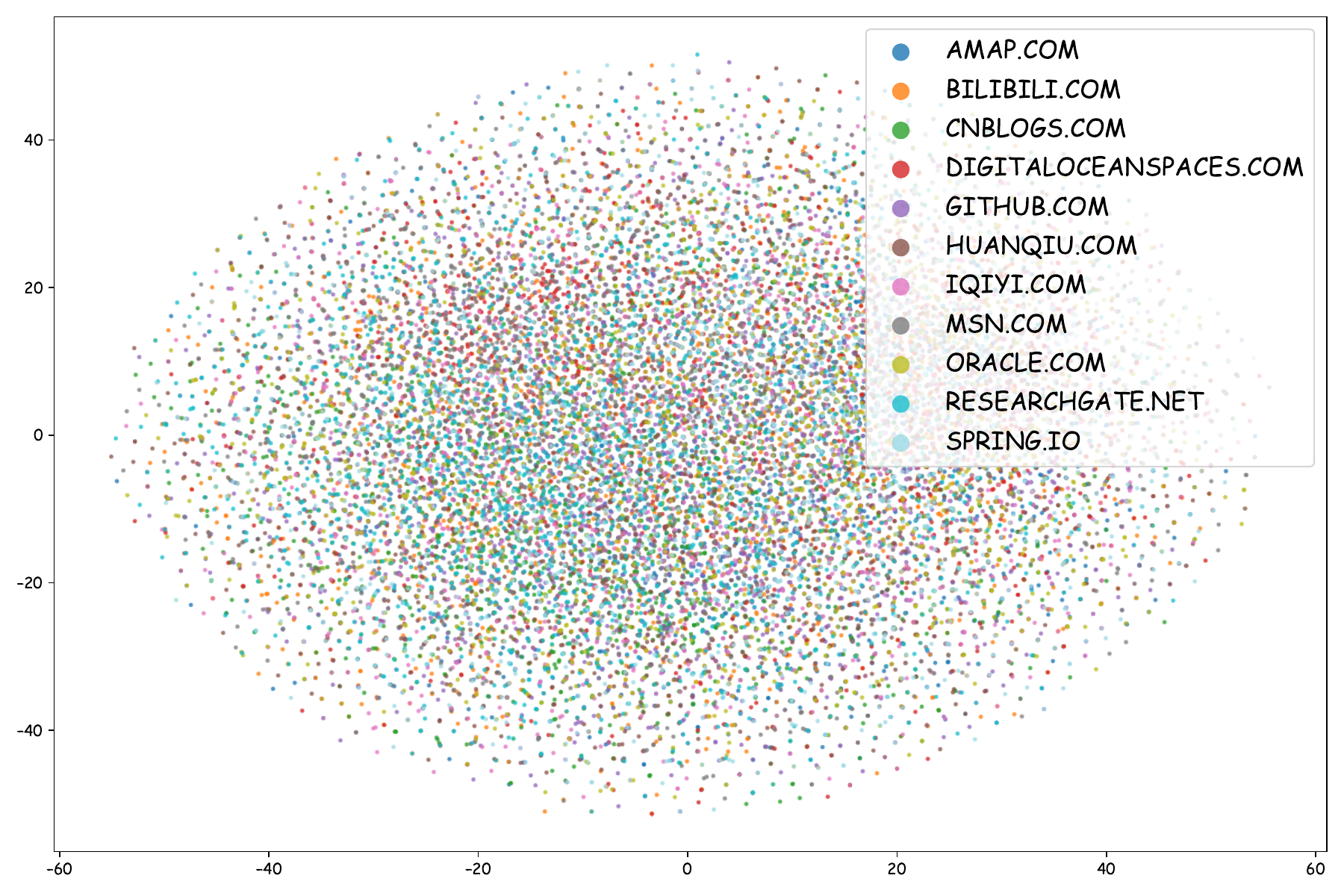} &
		\includegraphics[width=0.176\textwidth]{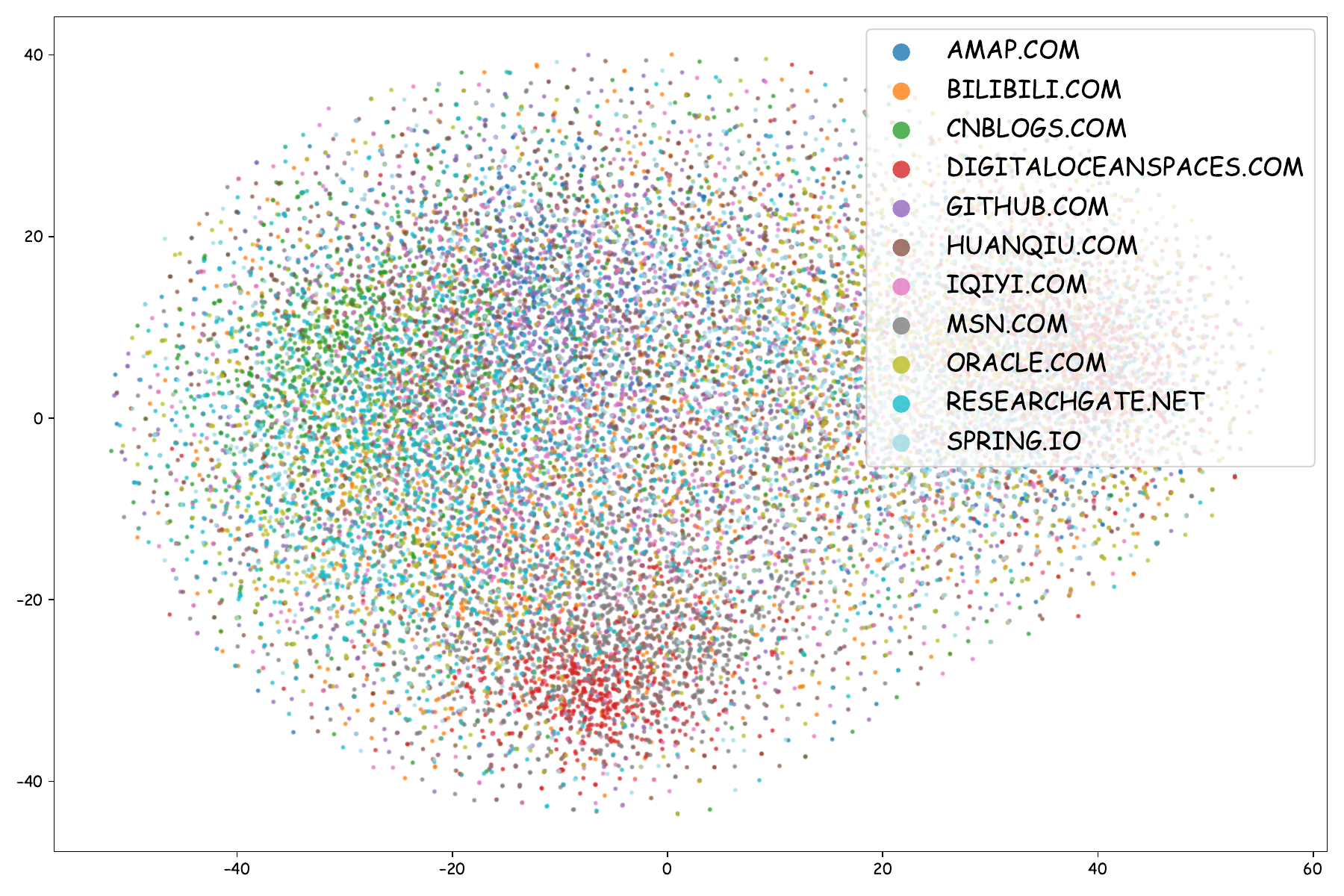} &
		\includegraphics[width=0.176\textwidth]{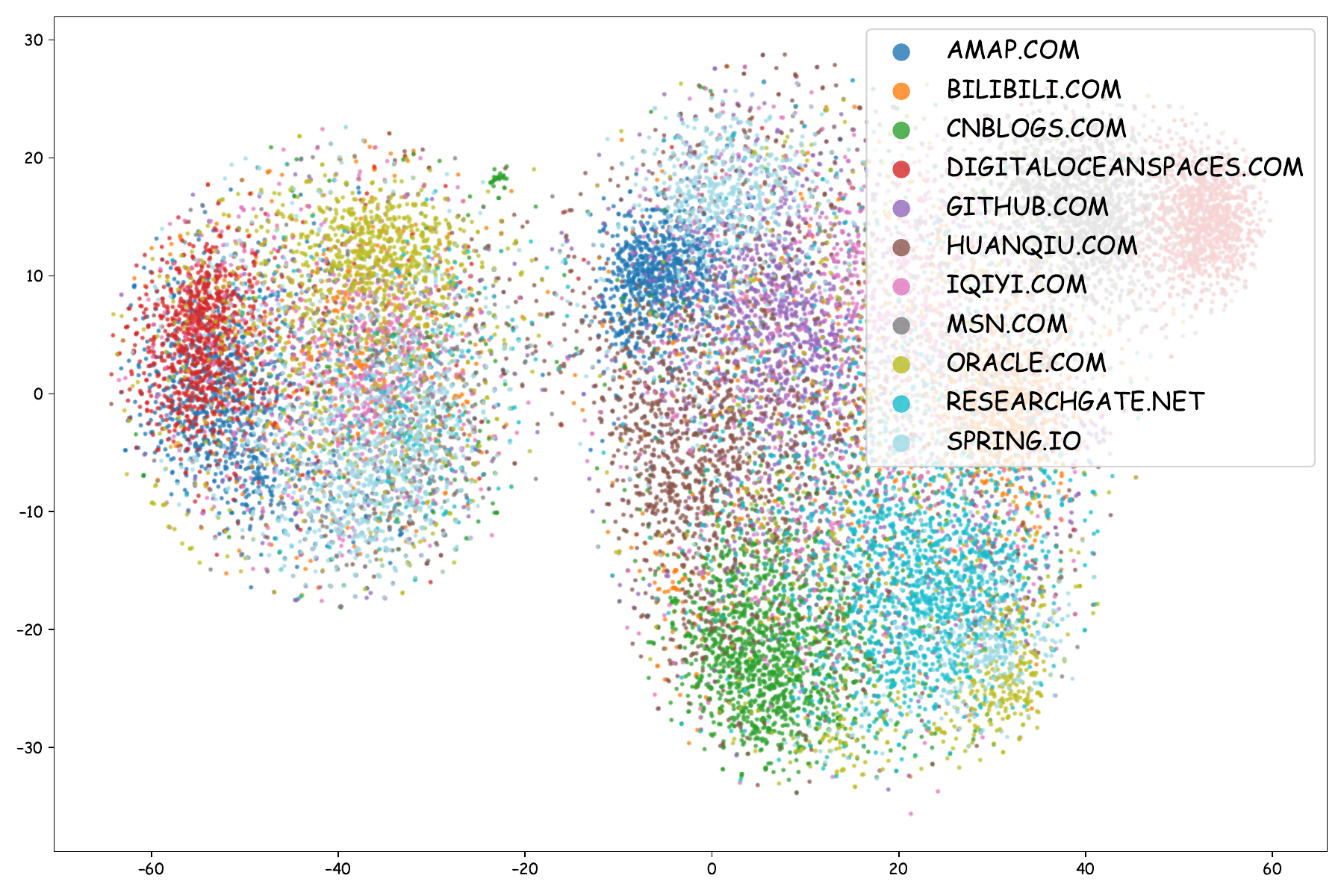} &
		\includegraphics[width=0.176\textwidth]{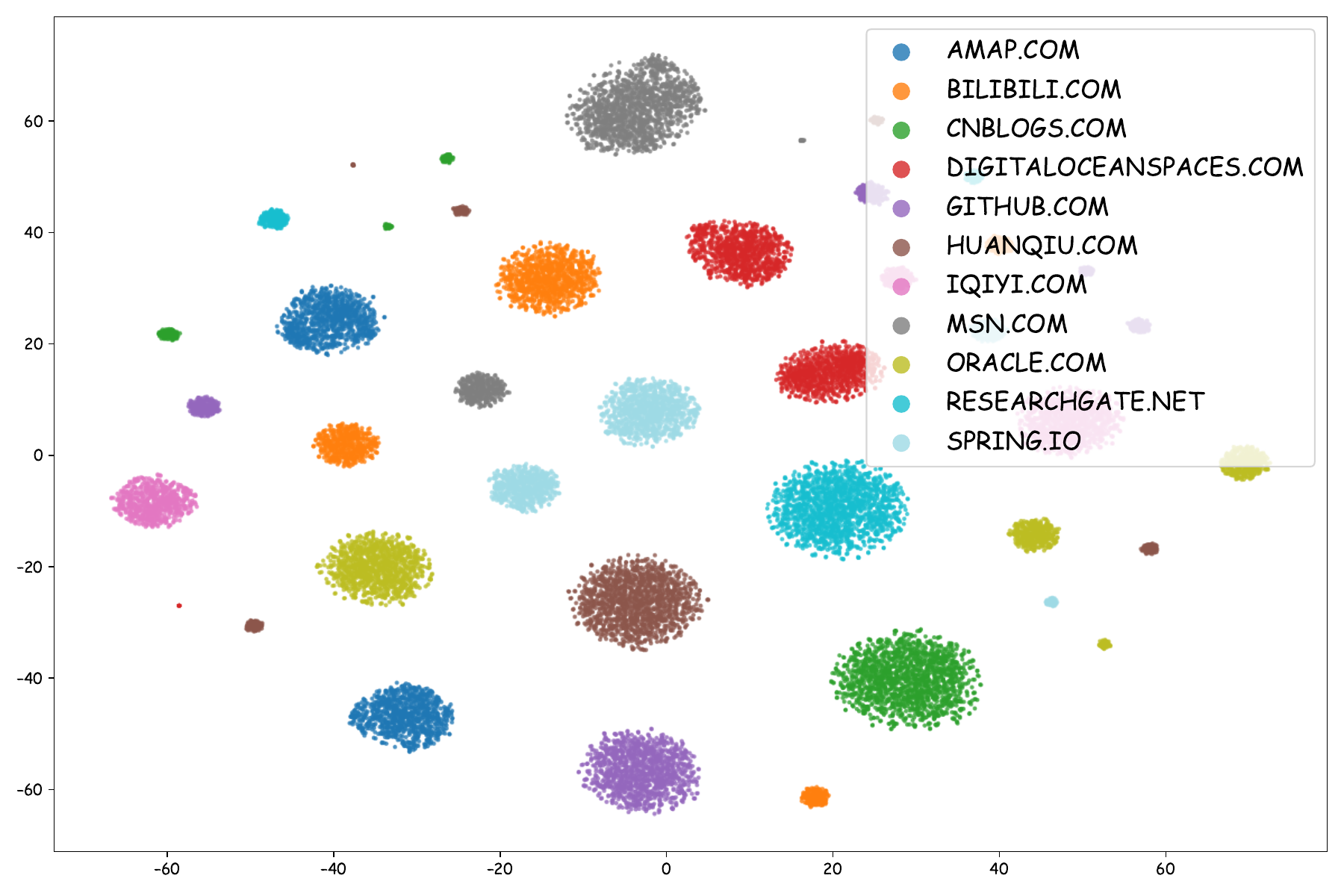} \\
		\bottomrule
	\end{tabular}
	\caption{Comparison of Reverse Denoising Processes in IDS and TLS Datasets (UG: unguided, G: guided).}
	\Description{A figure comparing the reverse denoising processes of TDDM with or without conditional guidance across five timesteps (900, 600, 400, 200, 0). The top row is a header indicating the timestep values. The middle row shows a sequence of t‑SNE visualizations where the generated samples (colored points) gradually converge towards the real data distribution (gray background points) without guidance, exhibiting more scattered and ambiguous intermediate states. The bottom row shows the guided process, where the samples converge more compactly and consistently towards the real class centroids, demonstrating the effectiveness of the centroid‑guided compact noise sampling mechanism.}
	\label{fig:denoise_step}
\end{figure*}

To thoroughly investigate the impact of the condition-guided compact noise sampling mechanism on the performance of TDDM, we design a targeted ablation study. In the experiment, TDDM first undergoes a forward diffusion process of \( T = 1000 \) timesteps, followed by 500 epochs of training for the noise prediction network. During the sample generation phase, we compare the distributional characteristics of the generated data under 2 settings: with or without condition-guided compact noise sampling. The denoising process is visualized using t-SNE dimensionality reduction \cite{t-SNE}. The results are presented in Figure~\ref{fig:denoise_step} (and in Appendix Figure~\ref{fig:denoise_step_add}). {\itshape It is worth noting that t-SNE transforms local similarities among data points in high-dimensional space into probability distributions, effectively revealing cluster structures. However, t-SNE cannot accurately reflect the true distances between clusters, and the visualization outcome is inherently stochastic.}

Observing the experimental results reveals that, in the unguided reverse denoising process, the noise predicted by the model exhibits ambiguity, which ultimately leads to inferior quality of the final generated samples. After the introduction of condition-guided compact noise sampling, the model is able to effectively steer the reverse denoising process. As the timestep decreases, the generated samples progressively converge toward the distribution of authentic samples. Considering the configured noise schedule parameters (\( \beta_{\text{start}} = 10^{-4}, \; \beta_{\text{end}} = 0.02 \)), the guidance strength is relatively weak during the initial stages of denoising; as the process progresses toward the final stages, the guidance strength intensifies significantly. From a theoretical perspective, by embedding class information into the noise sampling procedure, the conditional guidance mechanism effectively constrains the output space of the denoising generation, thereby reducing the risk of mode collapse while simultaneously enhancing the diversity of the generated samples.

\subsection{Ablation Analysis of the Memory Model}
\label{5.5}

\begin{figure}[htbp]
\centering
    \begin{subfigure}[b]{0.49\columnwidth}
    \centering
    \includegraphics[width=\linewidth]{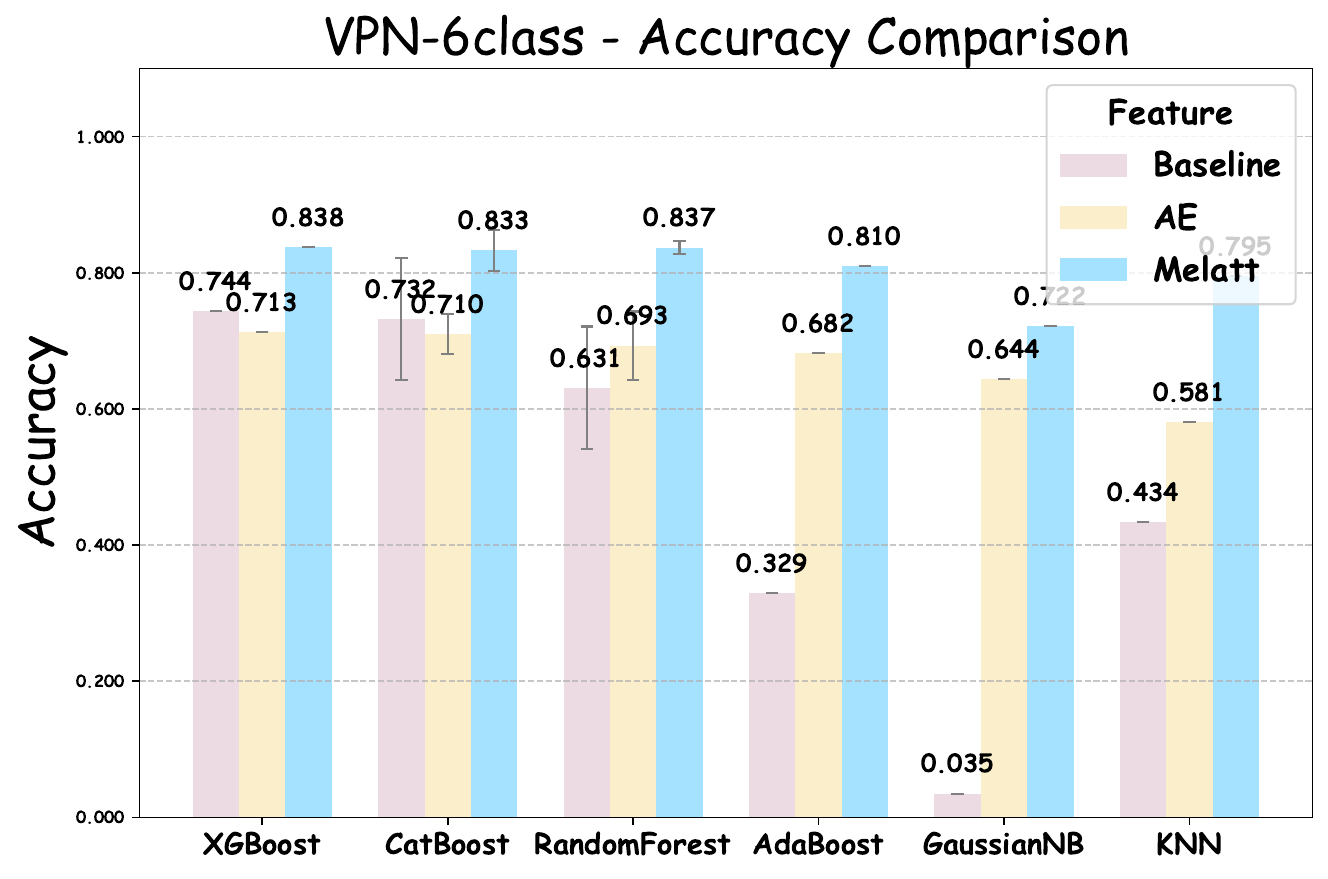}
    \end{subfigure}
\hfill
    \begin{subfigure}[b]{0.49\columnwidth}
    \centering
    \includegraphics[width=\linewidth]{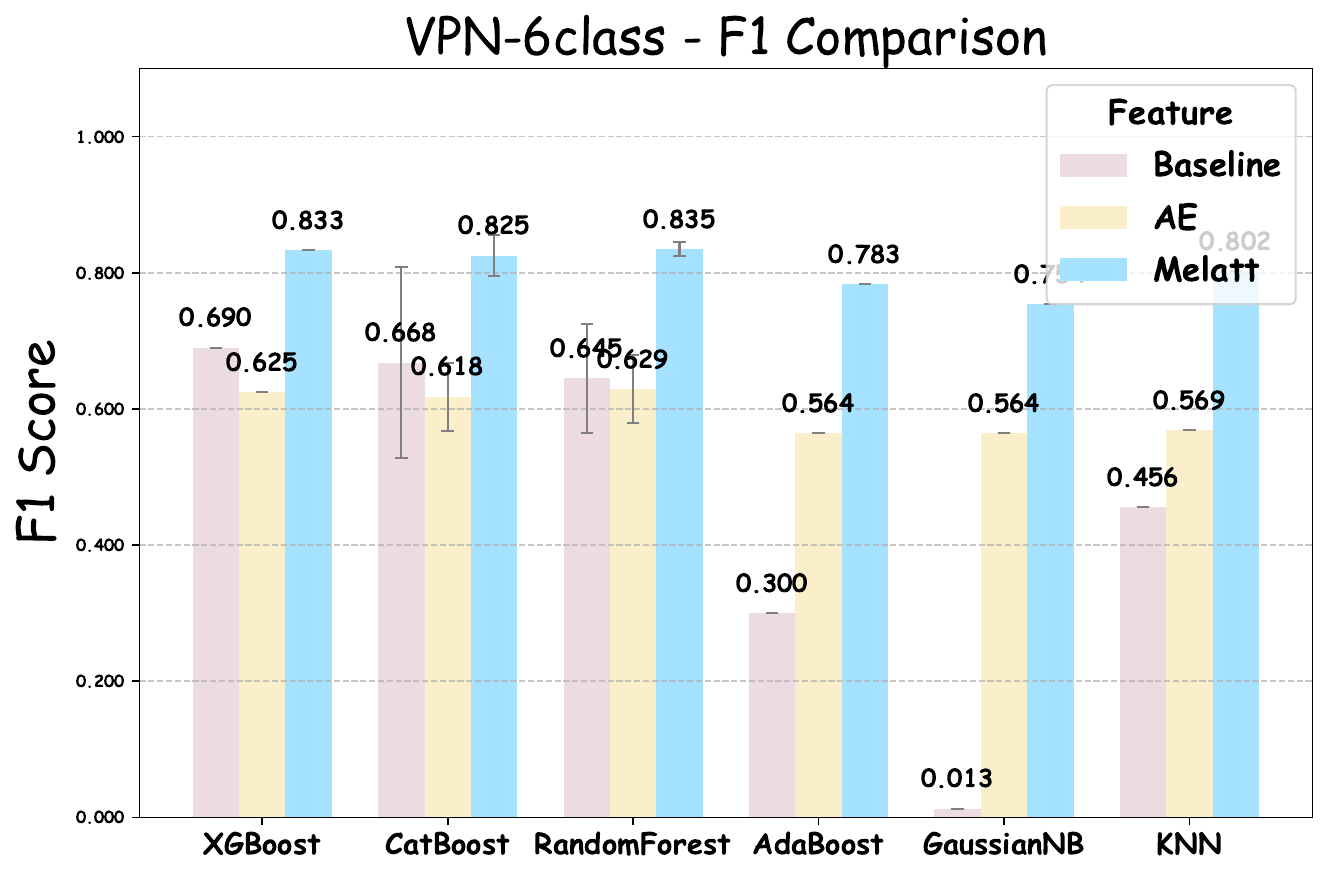}
    \end{subfigure}
\hfill \\
    \begin{subfigure}[b]{0.49\columnwidth}
    \centering
    \includegraphics[width=\linewidth]{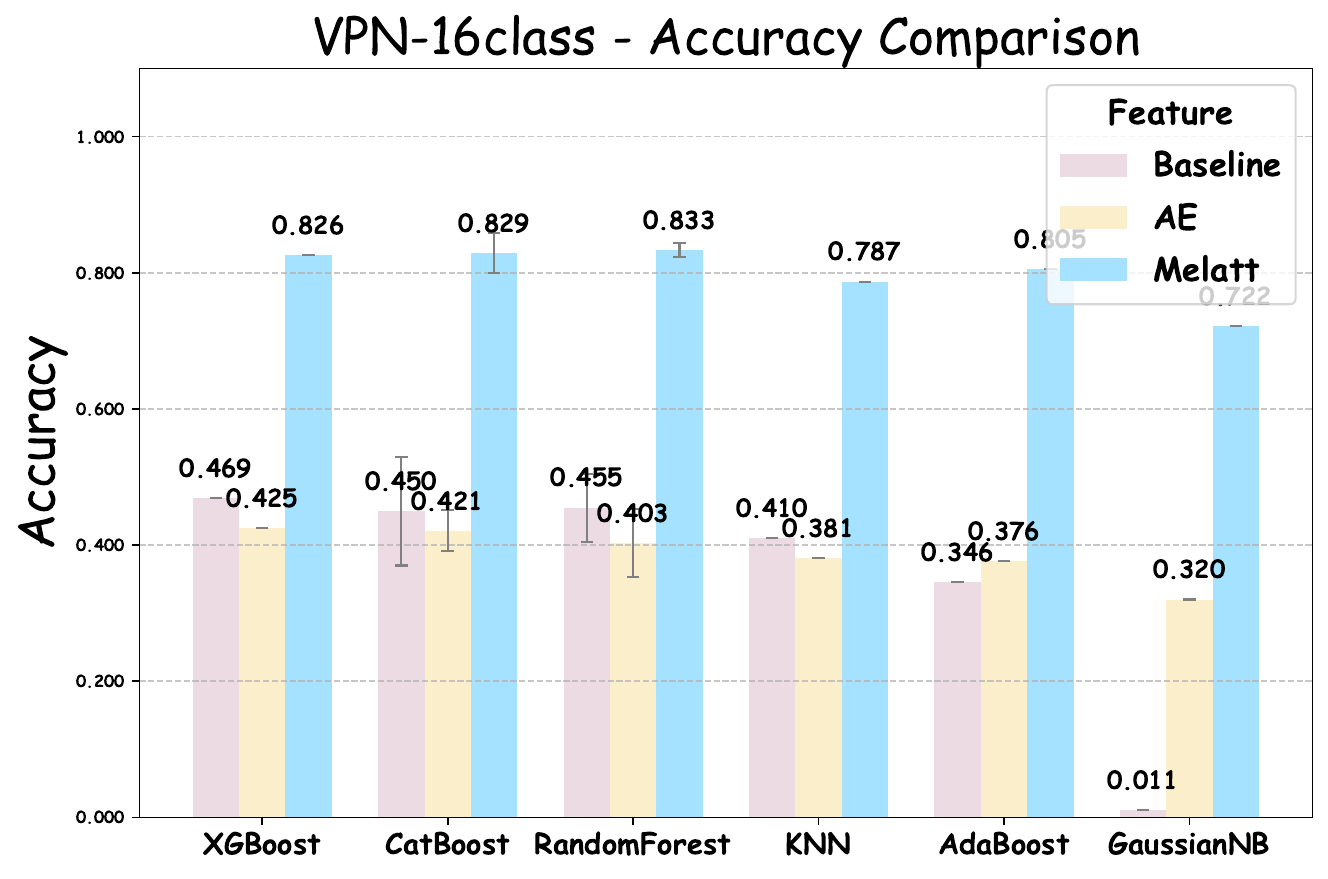}
    \end{subfigure}
\hfill
    \begin{subfigure}[b]{0.49\columnwidth}
    \centering
    \includegraphics[width=\linewidth]{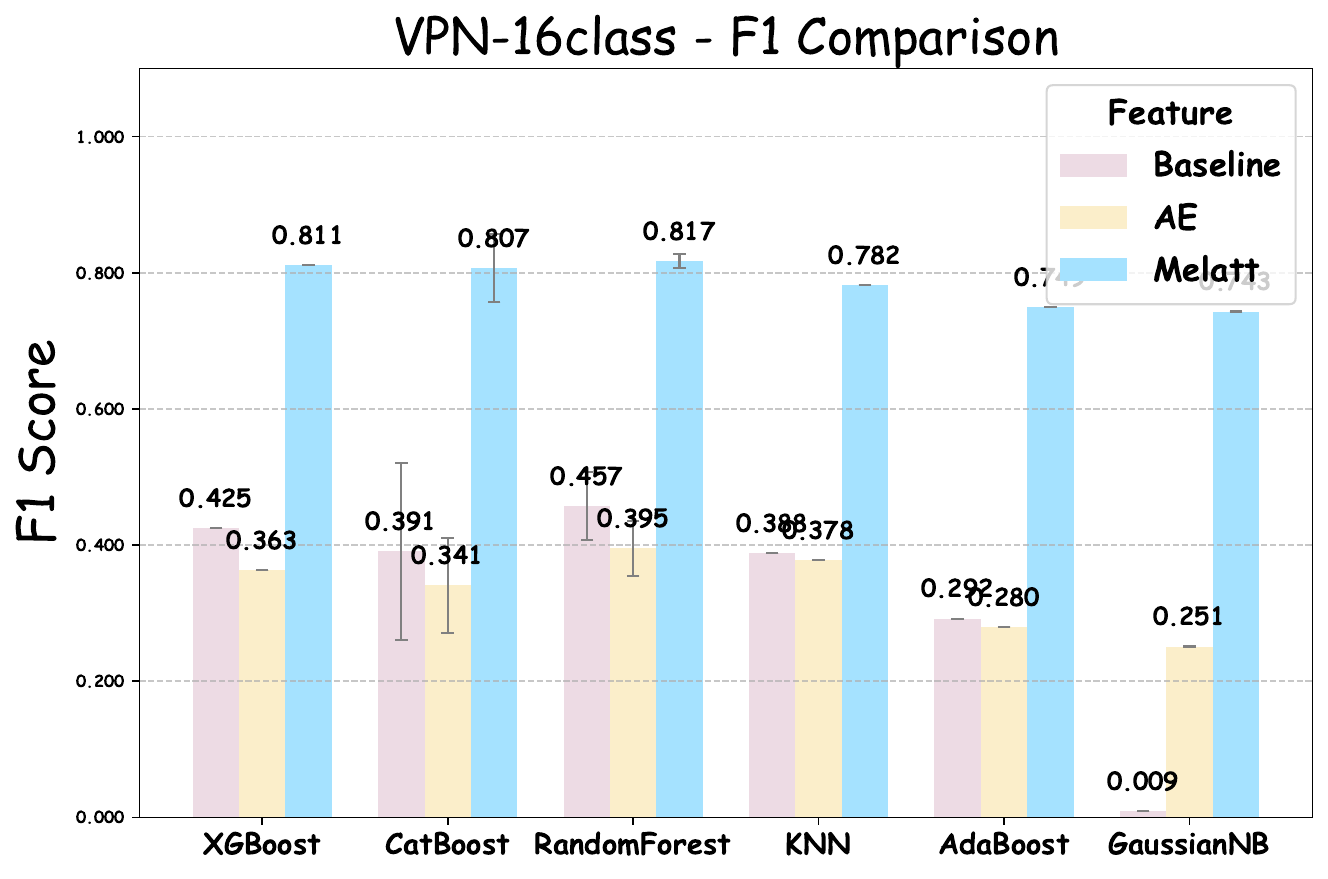}
    \end{subfigure}
{
\caption{Feature Extraction Performance in 6 Classifiers.}
\Description{Ablation study comparing three feature extraction strategies across six classifiers on VPN-6class and VPN-16class tasks. The four subfigures present, respectively, the accuracy on VPN-6class, the F1-score on VPN-6class, the accuracy on VPN-16class, and the F1-score on VPN-16class. Within each subfigure, the horizontal axis enumerates the classifiers (XGBoost, CatBoost, RandomForest, AdaBoost, KNN), and the vertical axis represents the performance metric. The three grouped bars correspond to the original features (blue), Autoencoder features (orange), and Melatt features (green).
A two-by-two grid of bar charts presenting the ablation results. The top-left chart shows the accuracy of XGBoost, CatBoost, RandomForest, AdaBoost, and KNN on VPN-6class using three different feature representations: original raw features, features extracted by a standard Autoencoder (AE), and features extracted by the proposed Melatt model. The top-right chart is structured identically but reports the F1-score for the same task. The bottom-left and bottom-right charts follow the same format for the VPN-16class task, reporting accuracy and F1-score respectively. Across all charts, the green bars representing Melatt consistently achieve the highest values for XGBoost, CatBoost, and RandomForest, while the improvements for KNN and AdaBoost are less pronounced. Error bars indicate the standard deviation over five repeated runs.}
\label{fig:ablation_classifiers}
}
\end{figure}

To validate the actual contribution of the memory model Melatt in feature pre-training and downstream classification, we design and conduct targeted ablation experiments. By comparing the generalization performance of different feature extraction strategies across multiple classifiers, we analyze the advantages of Melatt over a conventional Autoencoder (AE) \cite{AE-1,AE-2} and the original raw traffic features. The experiments are conducted on the 6-class and 16-class classification tasks of the ISCX-VPN-2016 dataset, comparing six classifiers: XGBoost \cite{XGBoost-2}, CatBoost \cite{CatBoost}, RandomForest \cite{Random-Forests}, AdaBoost \cite{AdaBoost}, GaussianNB \cite{GaussianNB}, and KNN \cite{KNN}. All comparison schemes share a unified experimental configuration, with the feature extraction method serving as the sole independent variable. Each experimental configuration is repeated independently five times, and the mean and standard deviation are reported to eliminate the influence of random fluctuations. The results are illustrated in Figure~\ref{fig:ablation_classifiers}, along with the error bars of the repeated trials.

Baseline does not incorporate any deep learning feature extraction module and retains only preprocessing. AE denotes a MLP autoencoder without the memory module. It extracts the latent vector and computes reconstruction errors (i.e., the squared differences between the input and reconstructed). These errors, concatenated with the latent vector, serve as the classification features. Melatt refers to our full model. Across all 6 classifiers, the features extracted by Melatt achieve the highest AC and F1, with notable improvements. Although AE adopts a pre-training paradigm based on reconstruction error, its lack of a content-based addressing mechanism for the memory bank causes its performance on certain classifiers (e.g., XGBoost, GaussianNB) to fall even below that of the Baseline. Features derived purely from reconstruction error may introduce additional noise or redundant information during the encoding and decoding processes. In contrast, by virtue of the memory module, Melatt can effectively capture and match the deep behavioral characteristics and anomalous patterns of traffic, thereby supplying more discriminative feature representations to downstream classifiers.

Further analysis of the performance across different classifiers reveals that the performance trends of the three feature extraction strategies remain consistent across all 6 classifiers. Gradient boosting tree models (XGBoost, CatBoost) exhibit the best overall performance. RandomForest is a parallel ensemble learning method and slightly underperforms gradient boosting trees. This is primarily attributable to its voting-based aggregation mechanism, whose residual fitting precision is inferior to serial iterative optimization of gradient boosting. AdaBoost, as an earlier boosting algorithm, delivers relatively limited performance on complex traffic classification tasks; its high sensitivity to noisy samples renders the model prone to overfitting when confronted with traffic data characterized by feature overlap and outliers. The probabilistically-driven GaussianNB and the distance-based KNN perform poorly, indicating the inherent limitations of linear assumptions and distance metrics in high-dimensional traffic feature spaces. In addition, the performance of CatBoost and RandomForest exhibits minor fluctuations across the five repeated runs. {\itshape It is worth noting that, although our model was not specifically designed with a sophisticated classifier, the insights gained from this experiment provide valuable directions for the design of downstream classifiers in future work.}

\begin{figure}[htbp]
	\centering
	\includegraphics[width=0.9\linewidth]{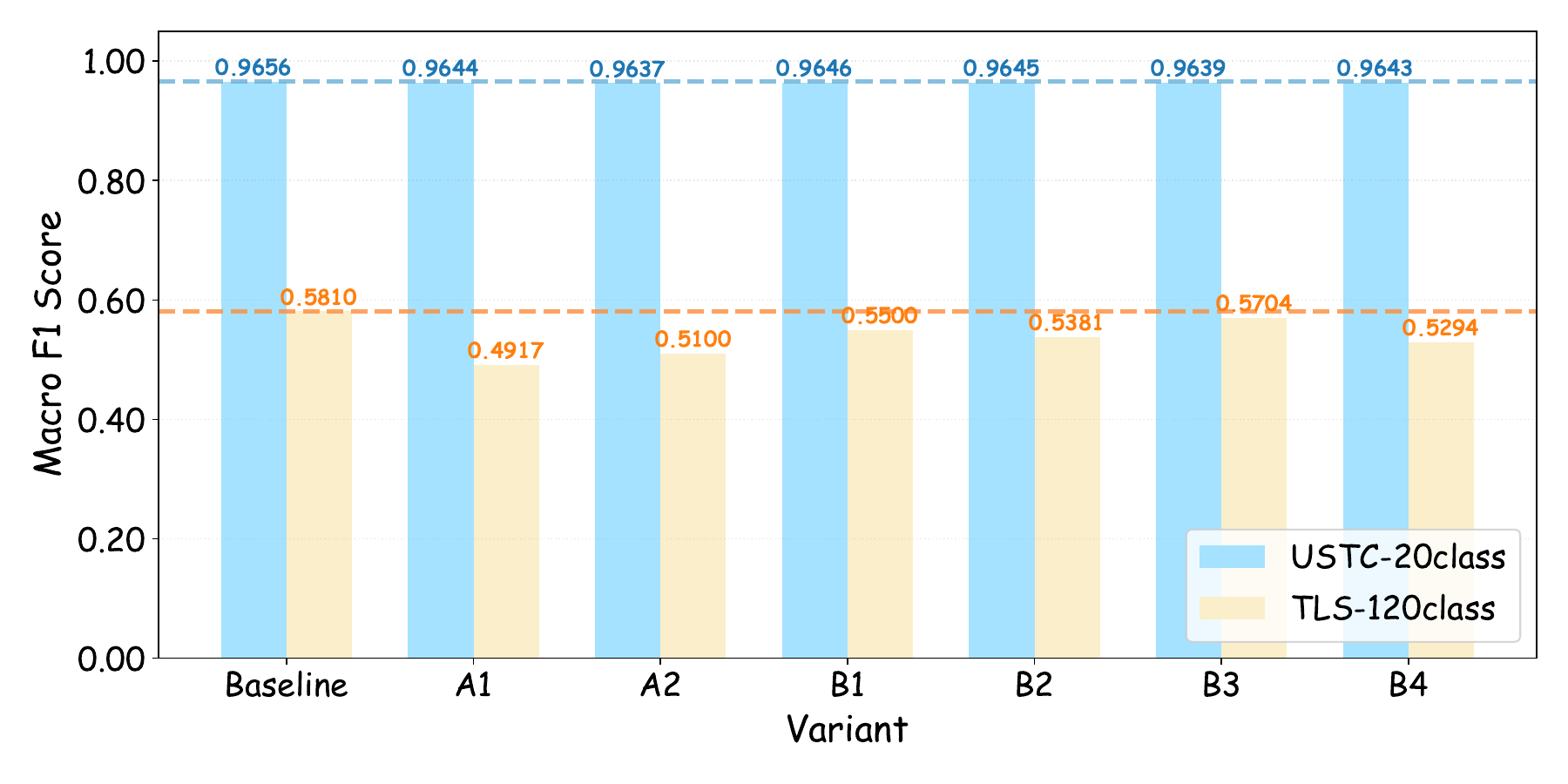}
	\caption{Component-level Ablation Study of Melatt.}
	\label{fig:melatt_ablation}
\end{figure}

To precisely identify the contribution of each core component in Melatt, we conduct additional ablation experiments on two representative tasks: USTC-20class (relatively balanced, Macro F1 = 96.56\%) and TLS-120class (highly imbalanced fine-grained task, Macro F1 = 58.10\%). We evaluate six variants: (A1) replacing CG-LSTM with standard LSTM, (A2) removing the multi-head cross-attention module, (B1) removing the compactness loss, (B2) removing the diversity loss, (B3) removing the entropy loss, and (B4) retaining only the reconstruction loss (i.e., removing the other three losses). Figure~\ref{fig:melatt_ablation} presents the F1 scores of all variants on both tasks.
﻿

The ablation results reveal a clear pattern: the effectiveness of each Melatt component is strongly correlated with task difficulty. On the relatively simple USTC-20class task (96.56\% Macro F1), removing any single component causes only marginal performance degradation (F1 drop $< 0.2\%$). In contrast, on the challenging TLS-120class task (58.10\% Macro F1), the contributions of all components become substantially more pronounced. This component-wise effectiveness aligns with the inspiration of our framework: in simple scenarios, the memory module acts as a robust yet non-critical auxiliary regulator, whereas in challenging fine-grained classification tasks, each component becomes indispensable for maintaining high-fidelity prototype representations.

\subsection{Model Efficiency}
\label{5.6}

Feature extraction serves as the critical link between data preprocessing and downstream classification, and its computational efficiency directly determines the overall system throughput and end-to-end detection latency. While the ablation study has confirmed the significant performance advantages of Melatt in classification, we further quantify the computational overhead introduced by the memory module through an efficiency comparison experiment, evaluating the time performance of Melatt across different traffic classification scenarios. The experiment encompasses eight classification tasks of varying scale and complexity across the four datasets. All experiments are conducted on an edge computing platform equipped with NVIDIA GeForce RTX 3050 Laptop GPU. To eliminate the influence of system caching and initialization overhead, five warm-up runs are performed first, followed by ten consecutive test runs, from which the arithmetic mean is reported.

\begin{figure}[tbp]
	\centering
	\begin{subfigure}[b]{0.48\columnwidth}
		\centering
		\includegraphics[width=\linewidth]{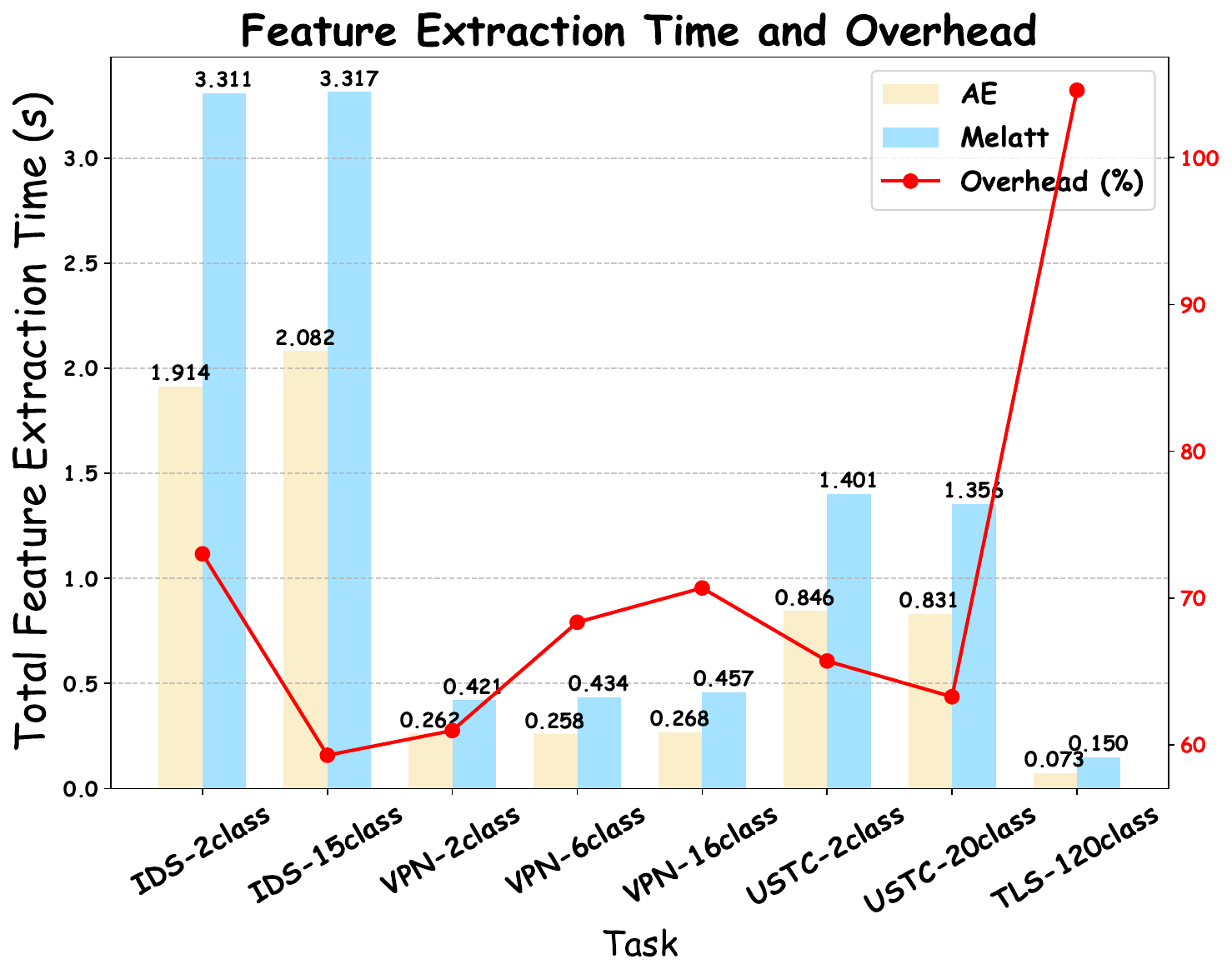}
		\Description{Bar chart comparing the total feature extraction time of a standard Autoencoder (AE) and Melatt across eight tasks. The x-axis lists the tasks, and the y-axis shows the time in seconds. The overhead introduced by the memory module of Melatt is indicated as a percentage above each pair of bars.}
	\end{subfigure}
	\hfill
	\begin{subfigure}[b]{0.48\columnwidth}
		\centering
		\includegraphics[width=\linewidth]{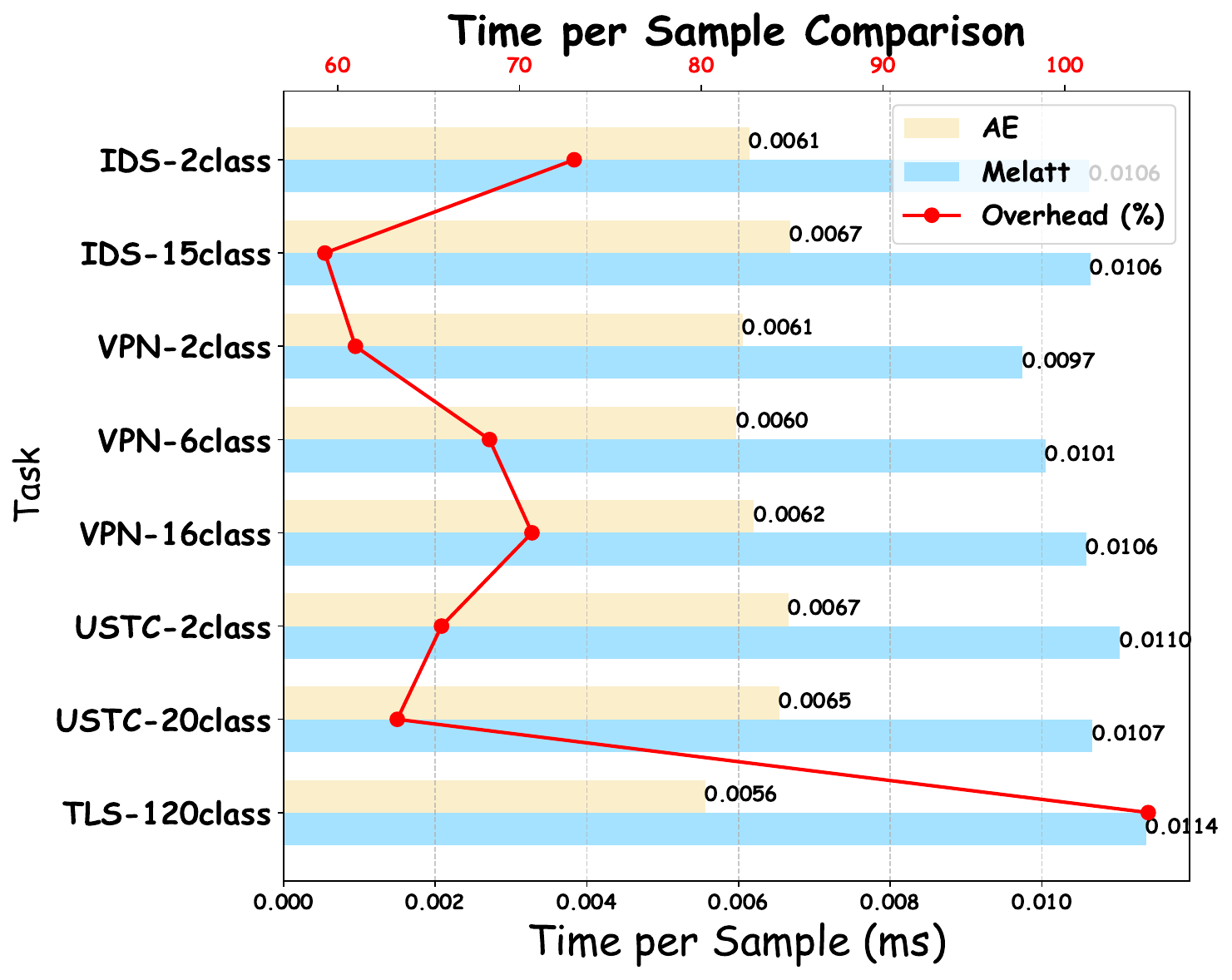}
		\Description{Bar chart comparing the per-sample feature extraction time of AE and Melatt across the same eight tasks. The y-axis indicates the time in seconds per sample, ranging from 0.0 to approximately 1.0. The per-sample time of Melatt remains consistently low across tasks of varying class counts, demonstrating that its computational complexity is largely decoupled from the number of classes.}
	\end{subfigure}
	{
		\caption{Efficiency Evaluation of Feature Extraction.}
		\Description{A figure composed of two bar charts evaluating the computational efficiency of feature extraction. The left chart displays the total extraction time for AE and Melatt on eight tasks (e.g., IDS-2class, IDS-15class, VPN-2class, VPN-6class, VPN-16class, USTC-2class, USTC-20class, TLS-120class). The overhead introduced by Melatt's memory module is labeled as a percentage above the bars. The right chart shows the per-sample processing time, where Melatt's time remains stable and low despite increasing task complexity.}
		\label{fig:efficiency}
	}
\end{figure}

The experimental results are presented in Figure~\ref{fig:efficiency}. The additional overhead introduced by the memory module ranges between 50\% and 110\%. For tasks with smaller sample sizes but higher class complexity (e.g., TLS-120class), the overhead reaches 104.59\%; for all other tasks, it remains below 80\%. This indicates that the memory mechanism incurs a relatively higher computational cost when handling complex tasks, yet the overhead remains within an acceptable range. Notably, task complexity exerts a minimal impact on the per-sample processing time. From the 2-class classification tasks to the 120-class task, the per-sample processing time increases by merely approximately 0.005 ms. This demonstrates that the computational complexity of Melatt is largely decoupled from the number of classes, exhibiting excellent scalability.

\begin{table}[htbp]
	\centering
	{
		\caption{Inference Latency Comparison.}
		\label{tab:efficiency_comparison}
		\begin{tabular}{l|c|c}
			\toprule
			\textbf{Model} & \textbf{Architecture} & \textbf{Latency (ms)} \\
			\midrule
			Melatt + XGBoost & CG-LSTM + Memory & $<12$ \\
			NetMamba & Mamba & $>15$ \\
			ET-BERT & BERT & $>50$  \\
			TrafficFormer & BERT & $>50$ \\
			Pcap-Encoder & T5 & $>180$ \\
			netFound & BERT & $>10^4$  \\
			\bottomrule
		\end{tabular}
	}
\end{table}

To evaluate the practical inference efficiency of the proposed Melatt framework, we conduct a quantitative latency comparison with the SOTA traffic classification models, with detailed results summarized in Table~\ref{tab:efficiency_comparison}. The Melatt framework integrated with XGBoost achieves optimal inference efficiency, with a per-sample feature extraction latency lower than 12 ms. This prominent efficiency advantage benefits from the framework’s decoupled design, which employs a lightweight XGBoost classifier on fixed pre-trained representations. This design avoids the necessity of end-to-end full network execution for each inference, fundamentally reducing inference overhead and enabling Melatt to deliver far superior practical deployment efficiency compared with all competing SOTA models.

\section{Discussion}
\label{6}

The core insight underpinning the proposed TDDM-Melatt is that the generalization bottleneck in encrypted traffic classification stems not from insufficient model capacity, but rather from a fundamental misalignment of learning objectives. Existing research has predominantly pursued performance gains in specific datasets while overlooking the dynamic variations in feature distributions within real-world network environments. Our experiments compellingly demonstrate that once spurious features such as IP and Port are rigorously removed, the performance of SOTA models suffers a precipitous decline. This phenomenon reveals a pervasive ``benchmark overfitting'' problem in the current research landscape. Through its memory-decoupled architecture, TDDM-Melatt fundamentally shifts the learning paradigm of traffic classification models from ``fitting the training data distribution'' to ``learning universal behavioral prototypes of traffic.''

Regarding data augmentation, our study indicates that generic generative models cannot be directly transplanted to network traffic data. The reason TDDM outperforms GANs and other diffusion-based models lies critically in the fact that we did not blindly transfer architectures designed for image generation. Instead, we proceeded from the intrinsic characteristics of traffic data and devised a suite of targeted strategies, including feature selection, class-wise training, centroid-guided sampling, and nearest-neighbor padding. Our experiments also reveal that data augmentation is not a panacea. When performance on the majority classes approaches saturation, merely increasing minority-class samples yields limited improvement in overall performance. This suggests that future research should not only focus on data augmentation techniques themselves, but should also explore more effective imbalance-aware loss functions and training strategies. Furthermore, we observe that the effectiveness of TDDM is not uniformly transferable across different base models.  This disparity suggests that TDDM's compatibility depends on the base model's representation learning paradigm. We discuss this limitation in detail in Appendix~\ref{Discussion}.

Although TDDM-Melatt has achieved outstanding performance on public datasets, we are fully aware that the complexity of real-world network environments far exceeds that of any laboratory dataset. Future work should address issues such as the inability of memory to adapt to dynamic traffic patterns and the insufficiency of incremental learning capabilities. We believe that interpretability and generalizability will be the defining characteristics of next-generation AI systems for cybersecurity. The decoupled memory architecture and the spurious-correlation-free training paradigm proposed in this paper lay the foundation for building such systems.

\section{Related Work}

Encrypted traffic classification constitutes a highly active research area in contemporary network security. In recent years, deep learning methods have emerged as the predominant paradigm. CNNs \cite{CNNs-2,CNNs-1} have been employed to capture local packet length patterns, while LSTM and GRU \cite{LSTM-GRU-1} have been applied to model temporal dependencies. ET-BERT \cite{ET-BERT-TLS120} draws inspiration from pre-trained language models. YaTC \cite{YaTC} adopts the Transformer, leveraging self-attention mechanisms to capture long-range dependencies within traffic flows. NetMamba \cite{NetMamba} introduces state space models to preserve the capacity for long-sequence modeling. However, explicit identifiers commonly retained in public datasets exhibit spurious correlations with traffic labels \cite{Pcap-Encoder,SoK,spurious-correlation-3}. This phenomenon causes the generalization capability of these models to deteriorate severely when deployed in real-world network environments.

Data augmentation techniques have been extensively explored as a means to address the problem of class imbalance. Conventional methods such as SMOTE \cite{SMOTE} and ADASYN \cite{ADASYN} generate synthetic samples via linear interpolation in the feature space, yet they struggle to capture the complex temporal dependencies and nonlinear characteristics inherent in traffic data. The generated samples often lack semantic coherence and may even introduce noise, thereby degrading model performance. In the realm of deep generative models, GANs \cite{GAN} and their variants, including WGAN-GP \cite{WGAN-GP} and ACGAN \cite{ACGAN}, have been widely investigated for traffic generation \cite{GraphCWGAN-GP,E-WACGAN}. Nevertheless, these methods suffer from intrinsic deficiencies such as training instability, mode collapse, and vanishing gradients \cite{GMA-SAWGAN-GP}. More recently, diffusion models have begun to be applied to traffic data augmentation \cite{NetDiff}. NetDiffus \cite{Netdiffus} maps traffic data into two-dimensional grayscale images and then directly adopts an image-oriented diffusion architecture. LRT-DDPM \cite{LRT-DDPM} attempts to leverage diffusion models for encrypted traffic classification. However, both of these methods directly transplant generic generative model architectures without adequately considering the distinctive characteristics of traffic data---namely, high dimensionality, sparsity, strong feature interdependencies, and discrete structured nature \cite{data-inba}. The generated samples consequently suffer from issues such as feature distribution drift and semantic information deficiency \cite{data-distribu-shift-1,data-distribu-shift-2}, which may, in turn, introduce noise that interferes with model training.

\section{Conclusion}

In this paper, to address the insufficient generalization capability caused by spurious correlations and the class imbalance problem in real-world encrypted traffic classification, we propose TDDM-Melatt, a decoupled memory and diffusion model framework for generalizable encrypted traffic classification. The modular decoupled memory representation model based on the CG-LSTM separates feature extraction from knowledge storage. A pre-training and inference paradigm free of spurious correlations is established to suppress shortcut learning, and a denoising diffusion model tailored to the structural properties of traffic data is designed to alleviate the class imbalance issue. Experiments on four public datasets demonstrate that TDDM-Melatt achieves superior performance across multiple classification tasks. The TDDM data augmentation method yields up to a 9.98\% improvement in F1-score, and ablation studies validate the effectiveness of each core component. This paper provides a new paradigm for encrypted traffic classification in real network environments, characterized by high accuracy, strong generalization, and rapid adaptability.

\bibliographystyle{ACM-Reference-Format}
\bibliography{myref}

\appendix

\section{Preprocessed Statistical Features}
\label{statistical features}

\begin{figure*}[htbp]
	\centering
	\begin{minipage}{0.9\textwidth}
		\centering
		\includegraphics[width=\linewidth]{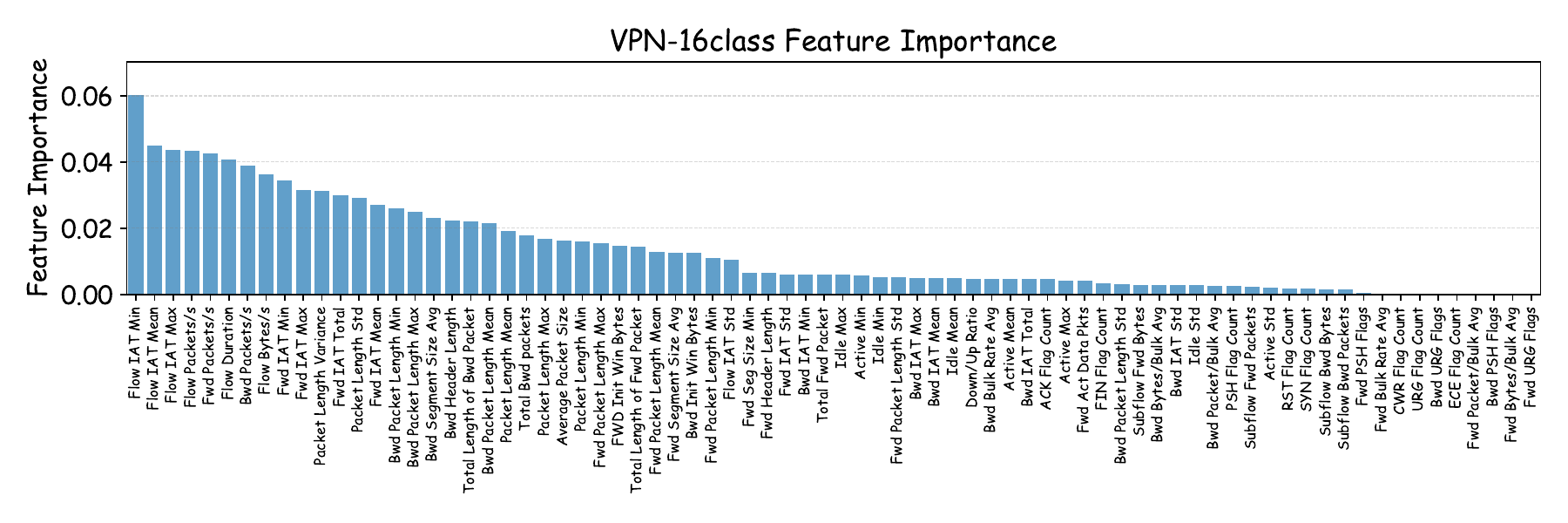}
	\end{minipage}
	\begin{minipage}{0.9\textwidth}
		\centering
		\includegraphics[width=\linewidth]{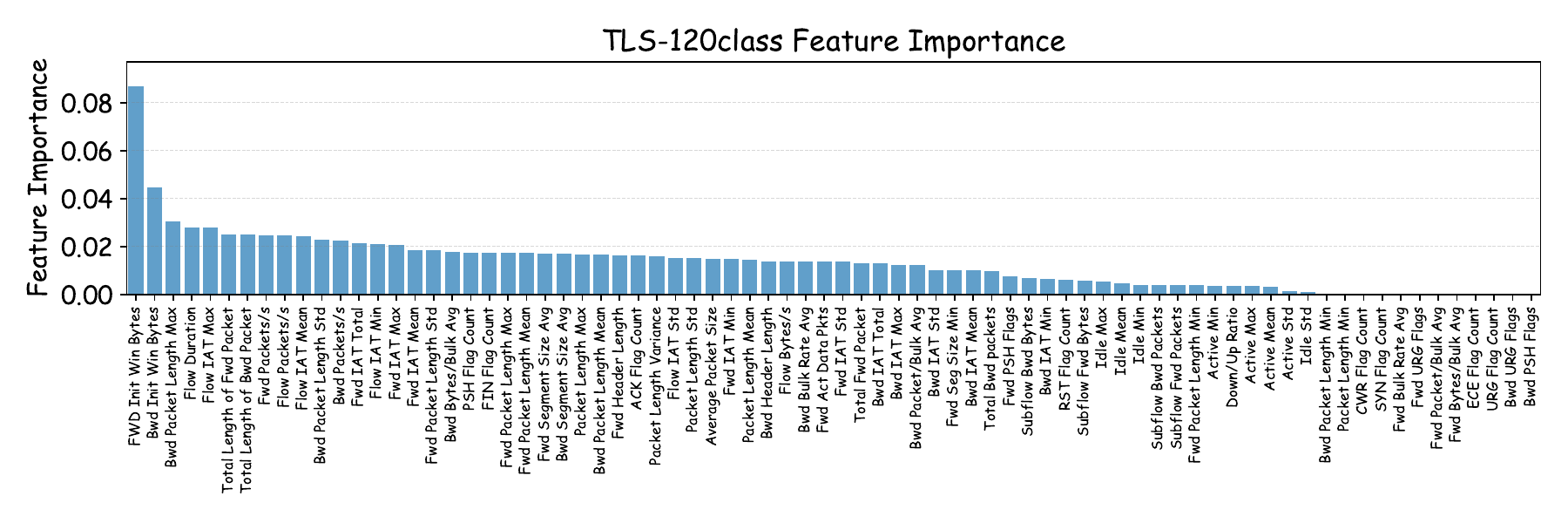}
	\end{minipage}
	\begin{minipage}{0.9\textwidth}
		\centering
		\includegraphics[width=\linewidth]{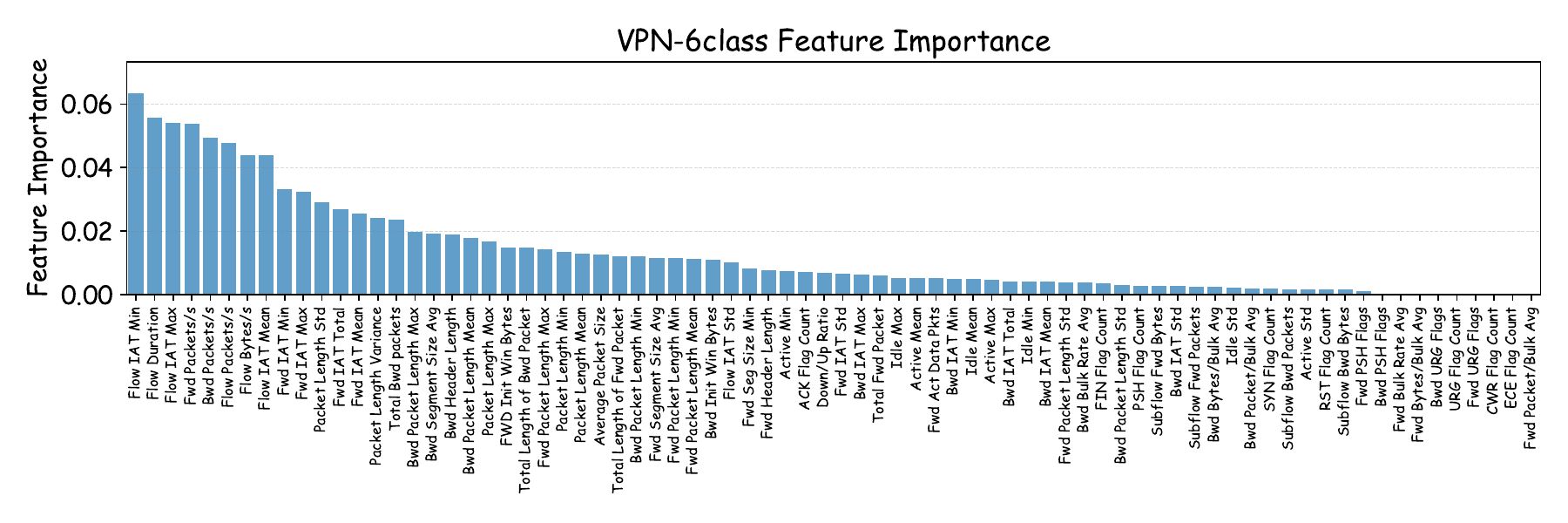}
	\end{minipage}
	\begin{minipage}{0.9\textwidth}
		\centering
		\includegraphics[width=\linewidth]{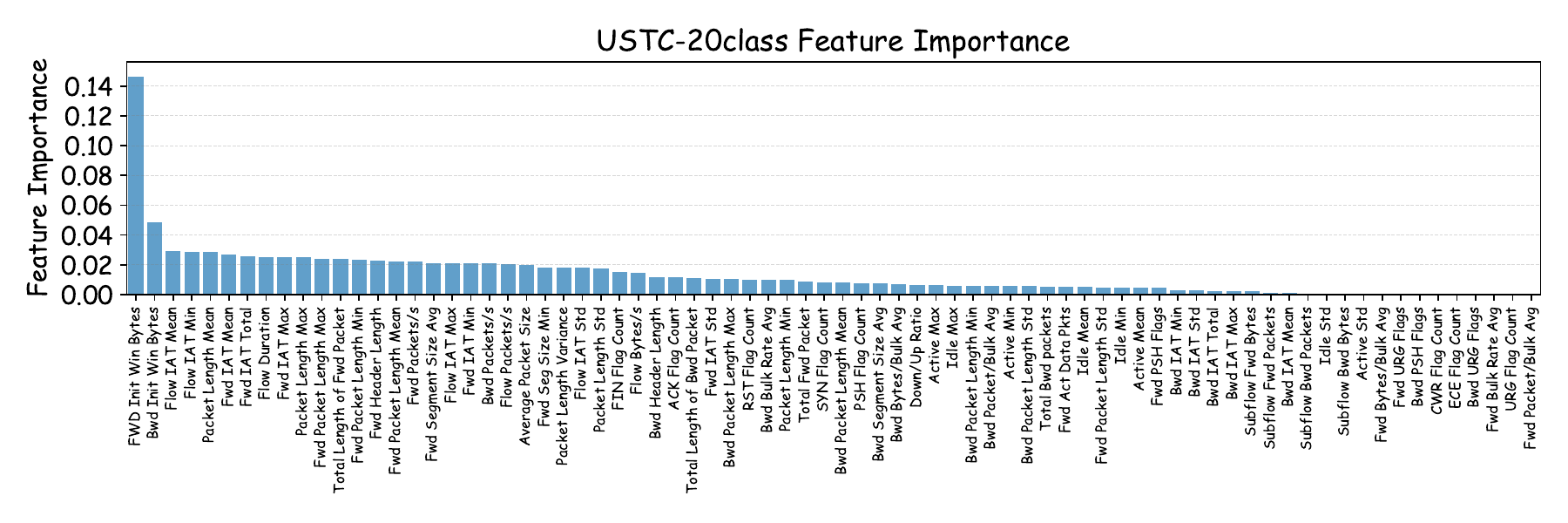}
	\end{minipage}
	\caption{Feature Importance Bar Charts for TDDM-Using Tasks.}
	\label{fig:bar_all_remaining}
\end{figure*}

This appendix provides the complete set of preprocessed features, as listed in Table \ref{tab:full-feature-list}. It also presents the full feature importance bar charts for 4 classification tasks (VPN-16class, TLS-120class, VPN-6class, and USTC-20class) that were generated using the TDDM, as shown in Figure \ref{fig:bar_all_remaining}.

\begin{table}[htbp]
	\centering
	\caption{Complete List of 76 Preprocessed Statistical Features.}
	\label{tab:full-feature-list}
	\small
	\begin{tabular}{|>{\centering\arraybackslash}m{0.28\linewidth}|m{0.68\linewidth}|}
		\hline
		\textbf{Category} & \multicolumn{1}{c|}{\textbf{Features}} \\
		\hline
		Length / Size \newline (16 features)
		& Total Length of Fwd Packet, Total Length of Bwd Packet, Fwd Packet Length Max/Min/Mean/Std, Bwd Packet Length Max/Min/Mean/Std, Packet Length Min/Max/Mean/Std/Variance, Average Packet Size, Fwd Segment Size Avg, Bwd Segment Size Avg, Fwd Seg Size Min \\
		\hline
		Temporal / IAT \newline (20 features)
		& Flow Duration, Flow IAT Mean/Std/Max/Min, Fwd IAT Total/Mean/Std/Max/Min, Bwd IAT Total/Mean/Std/Max/Min, Active Mean/Std/Max/Min, Idle Mean/Std/Max/Min \\
		\hline
		Packet / Byte Count \newline (12 features)
		& Total Fwd Packet, Total Bwd packets, Subflow Fwd Packets, Subflow Bwd Packets, Subflow Fwd Bytes, Subflow Bwd Bytes, Fwd Act Data Pkts, Down/Up Ratio, Fwd Header Length, Bwd Header Length \\
		\hline
		Rate Features \newline (10 features)
		& Flow Bytes/s, Flow Packets/s, Fwd Packets/s, Bwd Packets/s, Fwd Bulk Rate Avg, Bwd Bulk Rate Avg, Fwd Bytes/Bulk Avg, Fwd Packet/Bulk Avg, Bwd Bytes/Bulk Avg, Bwd Packet/Bulk Avg \\
		\hline
		Flag / Window Init \newline (18 features)
		& FIN Flag Count, SYN Flag Count, RST Flag Count, PSH Flag Count, ACK Flag Count, URG Flag Count, CWR Flag Count, ECE Flag Count, Fwd PSH Flags, Bwd PSH Flags, Fwd URG Flags, Bwd URG Flags, FWD Init Win Bytes, Bwd Init Win Bytes \\
		\hline
	\end{tabular}
\end{table}

\begin{figure}[htbp]
	\centering
	\begin{subfigure}[b]{0.49\columnwidth}
		\centering
		\includegraphics[width=\linewidth]{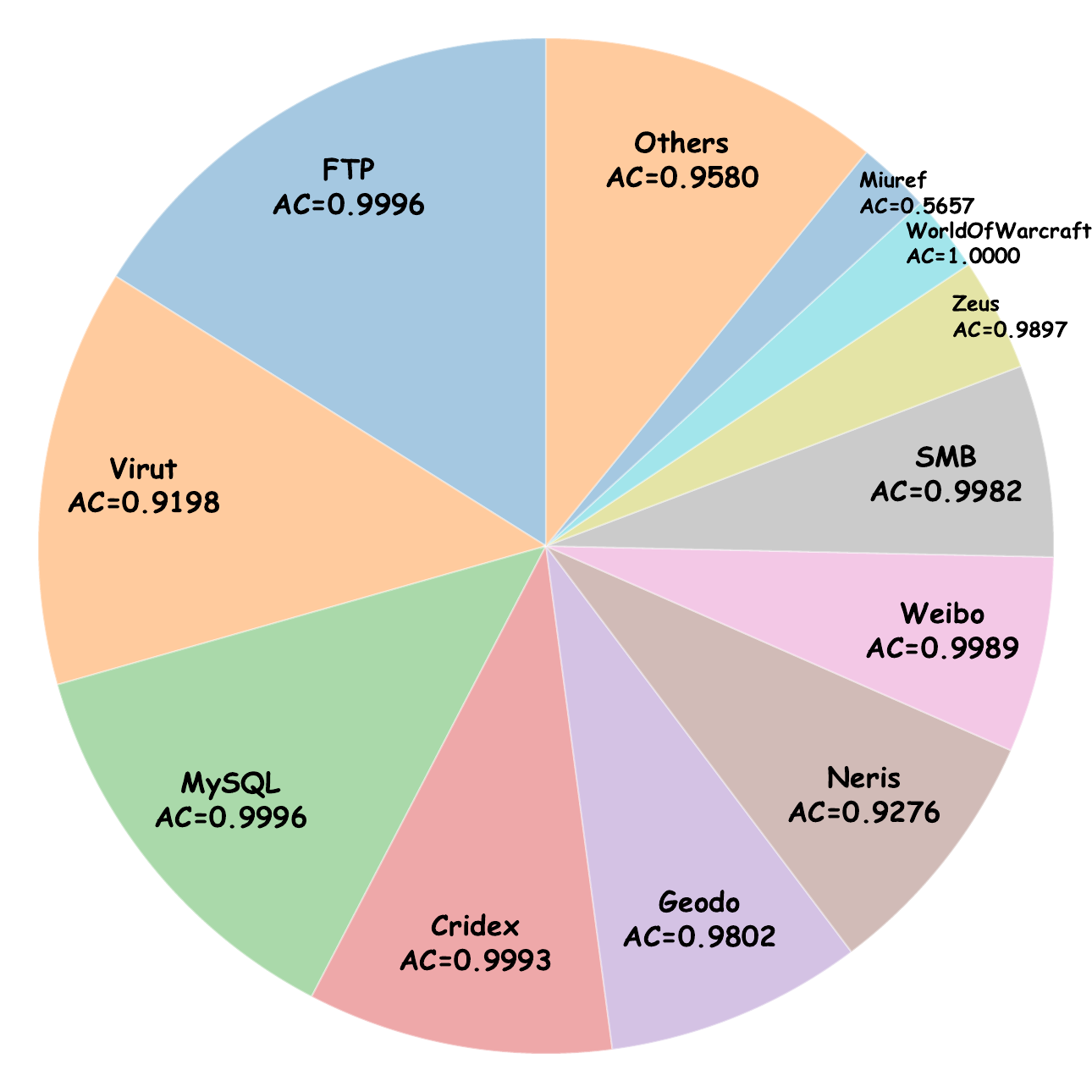}
		\caption{Original Accuracy}
		\Description{The original per-class accuracy of Melatt on USTC-20class without data augmentation.}
	\end{subfigure}
	\hfill
	\begin{subfigure}[b]{0.49\columnwidth}
		\centering
		\includegraphics[width=\linewidth]{img/USTC_pie_charts_F1_original.pdf}
		\caption{Original F1 Score}
		\Description{The original per-class F1-score of Melatt on USTC-20class without data augmentation.}
	\end{subfigure}
	\hfill
	\begin{subfigure}[b]{0.49\columnwidth}
		\centering
		\includegraphics[width=\linewidth]{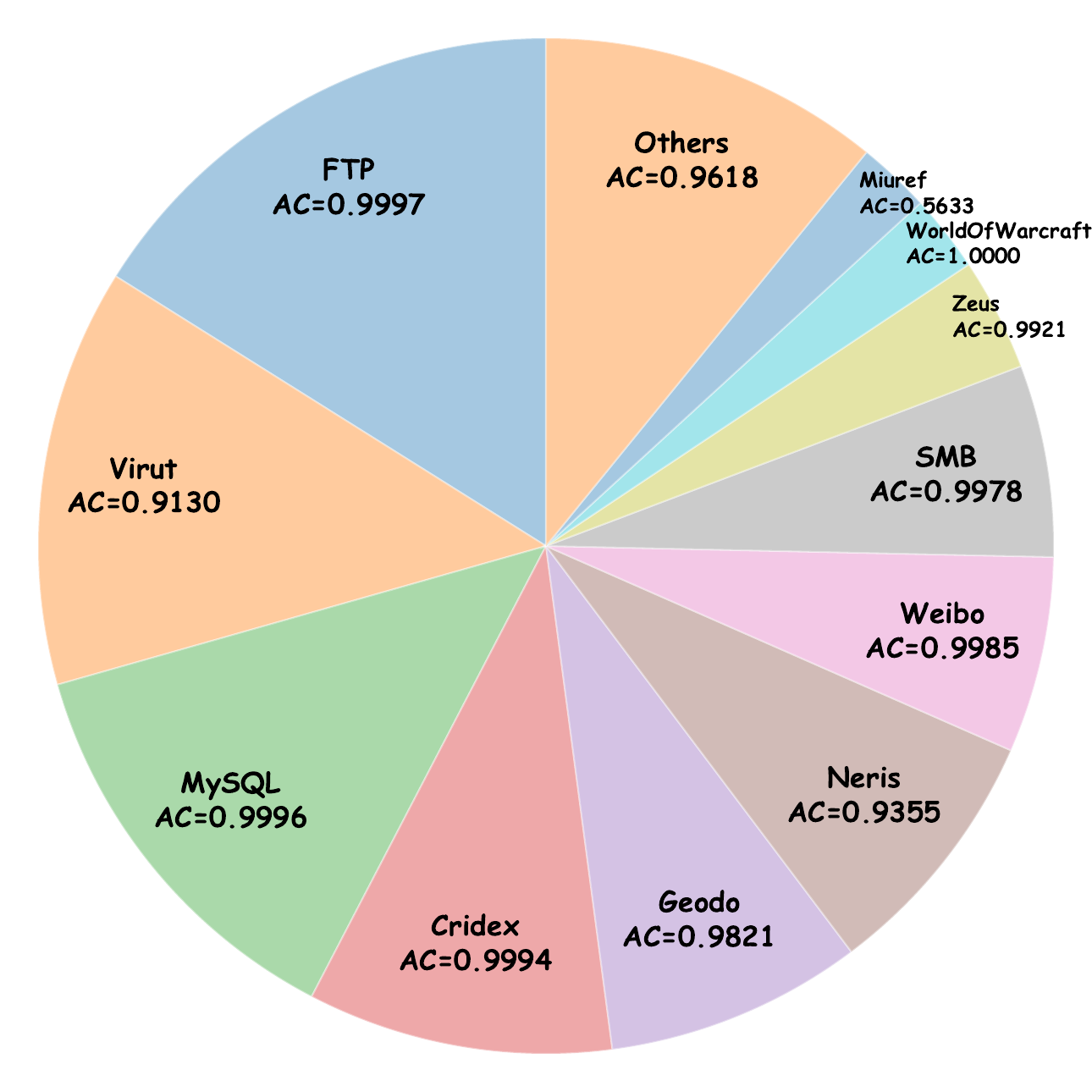}
		\caption{Augmented Accuracy}
		\Description{The per-class accuracy of TDDM-Melatt on USTC-20class after TDDM data augmentation.}
	\end{subfigure}
	\hfill
	\begin{subfigure}[b]{0.49\columnwidth}
		\centering
		\includegraphics[width=\linewidth]{img/USTC_pie_charts_F1_augmented.pdf}
		\caption{Augmented F1 Score}
		\Description{The per-class F1-score of TDDM-Melatt on USTC-20class after TDDM data augmentation.}
	\end{subfigure}
	\caption{Per-class Classification Performance on USTC-20class (AC and F1).}
	\Description{A set of four pie charts showing per-class classification results for USTC-20class. The top-left chart displays the original accuracy per class using Melatt without augmentation. The top-right chart displays the original F1-score per class using Melatt. The bottom-left chart displays the per-class accuracy after augmenting training data with TDDM. The bottom-right chart displays the per-class F1-score after TDDM augmentation. All charts indicate performance metrics across the 20 classes.}
	\label{fig:0}
\end{figure}

\section{Supplementary Figures and Tables}
\label{fig tab}

\begin{figure*}[!t]
	\centering
	\setlength{\tabcolsep}{0pt} 
	\begin{tabular}{@{} 
			>{\centering\arraybackslash}m{0.12\textwidth} 
			*{7}{>{\centering\arraybackslash}m{0.126\textwidth}} 
			@{}}
		\toprule
		\textbf{Timestep} & 900 & 700 & 500 & 300 & 200 & 100 & 0 \\
		\midrule
		\textbf{IDS (UG)} &
		\includegraphics[width=0.126\textwidth]{img/IDS-denoise/step_0900_no_guide.pdf} &
		\includegraphics[width=0.126\textwidth]{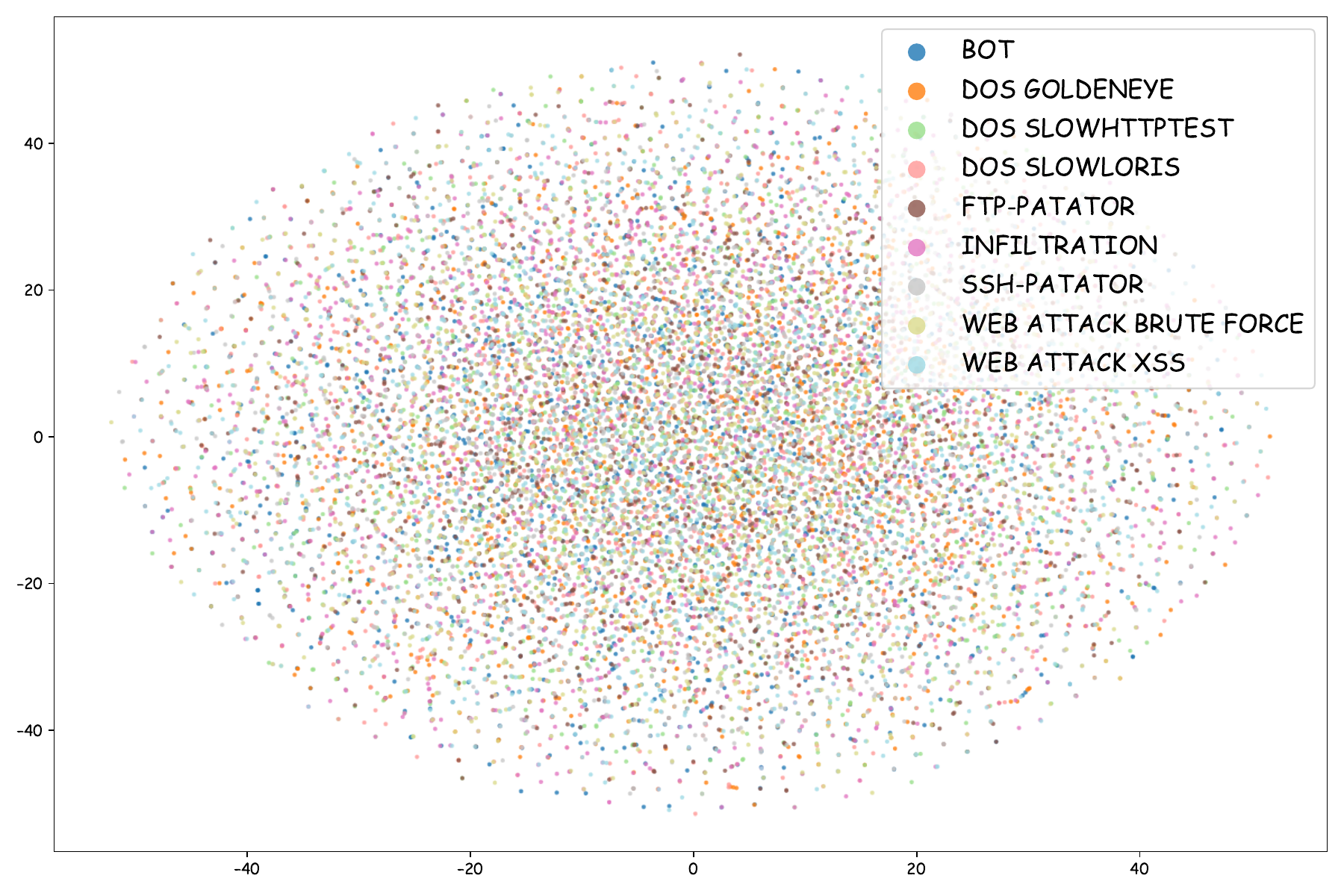} &
		\includegraphics[width=0.126\textwidth]{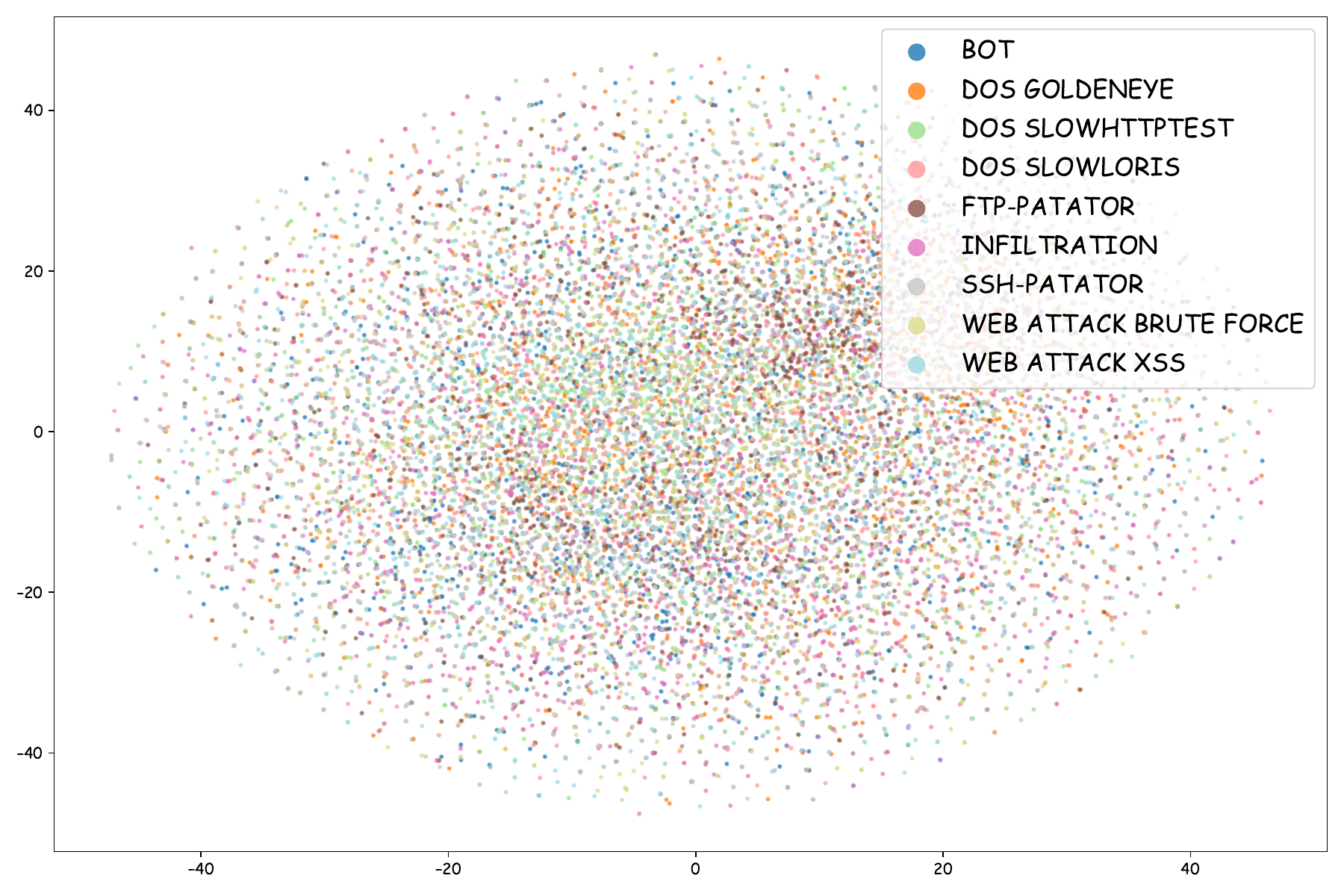} &
		\includegraphics[width=0.126\textwidth]{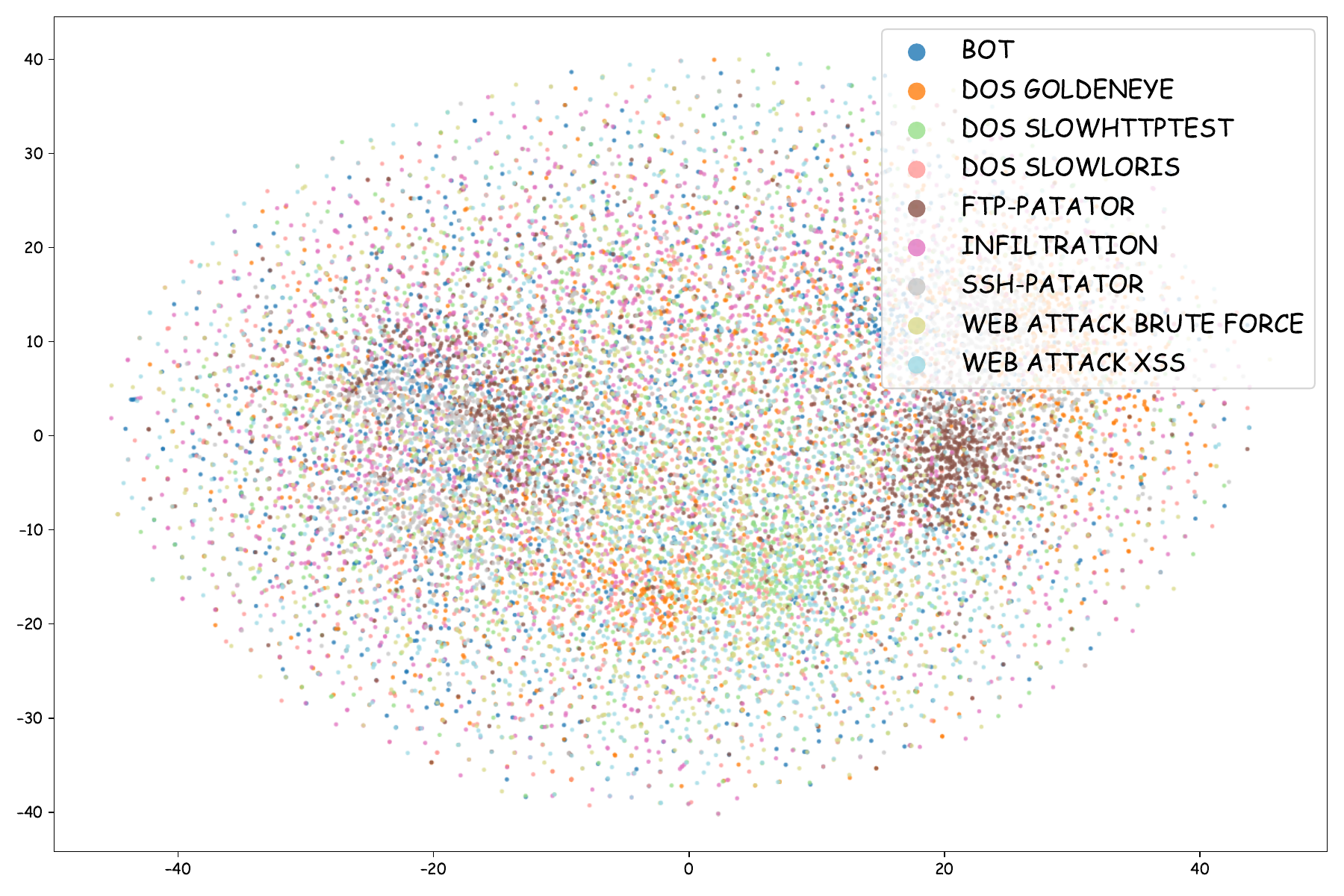} &
		\includegraphics[width=0.126\textwidth]{img/IDS-denoise/step_0200_no_guide.pdf} &
		\includegraphics[width=0.126\textwidth]{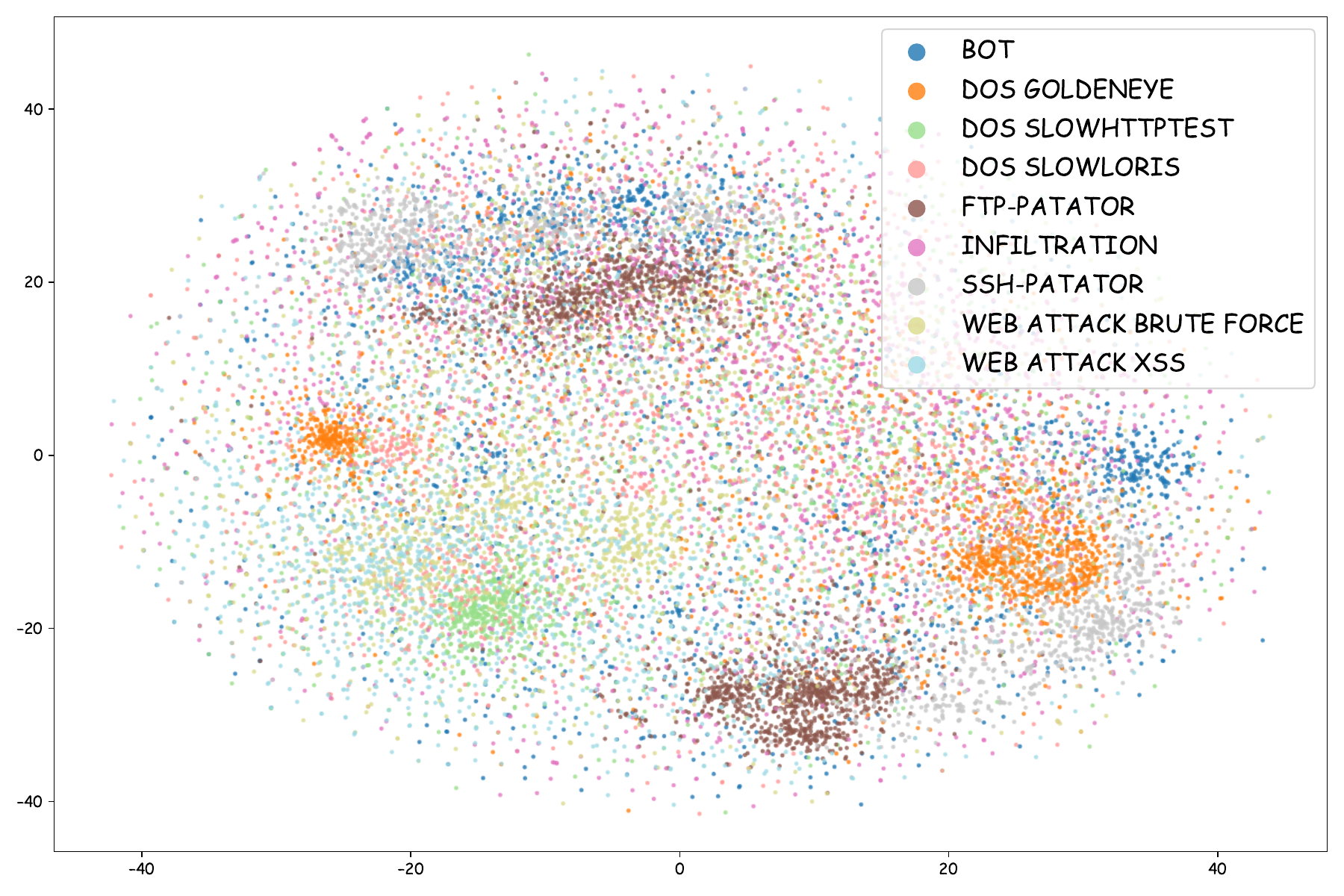} &
		\includegraphics[width=0.126\textwidth]{img/IDS-denoise/step_0000_no_guide.pdf} \\
		\midrule
		\textbf{IDS (G)} &
		\includegraphics[width=0.126\textwidth]{img/IDS-denoise/step_0900_guide.pdf} &
		\includegraphics[width=0.126\textwidth]{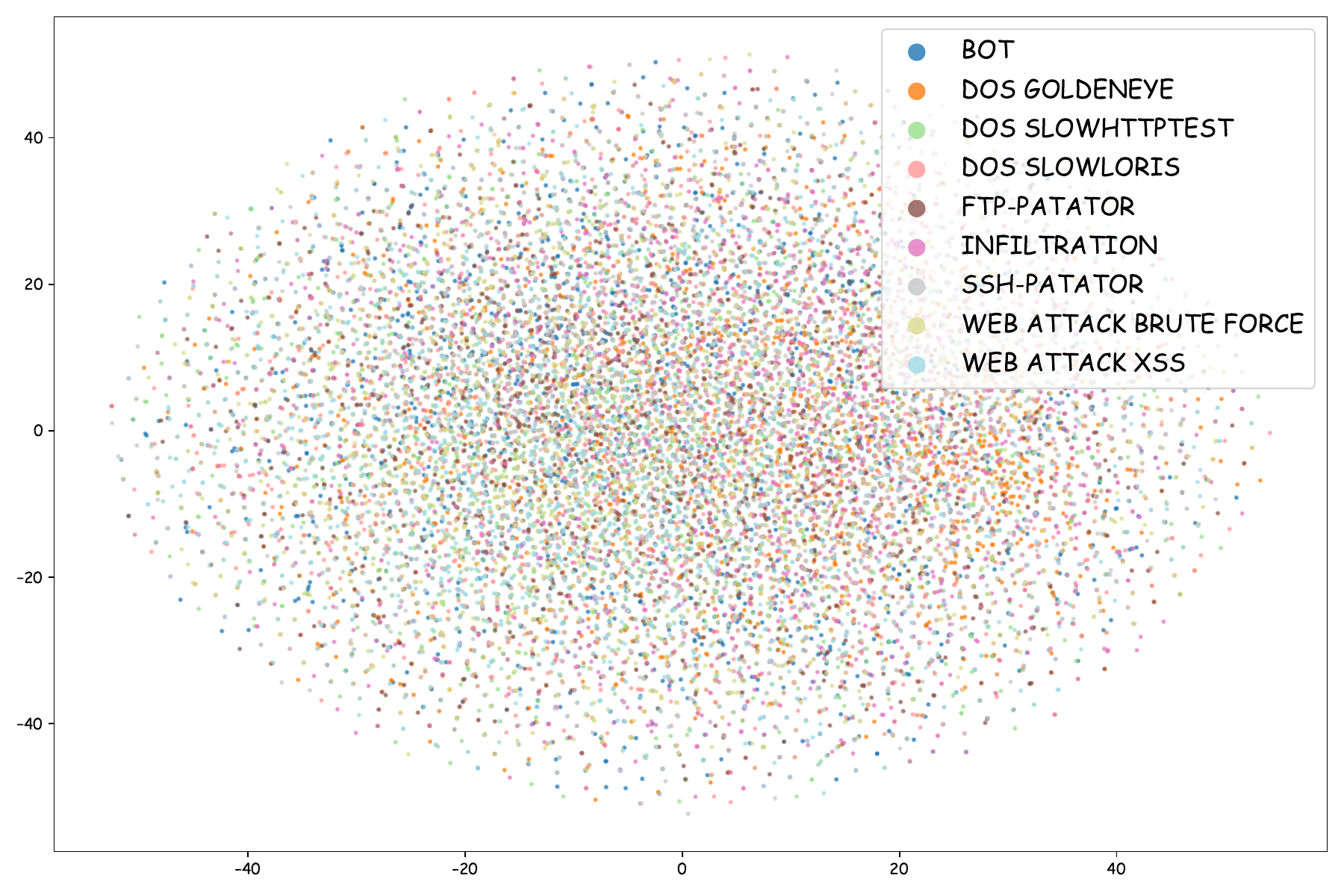} &
		\includegraphics[width=0.126\textwidth]{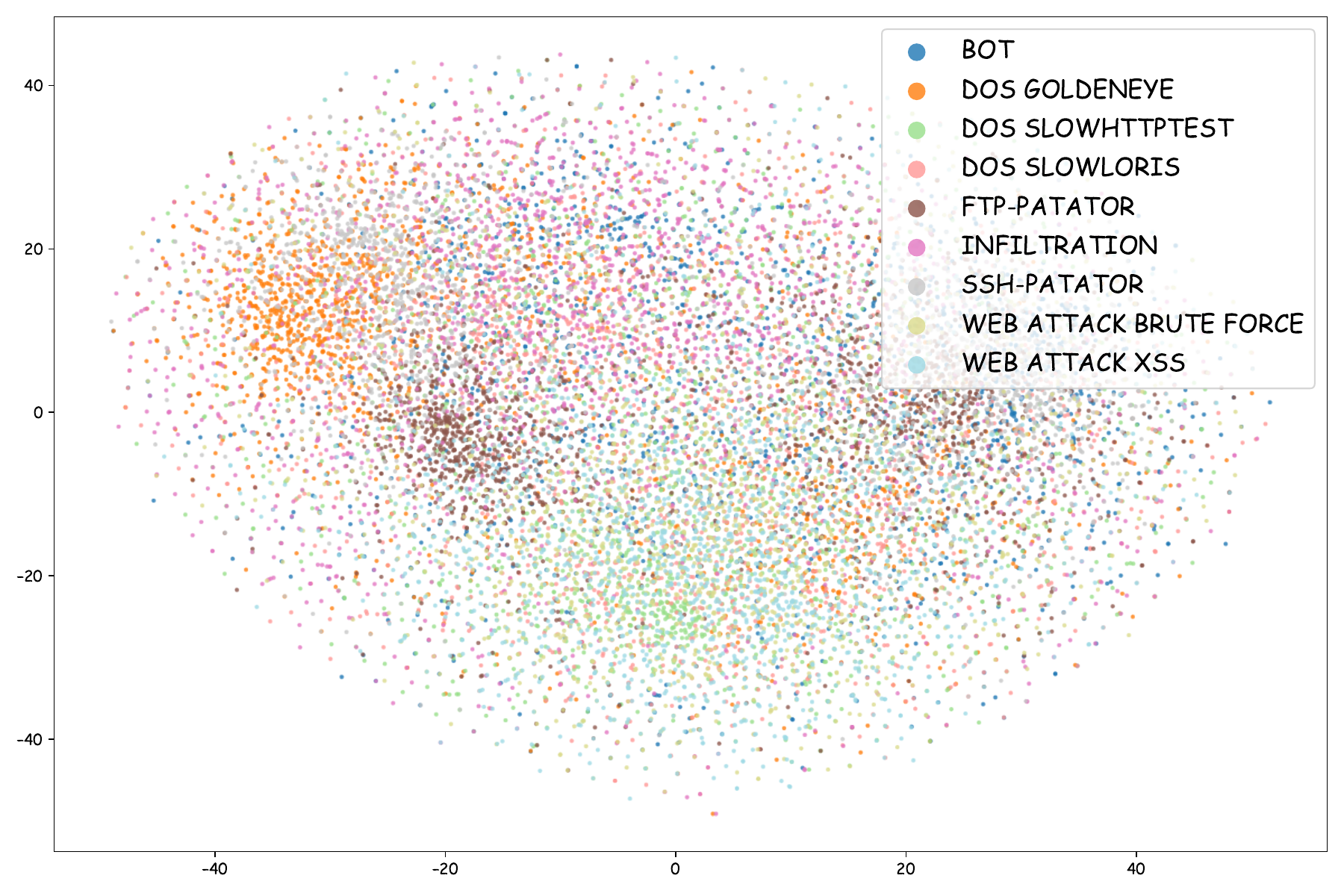} &
		\includegraphics[width=0.126\textwidth]{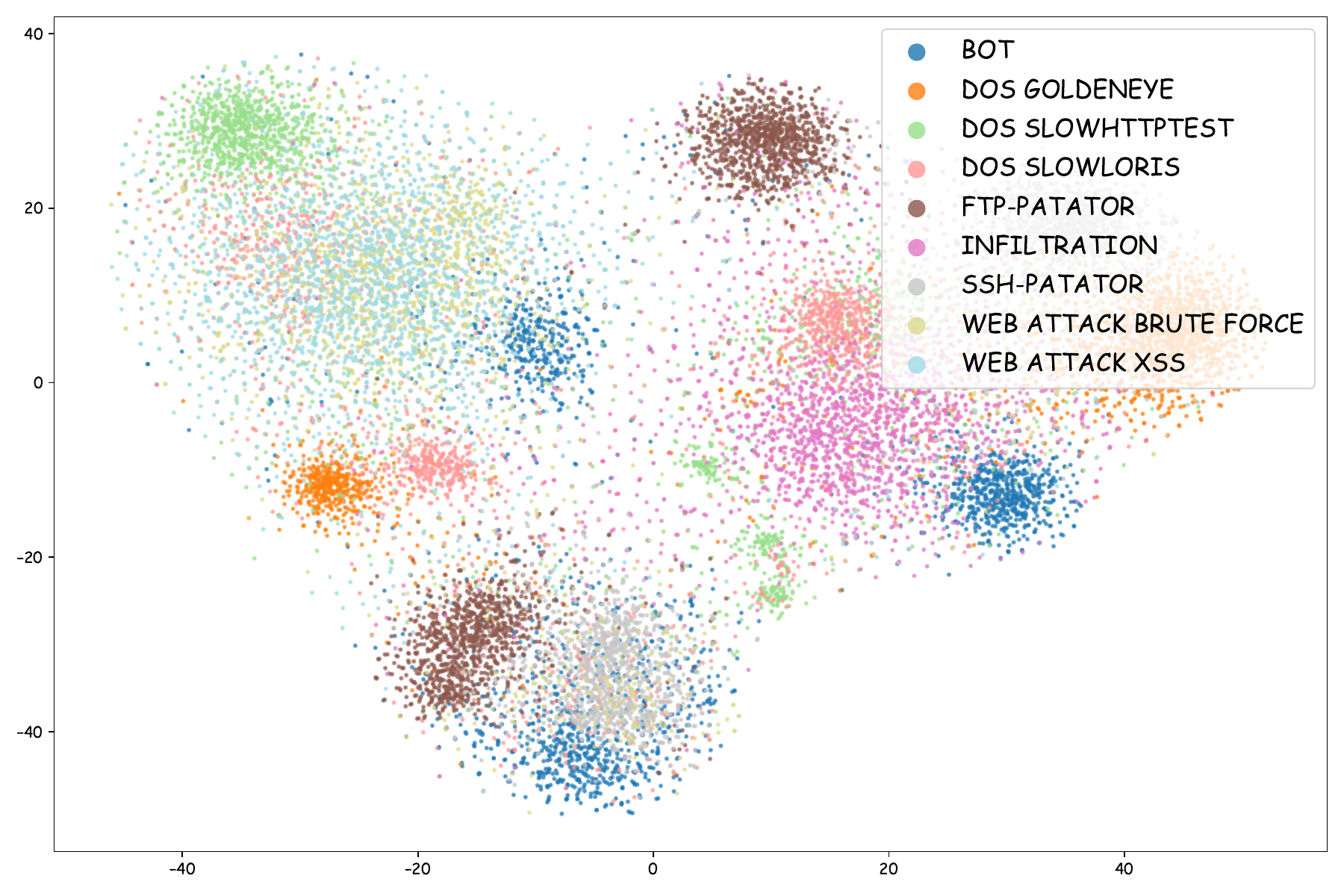} &
		\includegraphics[width=0.126\textwidth]{img/IDS-denoise/step_0200_guide.pdf} &
		\includegraphics[width=0.126\textwidth]{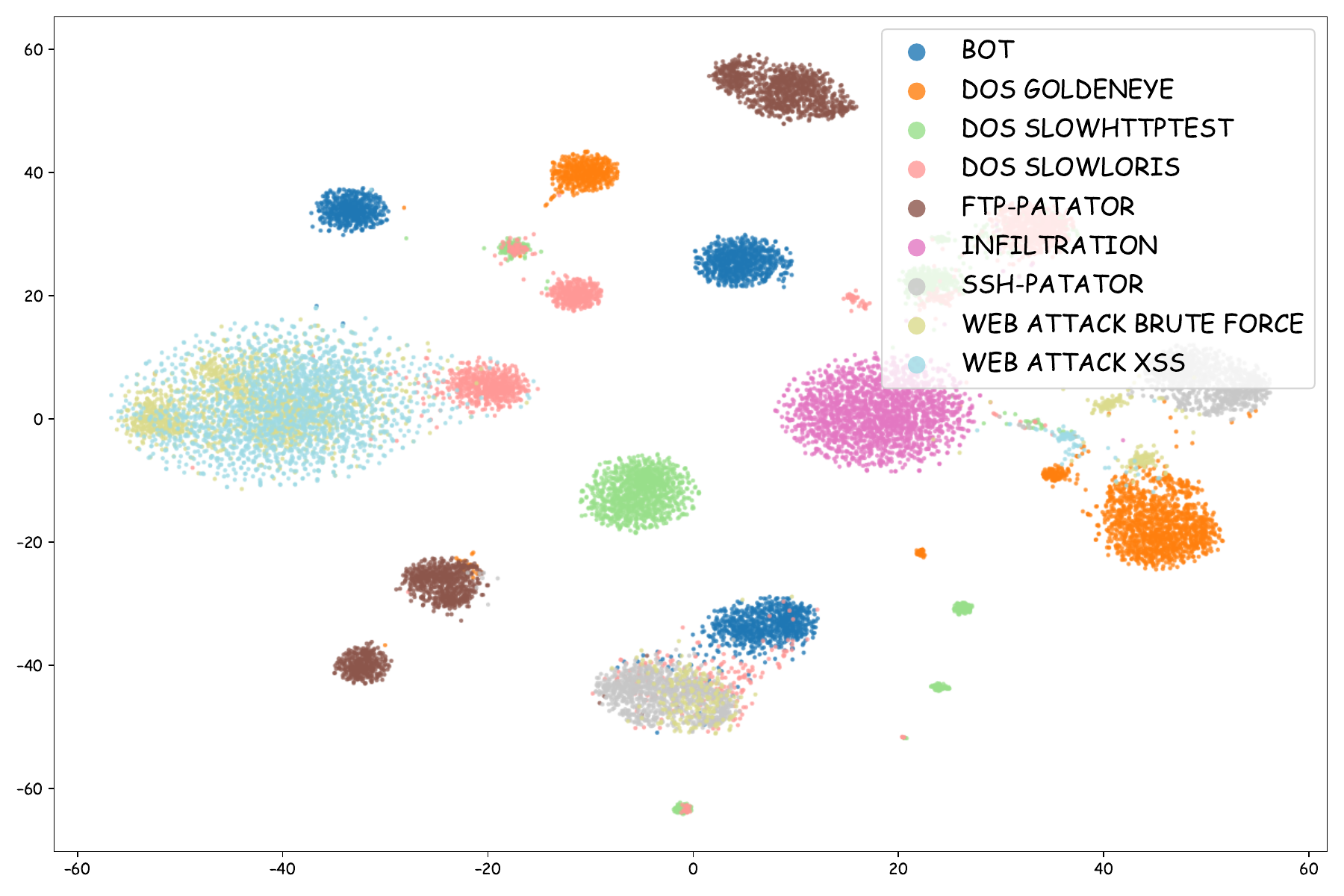} &
		\includegraphics[width=0.126\textwidth]{img/IDS-denoise/step_0000_guide.pdf} \\
		\midrule
		\textbf{USTC (UG)} &
		\includegraphics[width=0.126\textwidth]{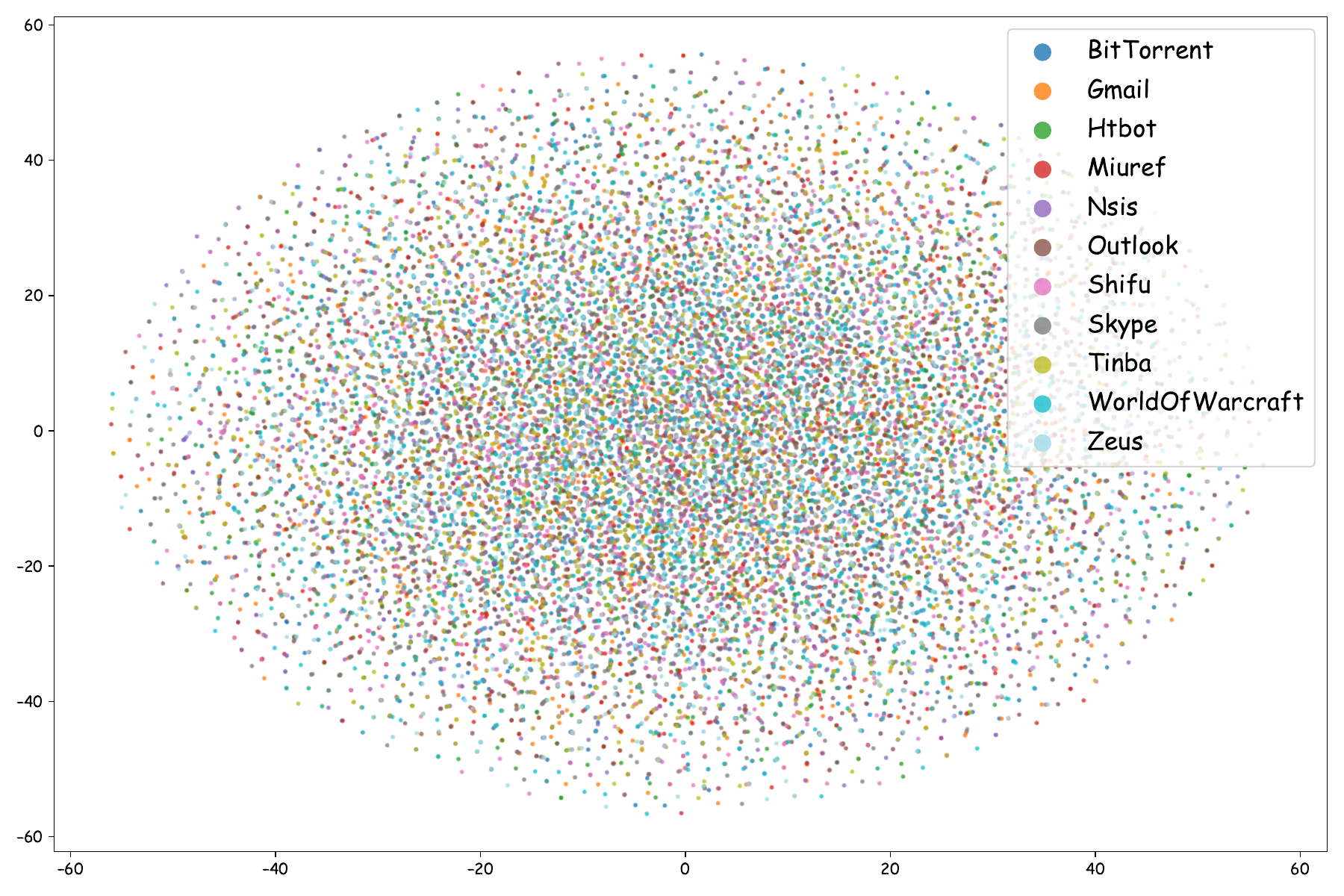} &
		\includegraphics[width=0.126\textwidth]{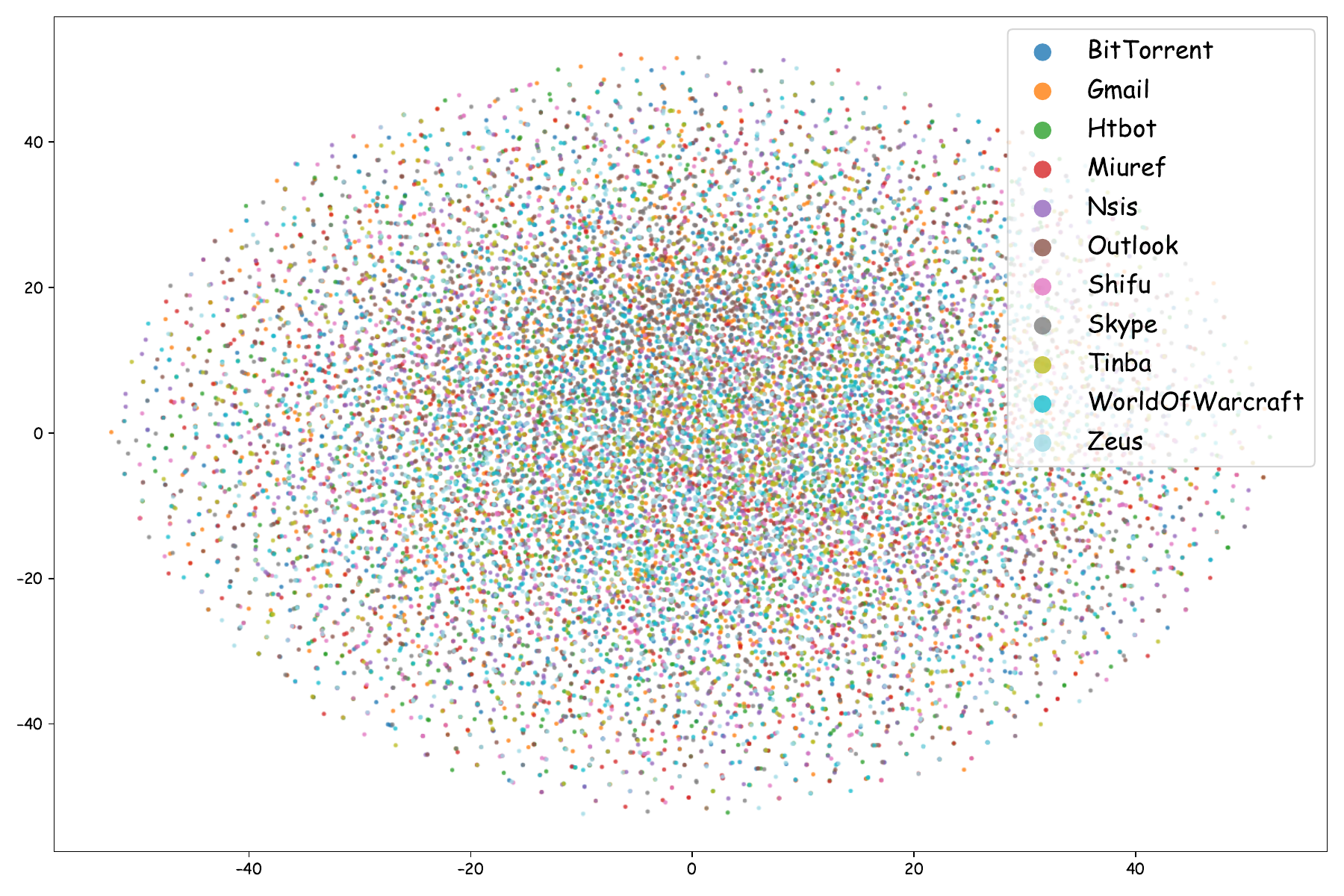} &
		\includegraphics[width=0.126\textwidth]{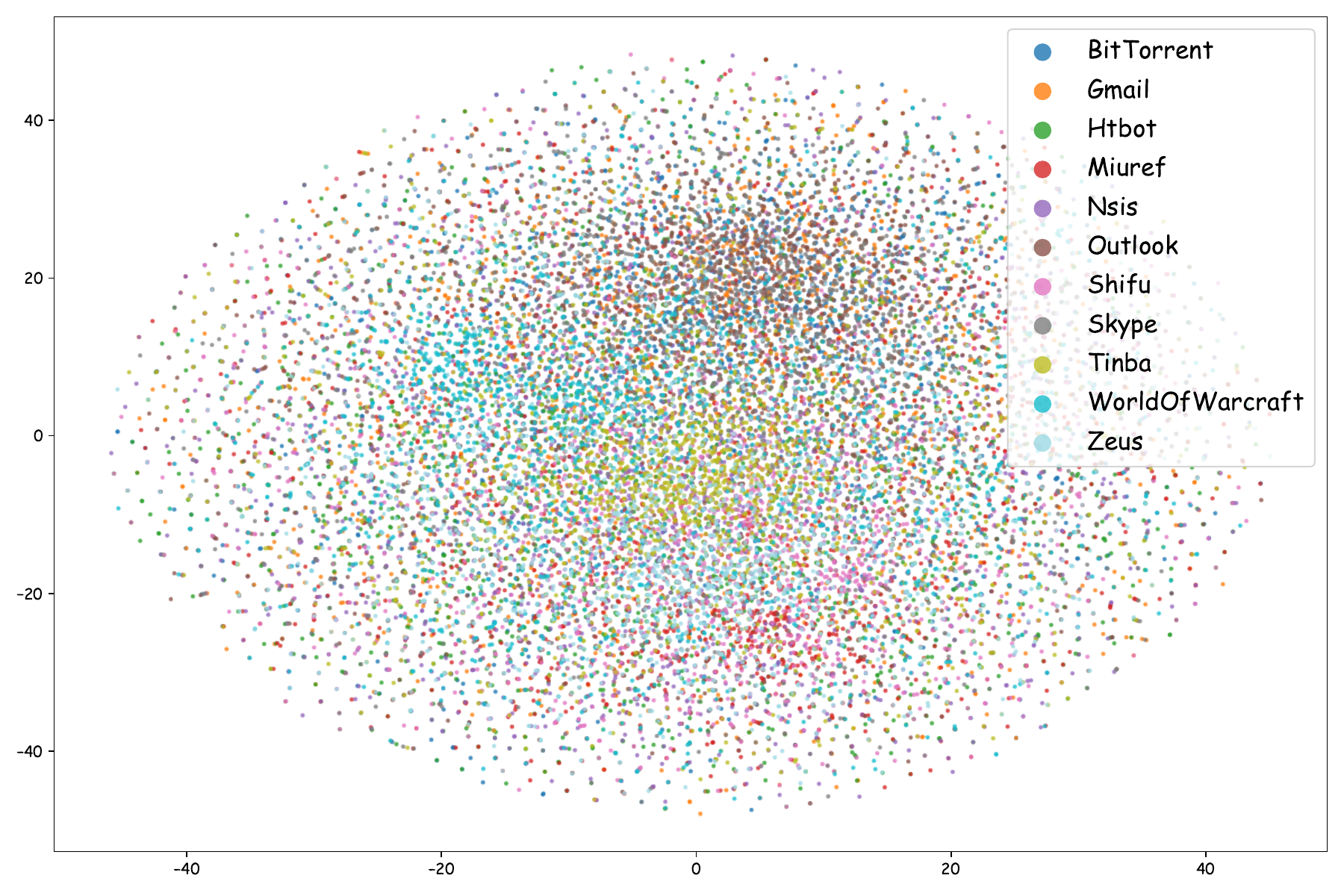} &
		\includegraphics[width=0.126\textwidth]{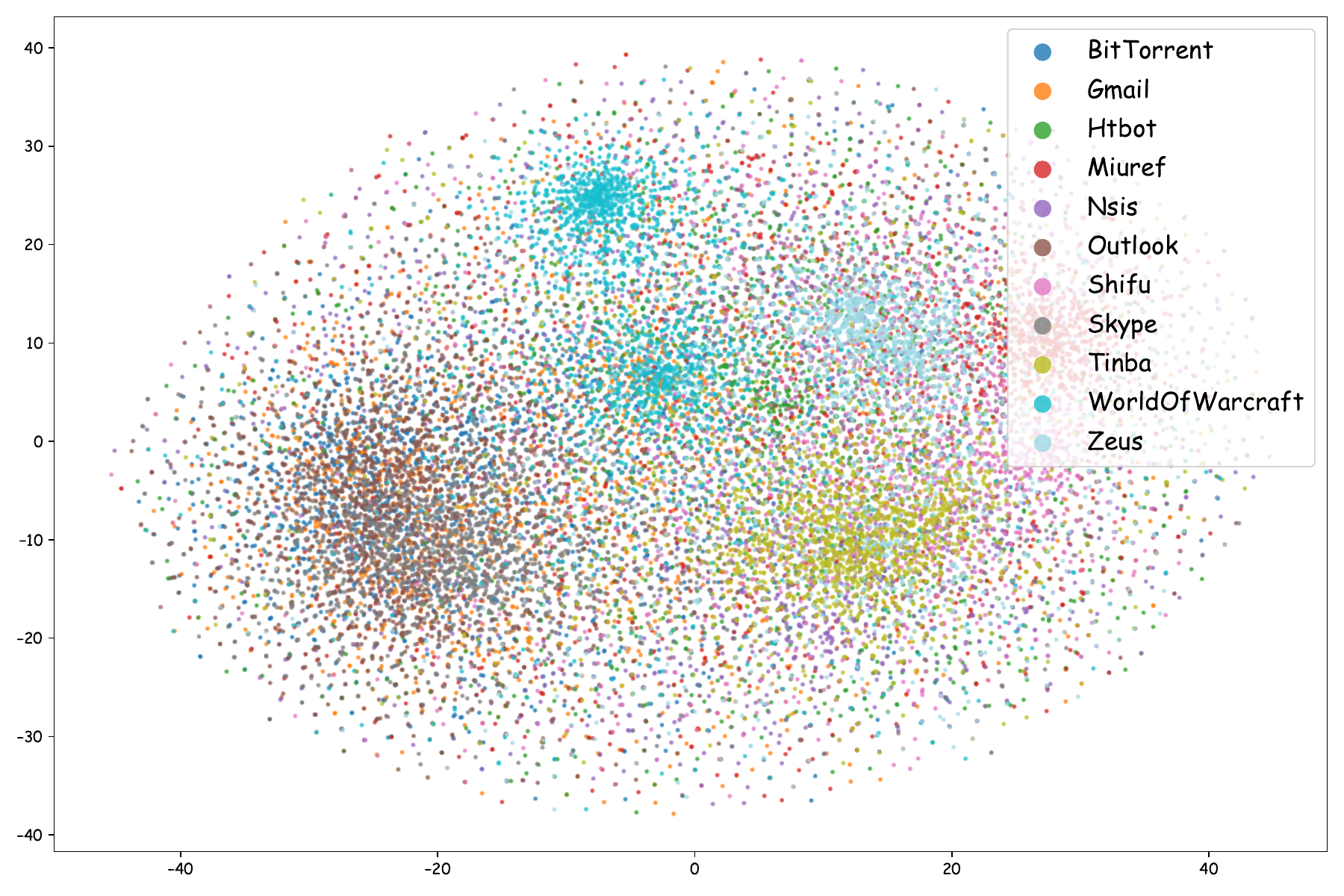} &
		\includegraphics[width=0.126\textwidth]{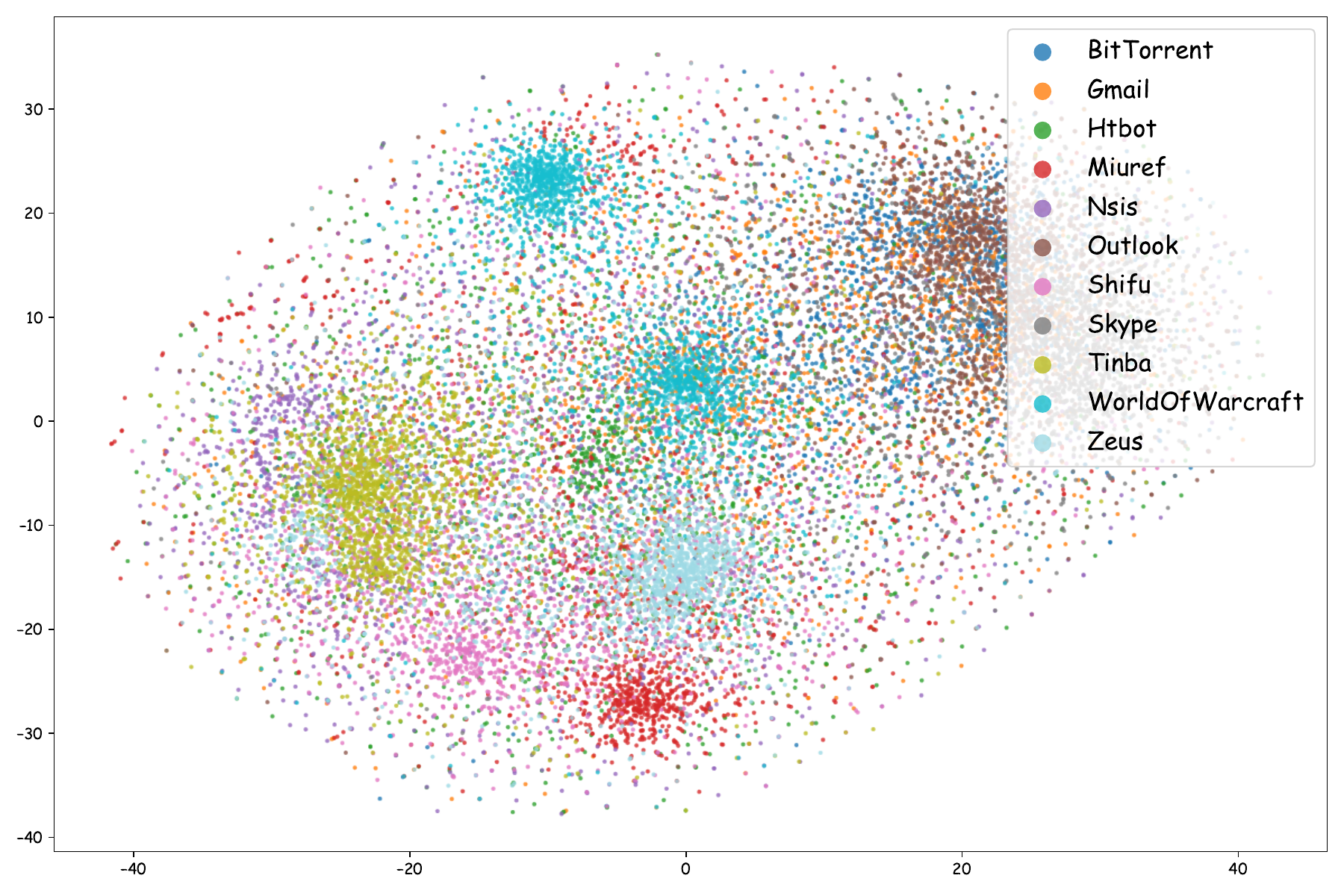} &
		\includegraphics[width=0.126\textwidth]{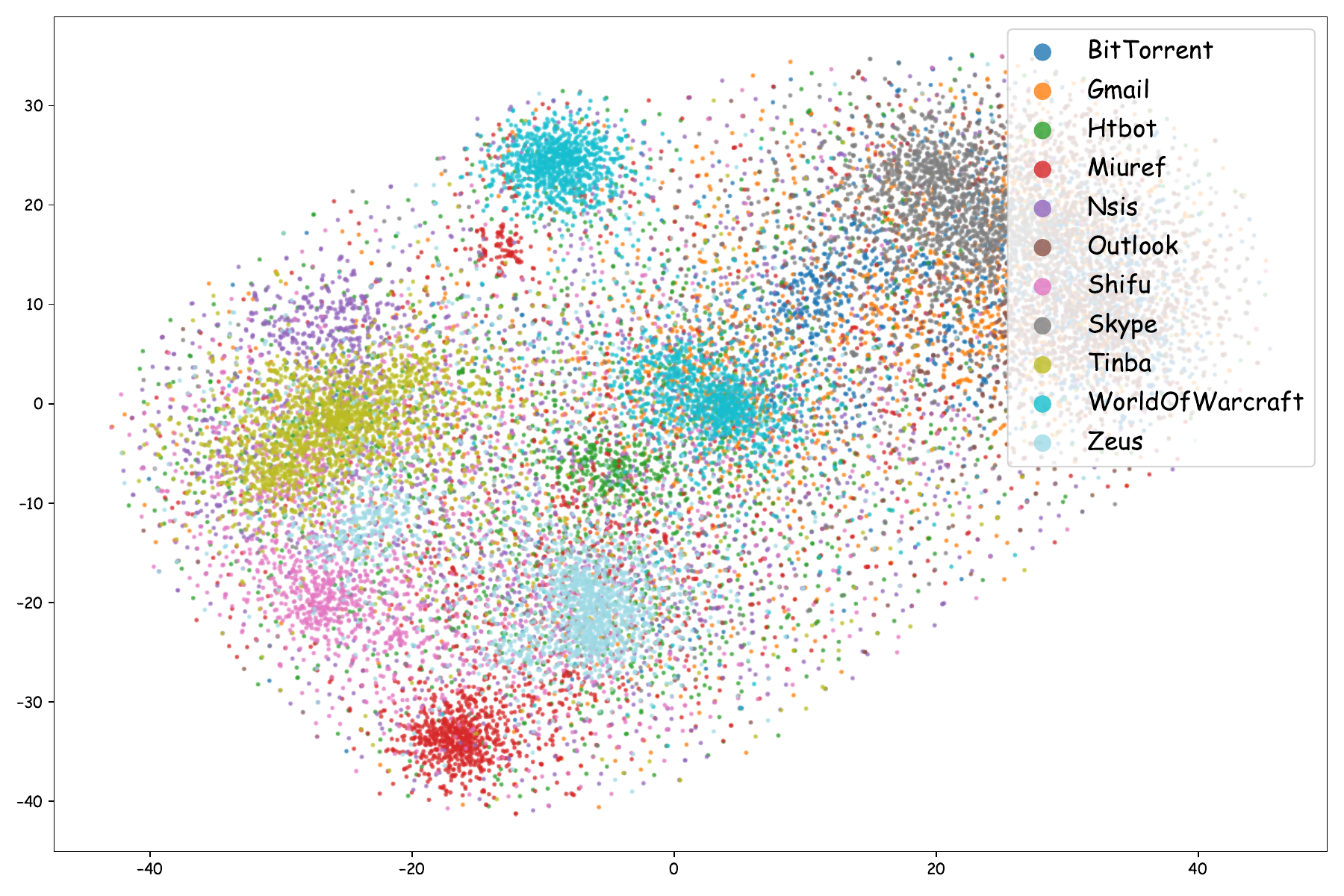} &
		\includegraphics[width=0.126\textwidth]{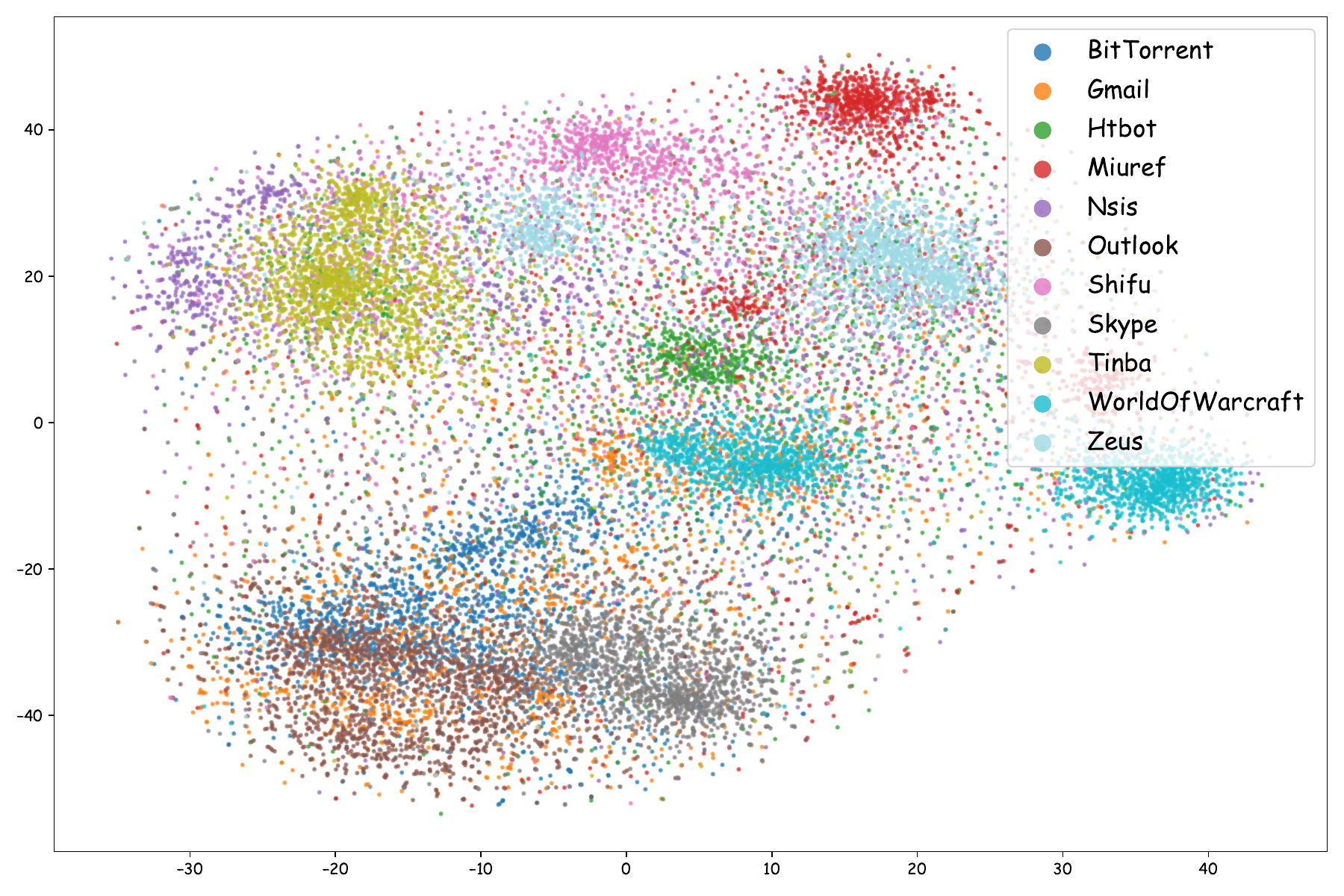} \\
		\midrule
		\textbf{USTC (G)} &
		\includegraphics[width=0.126\textwidth]{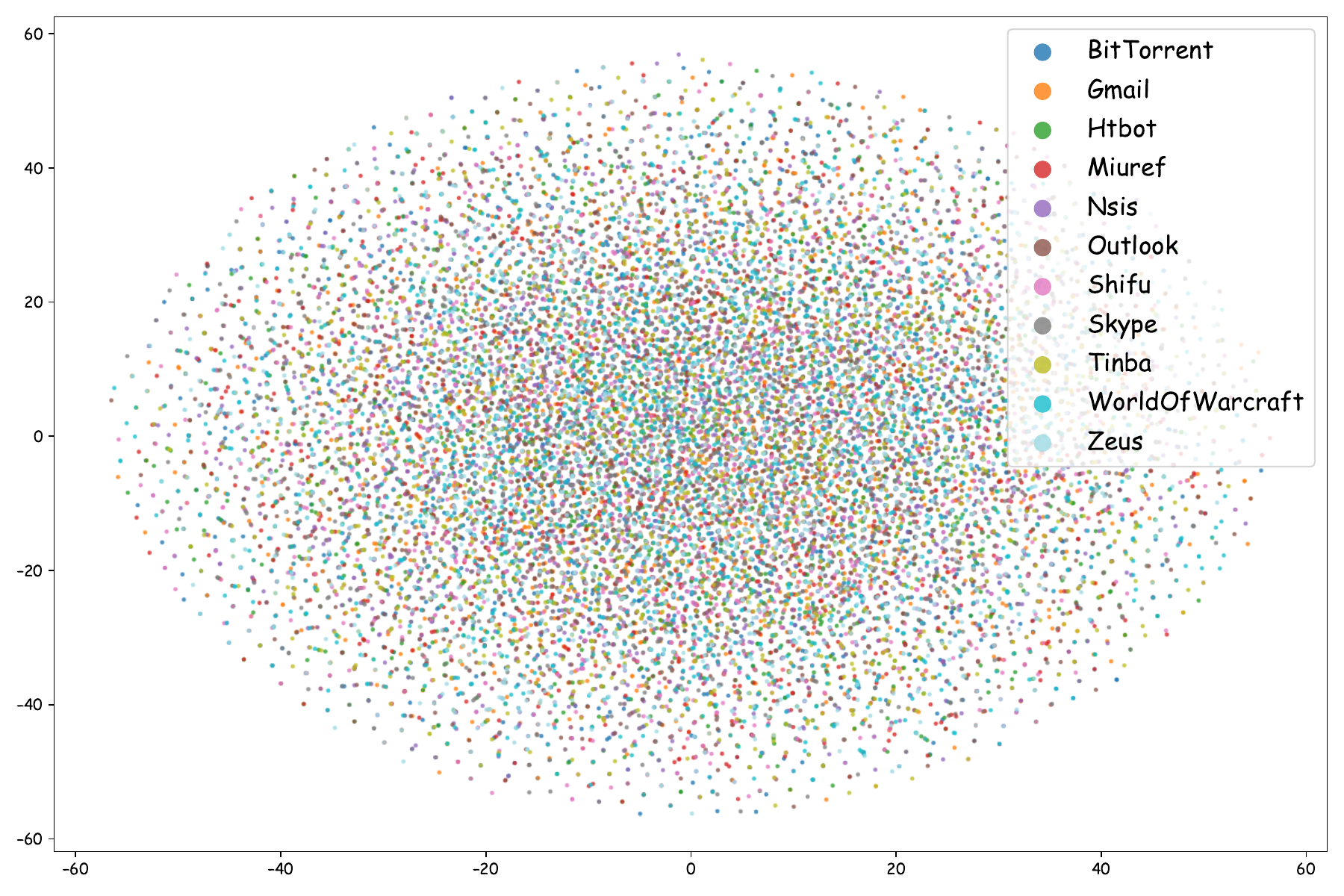} &
		\includegraphics[width=0.126\textwidth]{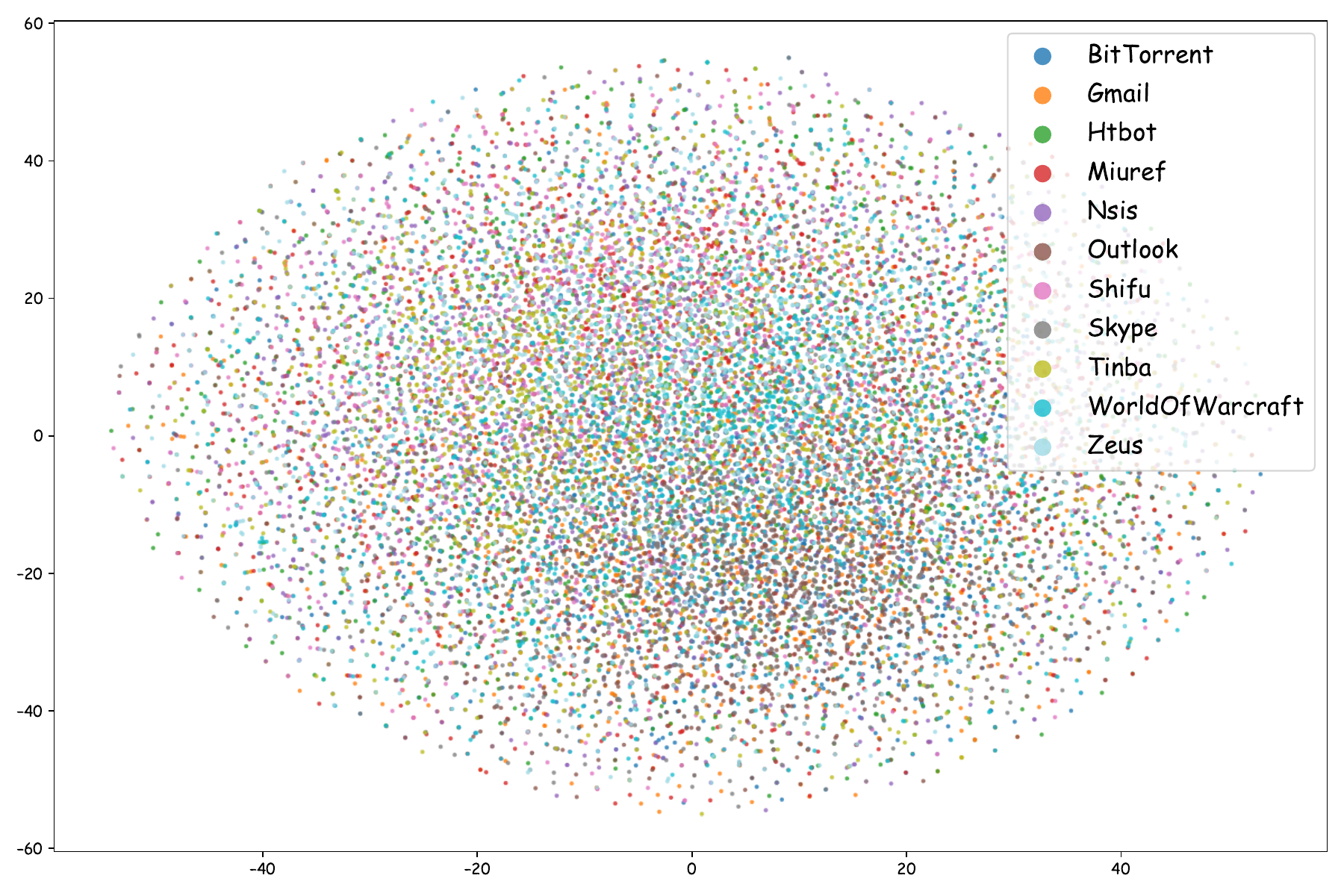} &
		\includegraphics[width=0.126\textwidth]{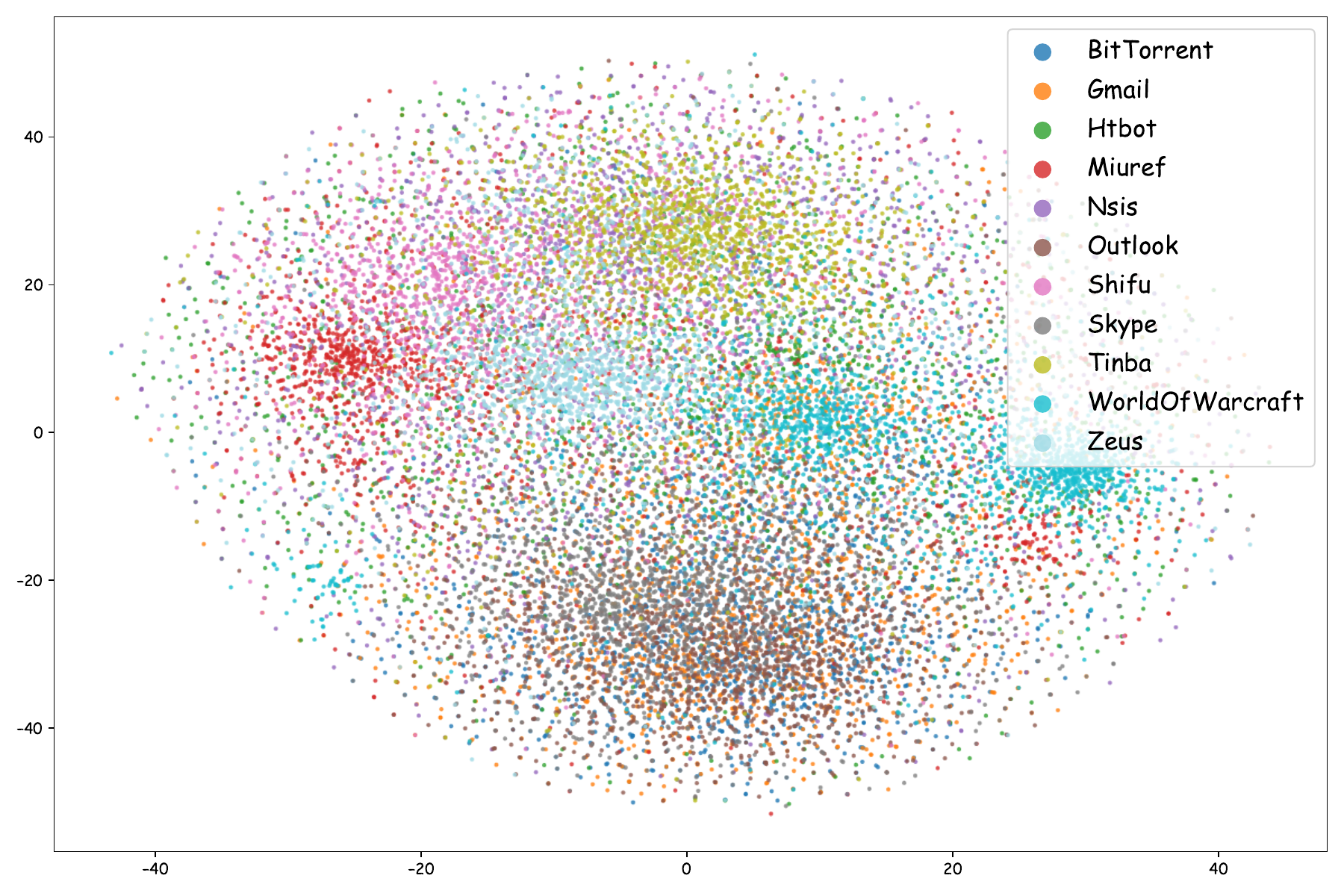} &
		\includegraphics[width=0.126\textwidth]{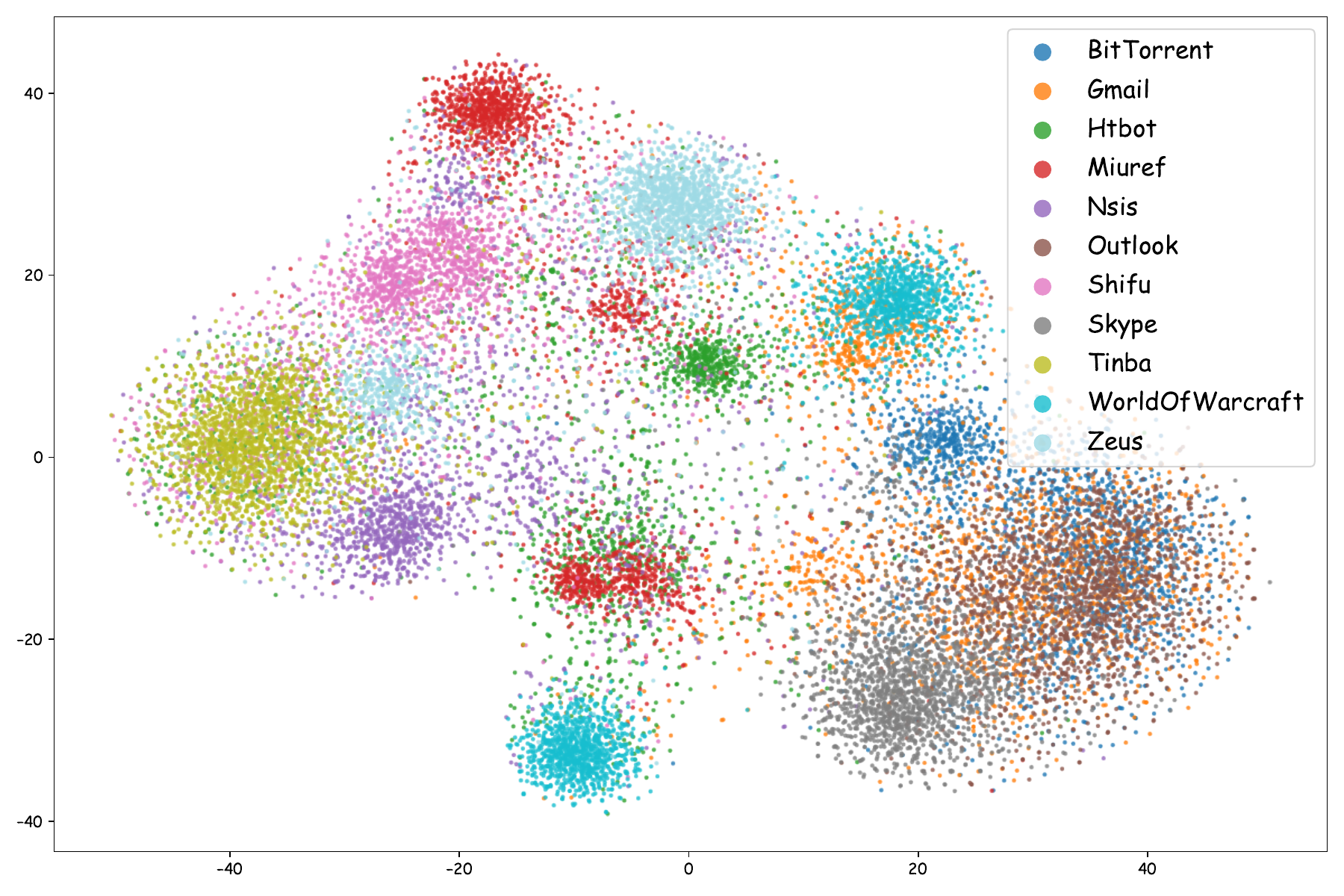} &
		\includegraphics[width=0.126\textwidth]{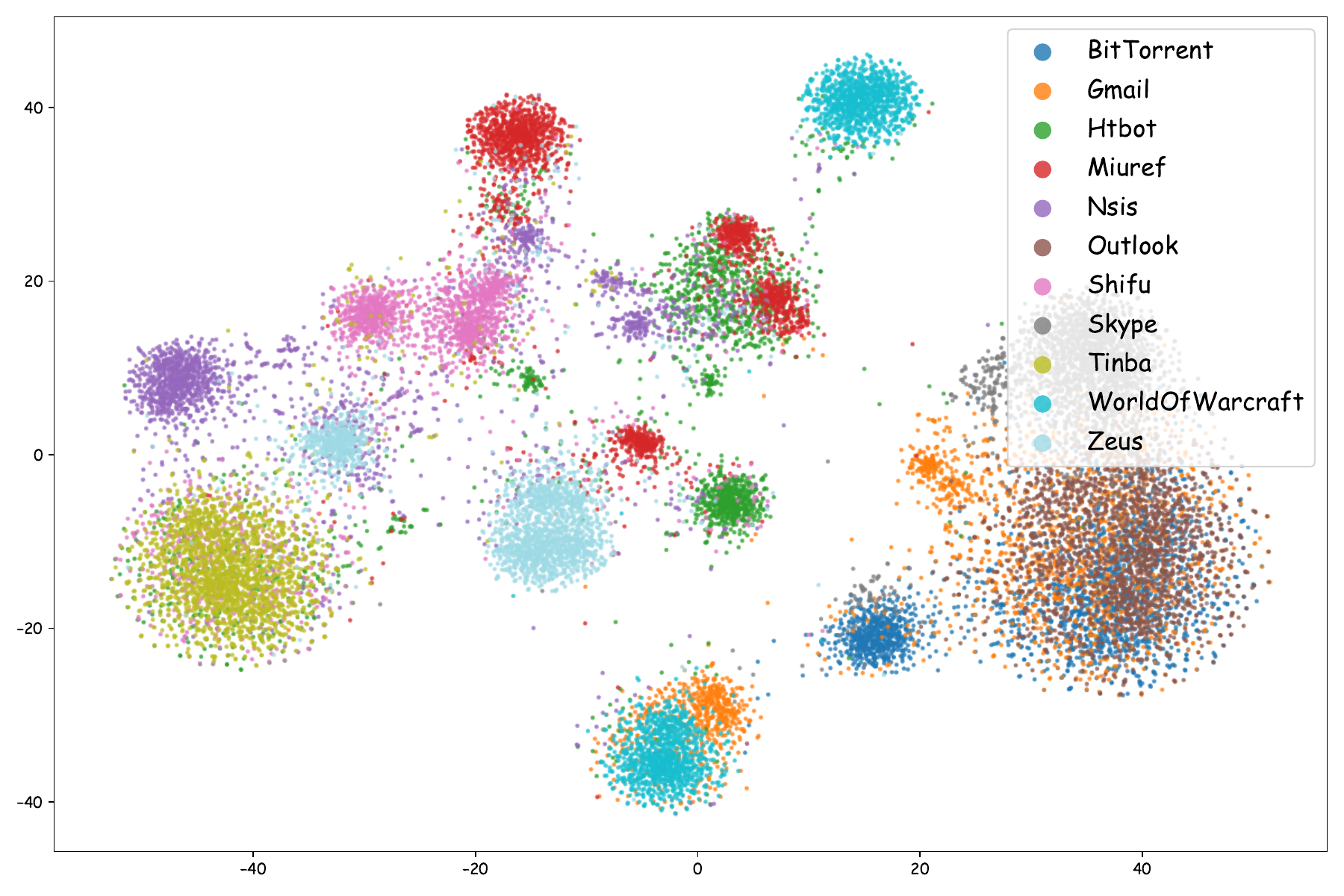} &
		\includegraphics[width=0.126\textwidth]{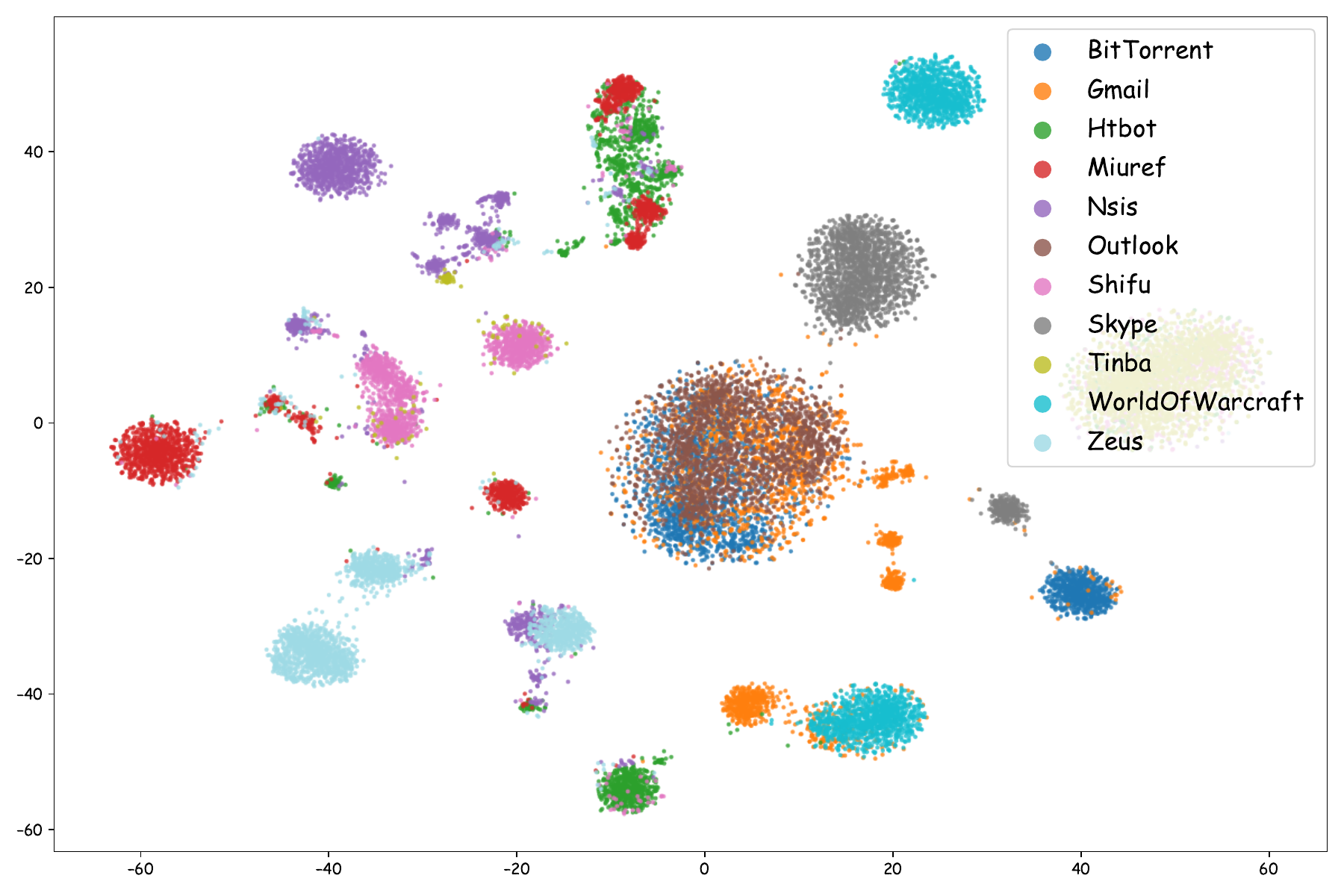} &
		\includegraphics[width=0.126\textwidth]{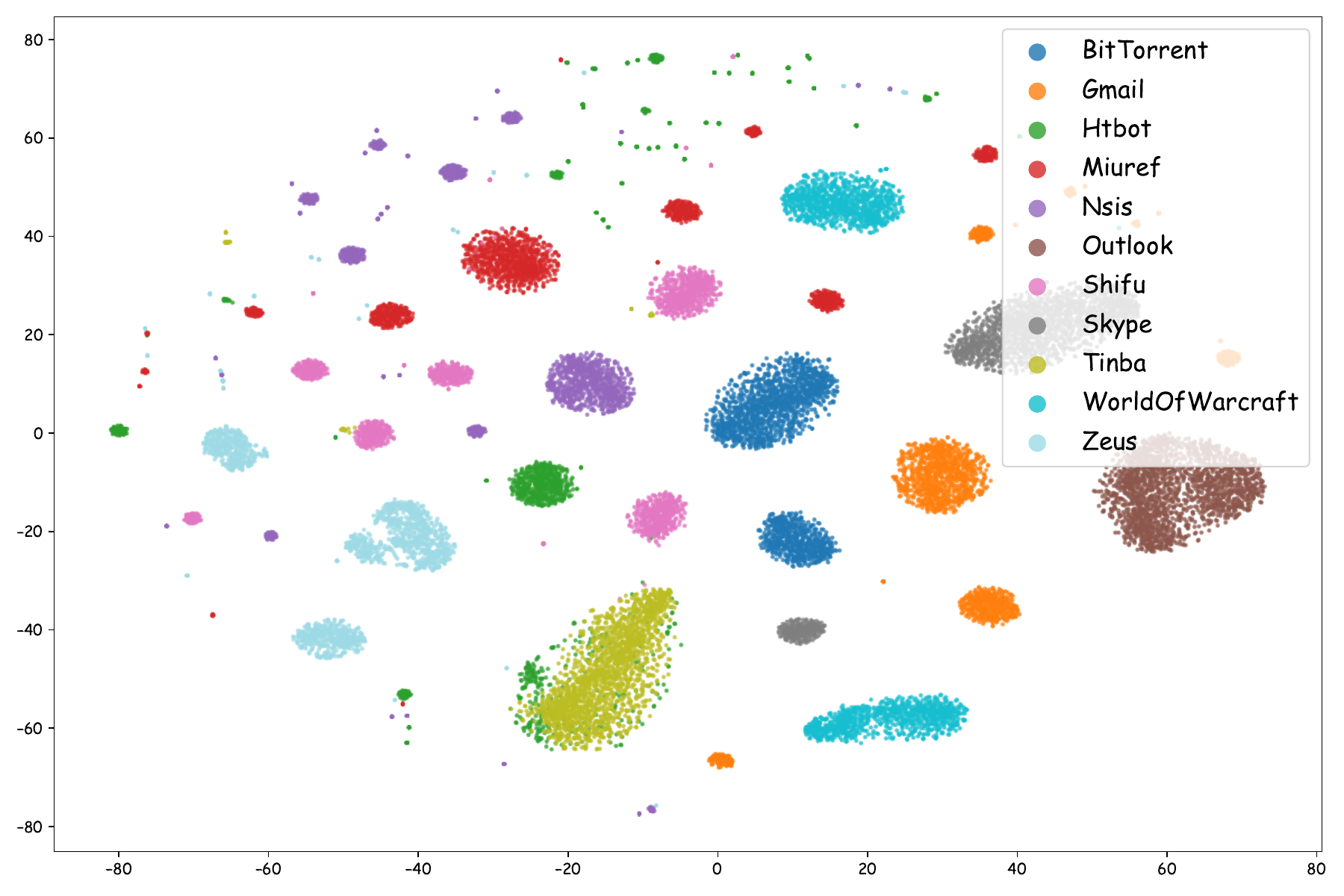} \\
		\midrule
		\textbf{TLS (UG)} &
		\includegraphics[width=0.126\textwidth]{img/TLS-denoise/step_0900_no_guide.pdf} &
		\includegraphics[width=0.126\textwidth]{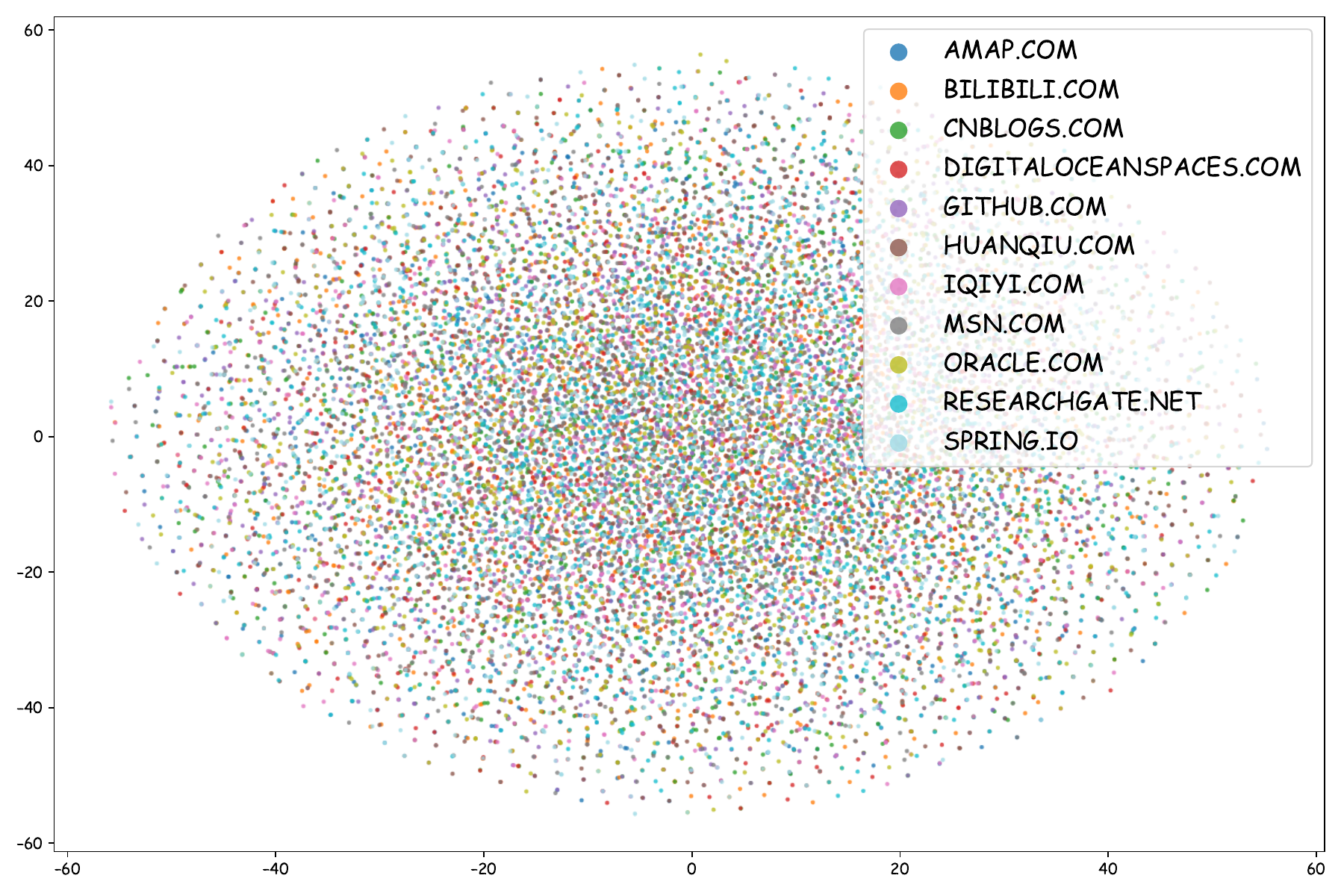} &
		\includegraphics[width=0.126\textwidth]{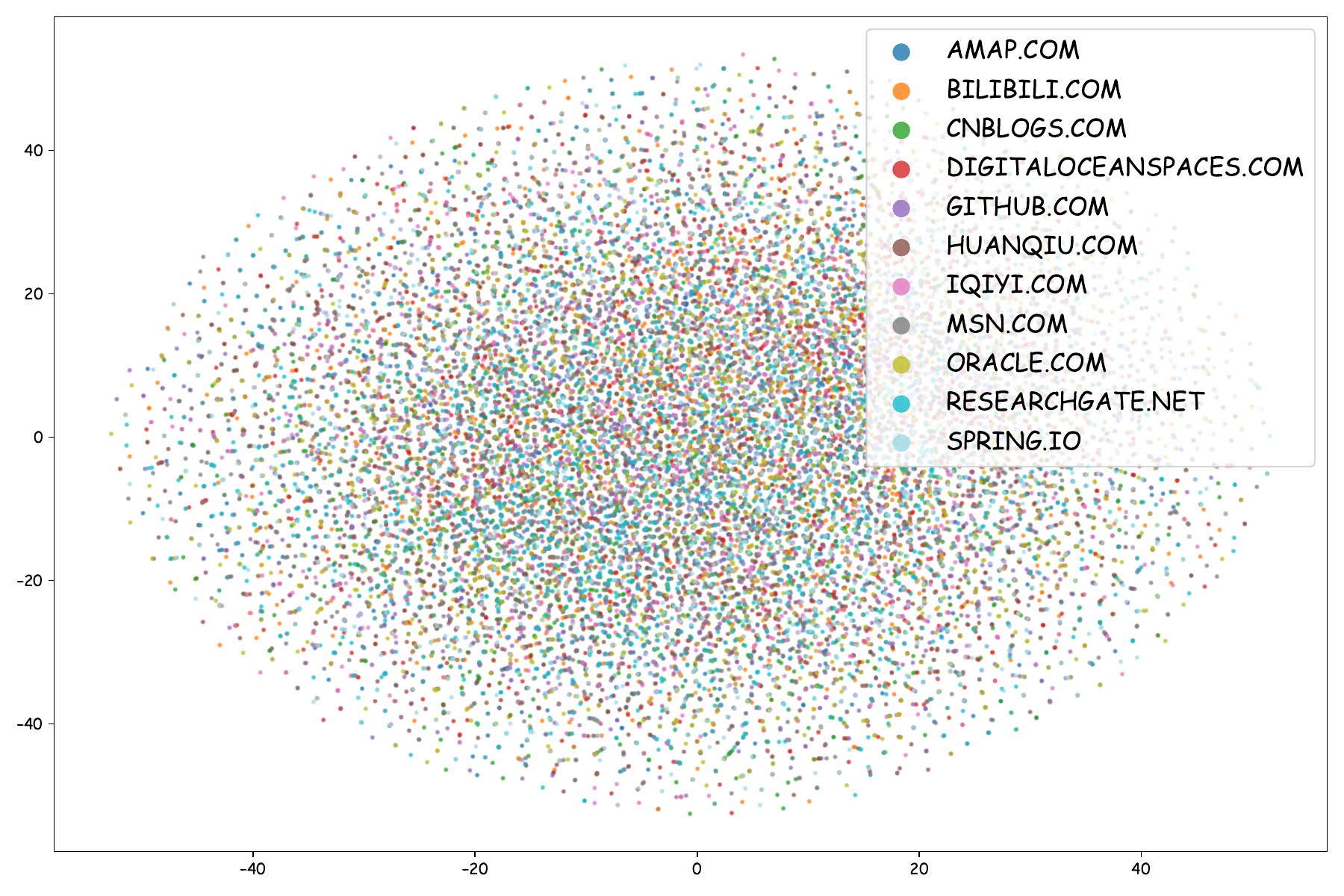} &
		\includegraphics[width=0.126\textwidth]{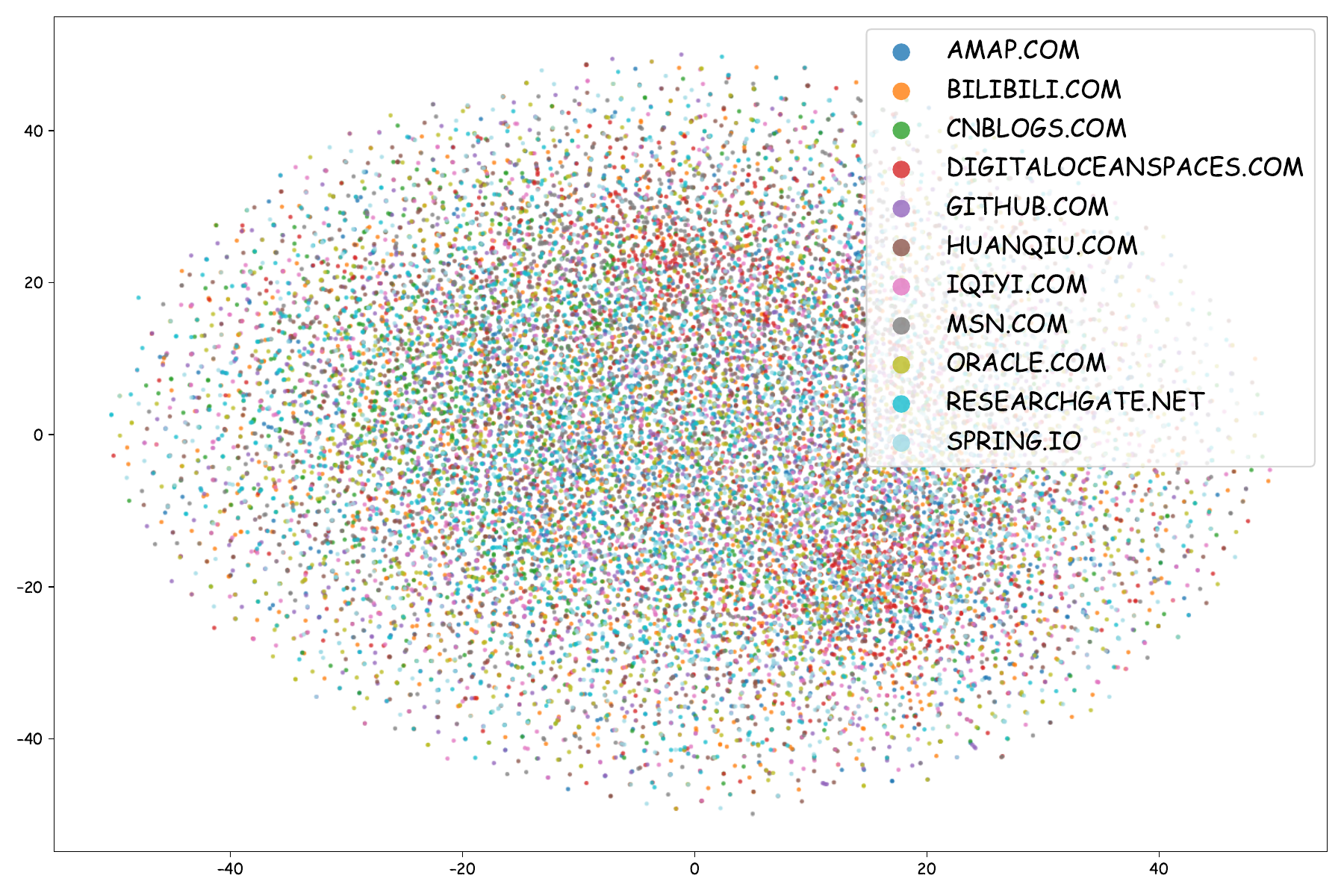} &
		\includegraphics[width=0.126\textwidth]{img/TLS-denoise/step_0200_no_guide.pdf} &
		\includegraphics[width=0.126\textwidth]{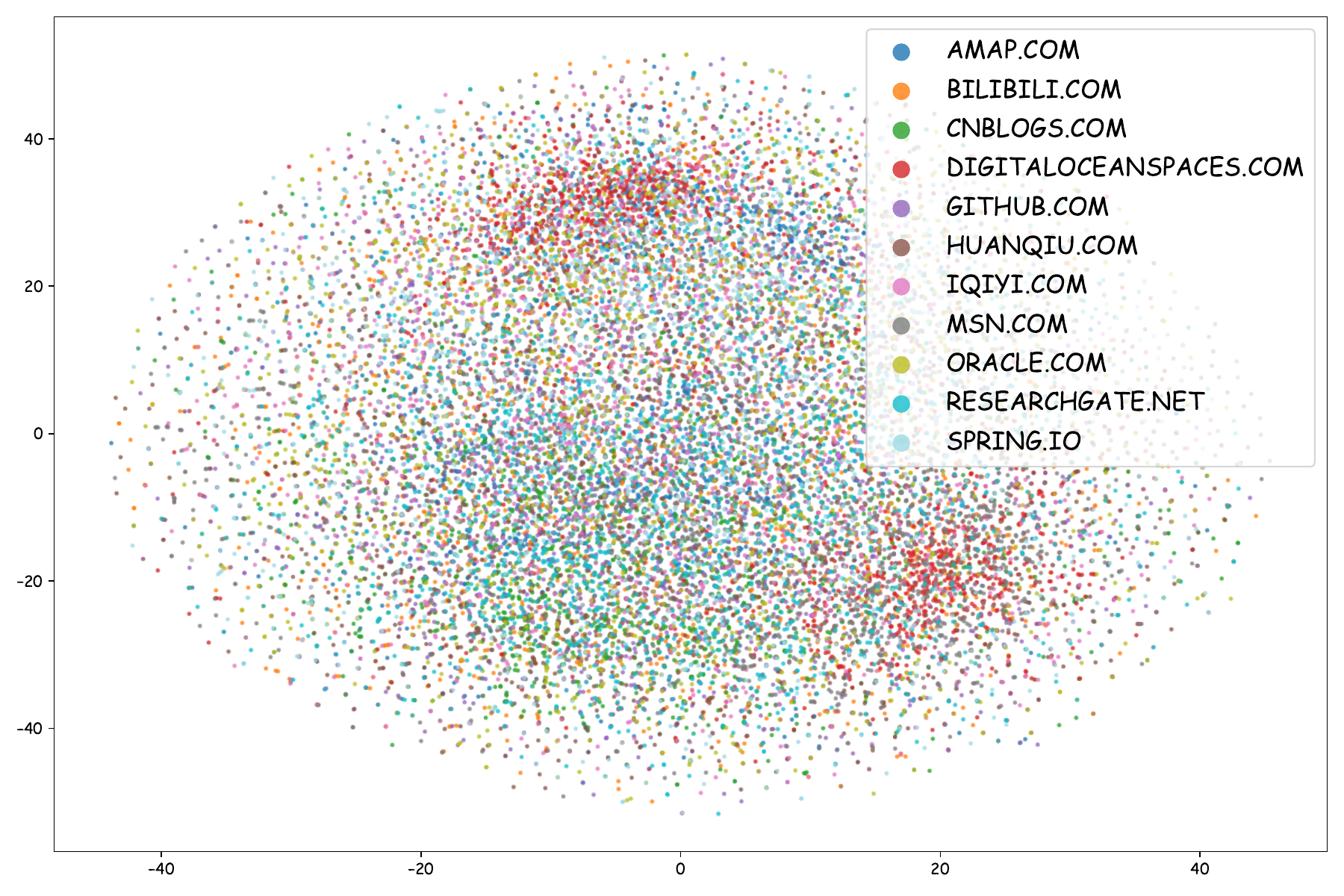} &
		\includegraphics[width=0.126\textwidth]{img/TLS-denoise/step_0000_no_guide.pdf} \\
		\midrule
		\textbf{TLS (G)} &
		\includegraphics[width=0.126\textwidth]{img/TLS-denoise/step_0900_guide.pdf} &
		\includegraphics[width=0.126\textwidth]{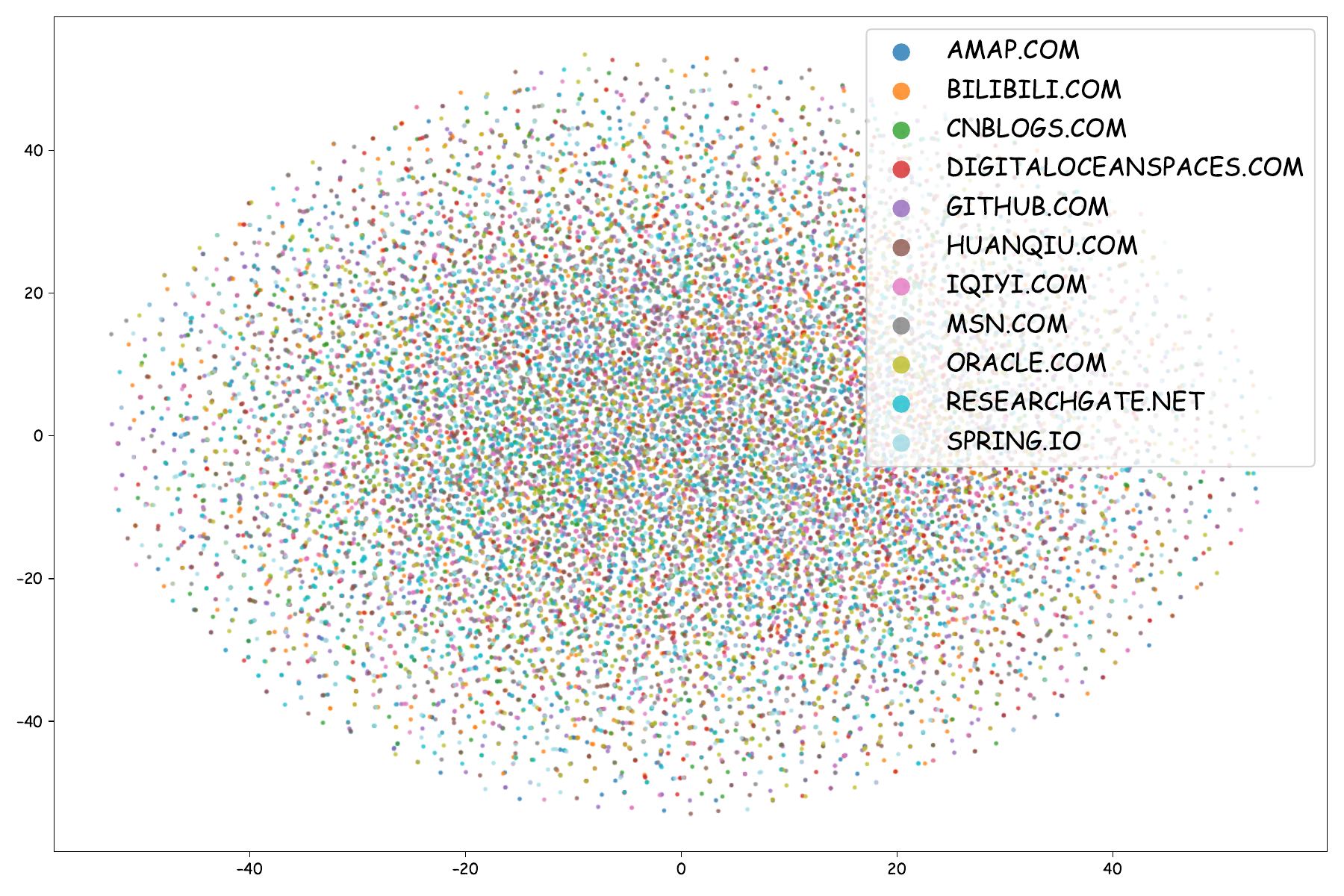} &
		\includegraphics[width=0.126\textwidth]{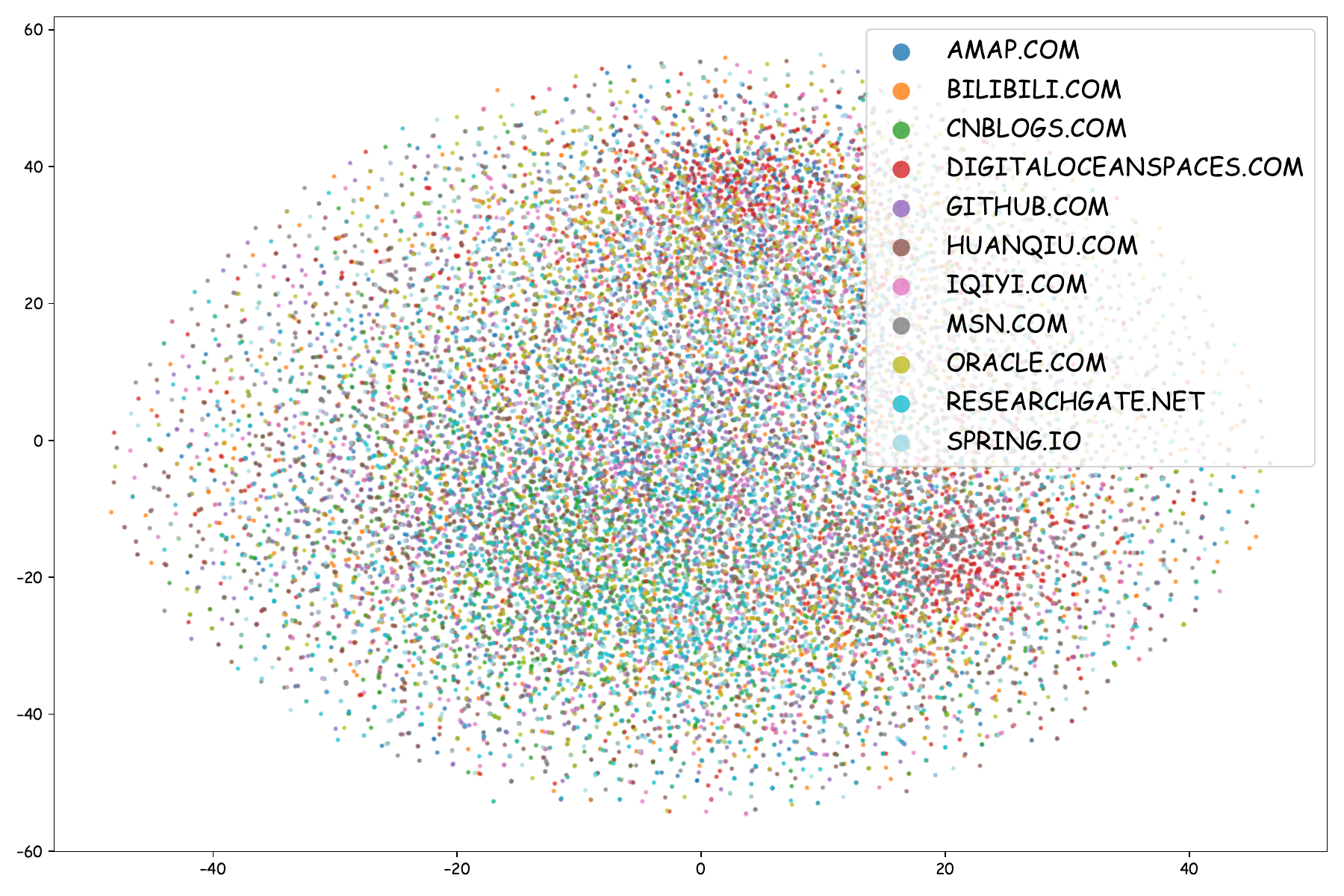} &
		\includegraphics[width=0.126\textwidth]{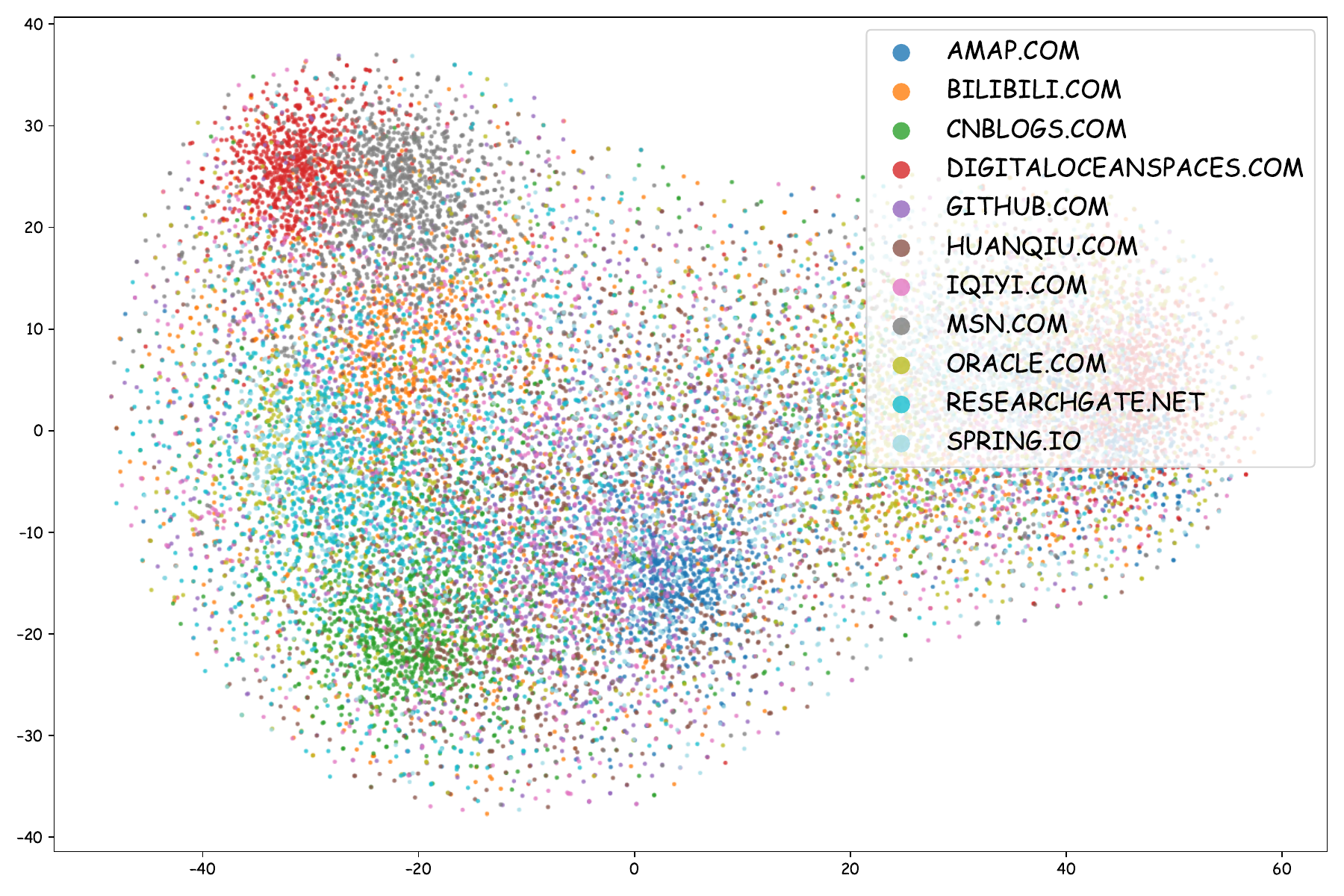} &
		\includegraphics[width=0.126\textwidth]{img/TLS-denoise/step_0200_guide.pdf} &
		\includegraphics[width=0.126\textwidth]{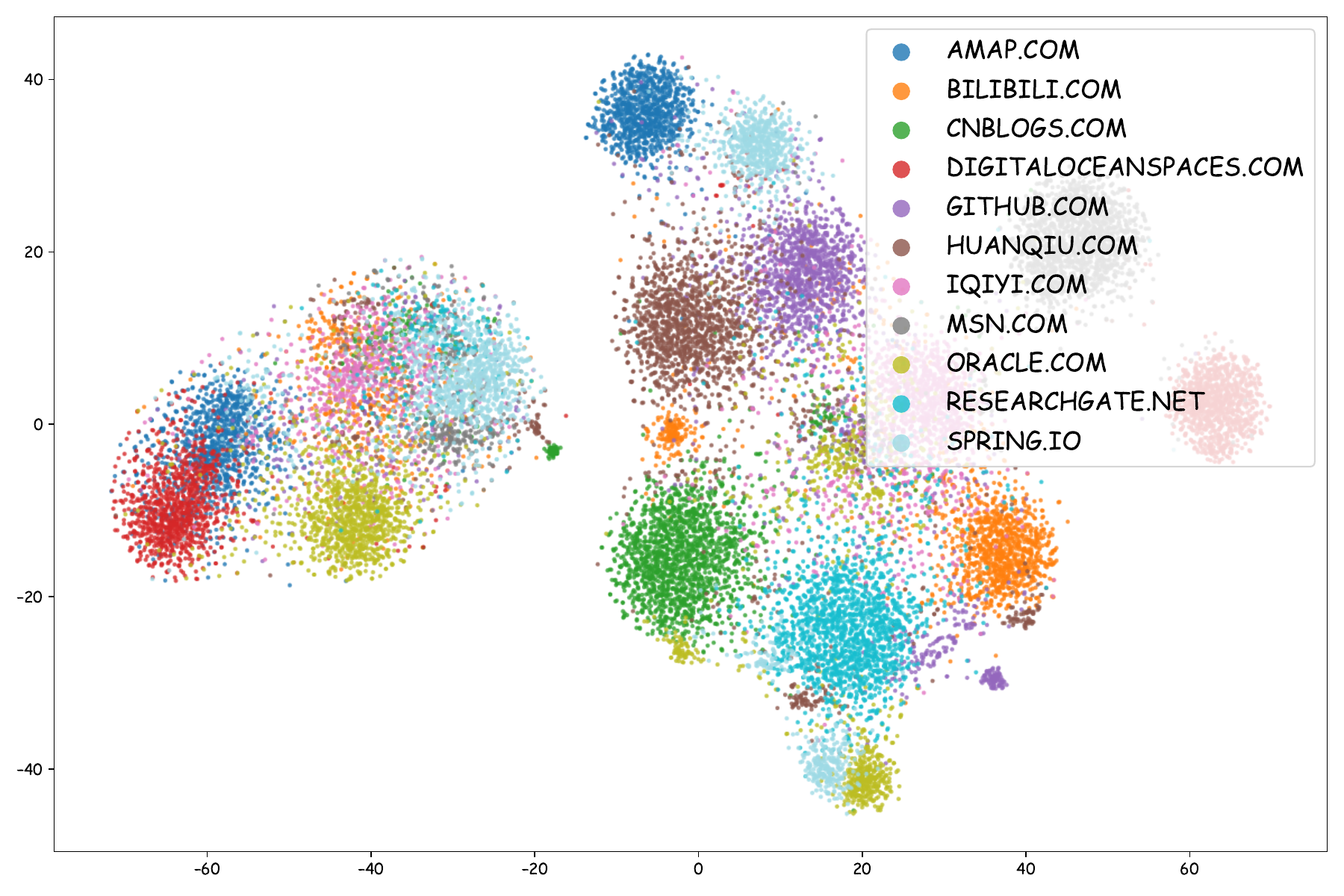} &
		\includegraphics[width=0.126\textwidth]{img/TLS-denoise/step_0000_guide.pdf} \\
		\bottomrule
	\end{tabular}
	\caption{Comparison of Reverse Denoising Processes in IDS, TLS, and USTC Datasets (UG: unguided, G: guided).}
	\label{fig:denoise_step_add}
\end{figure*}

This section collects figures and tables that cannot be fully presented in the main text due to page constraints. It streamlines the main manuscript and enhances the clarity of core illustrations therein.

\subsection{Per-class Performance on USTC-20class}

In the main text, Figure~\ref{fig:pie_chart} only presents the per-class F1-scores of TDDM-Melatt on the USTC-20class dataset after TDDM-based data augmentation. As a supplementary counterpart to Figure~\ref{fig:pie_chart}, Figure~\ref{fig:0} further reports both Accuracy (AC) and F1-score metrics.

\subsection{Comparison of Reverse Denoising Processes}

Figure~\ref{fig:denoise_step} visualizes the denoising diffusion process with and without guidance for partial classes on the IDS and TLS datasets. For comprehensive comparison, Figure~\ref{fig:denoise_step_add} further provides the contrast results of guided and unguided denoising diffusion across more timesteps over three datasets: IDS, USTC, and TLS.

\section{Discussion on Model Generalization Limitations}
\label{Discussion}

\begin{table}[htbp]
	\centering
	{
		\caption{AC and F1 Changes after Integrating TDDM on Different Models and Datasets.}
		\label{tab:tddm_perf_delta}
		\begin{tabular*}{\linewidth}{@{\extracolsep{\fill}} l|c|cc|cc@{}}
			\toprule
			\multirow{2}{*}{Model} & \multirow{2}{*}{Network} & \multicolumn{2}{c|}{TLS-120class} & \multicolumn{2}{c}{VPN-16class} \\
			\cmidrule(lr){3-4} \cmidrule(lr){5-6}
			& & $\Delta$ AC & $\Delta$ F1 & $\Delta$ AC & $\Delta$ F1 \\
			\midrule
			ET-BERT    & BERT  & +0.0154 & +0.0190 & +0.1975 & +0.1412 \\
			NetMamba   & Mamba & $-$0.0623 & $-$0.0346 & +0.0054 & $-$0.0126 \\
			YaTC       & ViT   & $-$0.1193 & $-$0.0729 & $-$0.0981 & $-$0.0661 \\
			\bottomrule
		\end{tabular*}
	}
\end{table}

While TDDM-Melatt demonstrates strong generalization through its memory-decoupled architecture, we acknowledge that the effectiveness of the TDDM data augmentation module is not uniformly transferable across all base models, as TDDM is originally designed to adapt to Melatt. We integrate TDDM with three representative SOTA models—ET-BERT, NetMamba, and YaTC—on the TLS-120class and VPN-16class tasks, and report the performance changes in Table~\ref{tab:tddm_perf_delta}.

The results reveal a significant disparity: TDDM yields substantial improvements on ET-BERT (F1 +1.90\% on TLS-120class, +14.12\% on VPN-16class), yet causes notable degradation on NetMamba (F1 -3.46\% on TLS-120class) and YaTC (F1 -7.29\% on TLS-120class, -6.61\% on VPN-16class). We attribute this disparity to the differences in preprocessing paradigms and feature representations among these models. ET-BERT employs a BERT-style pre-training that learns contextualized byte-level representations; the additional augmented samples from TDDM provide complementary statistical patterns that enrich this representation space. In contrast, YaTC and NetMamba rely on statistical features or raw payload sequences that are more tightly coupled to the specific feature distributions of the training dataset. When TDDM introduces synthetic samples that slightly deviate from the original training distribution, these models, being more sensitive to the exact feature manifold, suffer from degraded decision boundaries.

This observation highlights a critical limitation of our approach: the effectiveness of TDDM is contingent on the compatibility between the base model's representation learning paradigm and the nature of the augmented features. For models that already incorporate robust pre-training on diverse data distributions, TDDM serves as an effective complementary augmentation strategy. However, for models that rely heavily on task-specific shortcut features or have limited capacity to accommodate distributional shifts, naive application of TDDM may introduce noise that undermines performance. This finding underscores the importance of considering the interaction between data augmentation and model architecture when designing generalizable traffic classification systems, and points to the need for model-aware augmentation strategies in future work.

\end{document}